\documentclass{article}
\DeclareUnicodeCharacter{2744}{\textsnowflake}

\PassOptionsToPackage{numbers, compress}{natbib}
\usepackage[preprint]{neurips_2024}
\usepackage{tikz}
\usetikzlibrary{shapes,arrows,positioning,fit,calc,backgrounds}

\usepackage[T1]{fontenc}    
\usepackage{url}            
\usepackage{booktabs}       
\usepackage{amsfonts}       
\usepackage{nicefrac}       
\usepackage{microtype}      
\usepackage{lettrine}
\usepackage{colortbl}
\usepackage{booktabs}
\usepackage{longtable}
\usepackage{multirow}
\usepackage{array}
\usepackage{xcolor}
\usepackage{caption}
\usepackage{booktabs}
\usepackage{multirow}
\usepackage{xcolor}
\newcolumntype{L}{>{\raggedright\arraybackslash}p{4cm}}
\newcolumntype{M}{>{\raggedright\arraybackslash}p{3.2cm}}
\newcolumntype{S}{>{\raggedright\arraybackslash}p{2.5cm}}

\usepackage{pdflscape}
\usepackage{booktabs}
\usepackage{multirow}
\usepackage{graphicx}
\usepackage{amsmath}
\usepackage{amsfonts}
\usepackage{amssymb}
\usepackage[colorlinks,urlcolor=blue,linkcolor=blue,citecolor=blue]{hyperref}
\usepackage{float}
\usetikzlibrary{trees,shapes,arrows,positioning,calc,fit}
\usepackage{color}
\newcolumntype{Z}{>{\setbox0=\hbox\bgroup}c<{\egroup}@{\hspace*{-\tabcolsep}}}
\usepackage{longtable}
\usepackage{graphicx}
\usepackage{multirow}
\usepackage[utf8]{inputenc}
\usepackage{amsmath, amssymb, amsfonts}
\usepackage{booktabs}
\usepackage{fontawesome5}
\usepackage{times}
\usepackage{latexsym}
\usepackage{blindtext}
\usepackage{titlesec}
\usepackage{microtype}
\usepackage{graphicx}
\usepackage{subfigure}
\usepackage{booktabs} 
\usepackage{colortbl}
\usepackage{multirow}
\usepackage{amsmath}
\usepackage{amssymb}
\usepackage{mathtools}
\usepackage{amsthm}
\usepackage{geometry}

\usepackage{minitoc}

\theoremstyle{plain} 

\usepackage[T1]{fontenc}

\usepackage{inconsolata}

\usepackage{subcaption}

\usepackage{algorithm2e}  
\usepackage{algorithmic}
\usepackage{wrapfig}

\title{PEFT A2Z: Parameter-Efficient Fine-Tuning Survey for Large Language and Vision Models}

\author{
\textbf{Nusrat Jahan Prottasha}\textsuperscript{1}, 
\textbf{Upama Roy Chowdhury}\textsuperscript{2}\thanks{Equal contribution.}, 
\textbf{Shetu Mohanto}\textsuperscript{3}\footnotemark[1], 
\textbf{Tasfia Nuzhat}\textsuperscript{4}\thanks{Equal contribution.}, \\ 
\textbf{Abdullah As Sami}\textsuperscript{5}\footnotemark[2], 
\textbf{Md Shamol Ali}\textsuperscript{6}\footnotemark[2], 
\textbf{Md Shohanur Islam Sobuj}\textsuperscript{7}, 
\textbf{Hafijur Raman}\textsuperscript{1}, \\
\textbf{Md Kowsher}\textsuperscript{1}, 
\textbf{Ozlem Ozmen Garibay}\textsuperscript{1}\\
\textsuperscript{1}University of Central Florida, USA,  \textsuperscript{2}Khulna University of Engineering \& Technology, Bangladesh,  \\ 
\textsuperscript{3}Delineate Inc. USA, \textsuperscript{4}Universiti Tenaga Nasional, Malaysia,   \textsuperscript{5}University of South Florida, USA,  \\
\textsuperscript{6}Daffodil International University, Bangladesh,  \textsuperscript{7}Anymate Me, Germany \\
\faGithub~\textcolor{red}{\href{https://github.com/Nusrat-Prottasha/PEFT-A2Z}{\textcolor{red}{\texttt{https://github.com/Nusrat-Prottasha/PEFT-A2Z}}}}
}

\begin{document}
\maketitle

\begin{center}
\noindent
\rotatebox{0}{
\hspace*{-6.0em} 
    \includegraphics[height=0.95\textwidth]{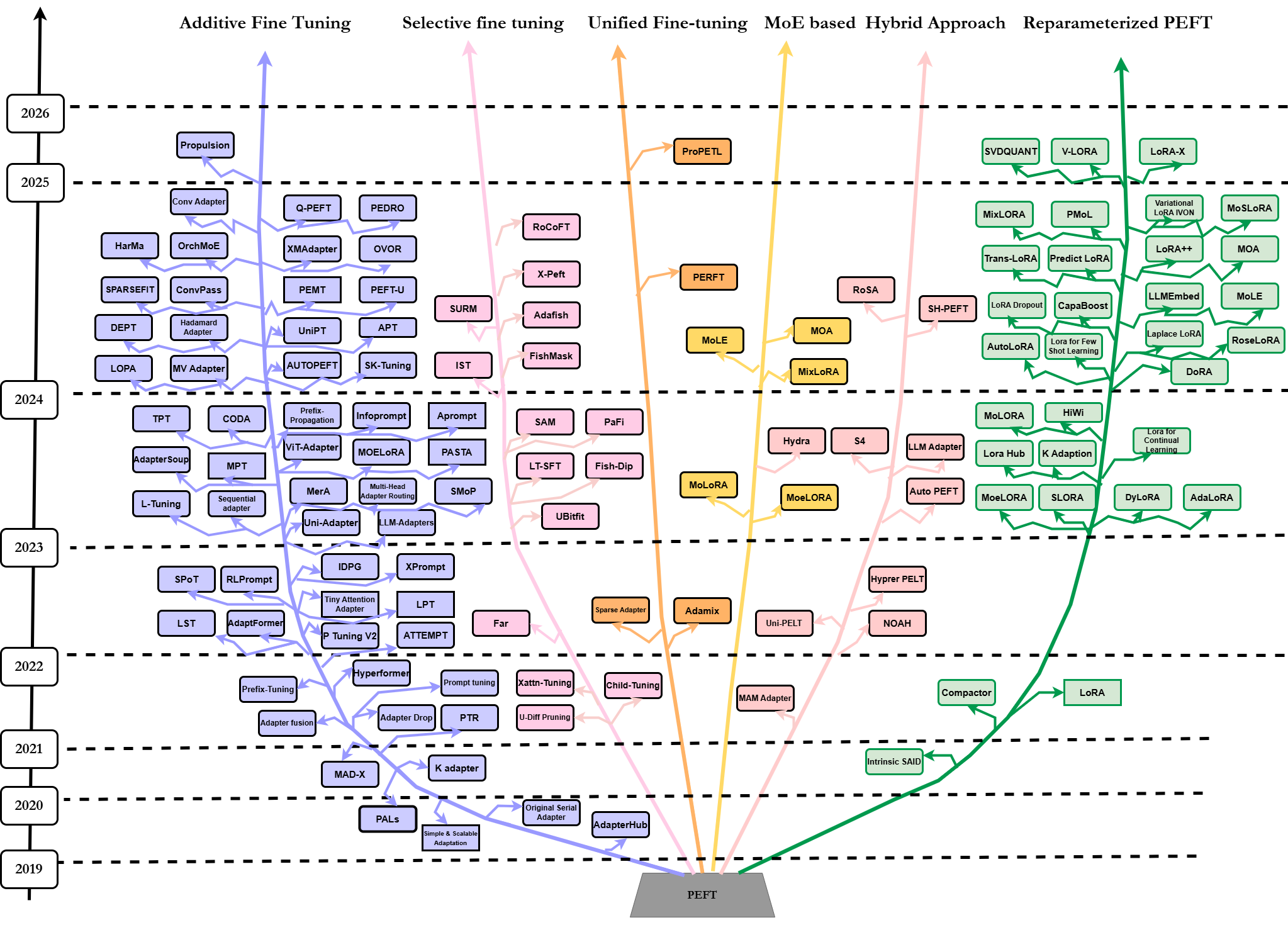}
}
\vspace{0.5em}
\noindent
\small
Evolution of PEFT methods from 2019 to 2025.
\end{center}

\vspace{1em} 

\begin{abstract}
Large models such as Large Language Models (LLMs) and Vision Language Models (VLMs)  have transformed artificial intelligence, powering applications in natural language processing, computer vision, and multimodal learning. However, fully fine-tuning these models remains expensive, requiring extensive computational resources, memory, and task-specific data. Parameter-Efficient Fine-Tuning (PEFT) has emerged as a promising solution that allows adapting large models to downstream tasks by updating only a small portion of parameters. This survey presents a comprehensive overview of PEFT techniques, focusing on their motivations, design principles, and effectiveness. We begin by analyzing the resource and accessibility challenges posed by traditional fine-tuning and highlight key issues, such as overfitting, catastrophic forgetting, and parameter inefficiency. We then introduce a structured taxonomy of PEFT methods—grouped into additive, selective, reparameterized, hybrid, and unified frameworks—and systematically compare their mechanisms and trade-offs. Beyond taxonomy, we explore the impact of PEFT across diverse domains, including language, vision, and generative modeling, showing how these techniques offer strong performance with lower resource costs. We also discuss important open challenges in scalability, interpretability, and robustness, and suggest future directions such as federated learning, domain adaptation, and theoretical grounding. Our goal is to provide a unified understanding of PEFT and its growing role in enabling practical, efficient, and sustainable use of large models.
\end{abstract}

\tableofcontents

\section{Introduction}
\label{sec:introduction}



\lettrine{L}{arge} Language Models (LLMs) \cite{Steyvers2025, jin2024position} and Pre-trained Language Models (PLMs)  \cite{luo2024pre, ma2020charbert,schramowski2022large, wu2019enriching}  have revolutionized artificial intelligence \cite{venkatesan2023high, daneshvar2024artificial} , driving transformative advancements across domains such as Natural Language Processing (NLP)  \cite{chowdhary2020natural,nabi2021sondhan} , Computer Vision (CV)  \cite{ballard1982computer,wiley2018computer} , and multimodal learning  \cite{brown2020language,giannakos2023role,pang2015deep}. Built on billions of parameters and trained on vast datasets, these models have demonstrated unparalleled capabilities in applications like text generation \cite{mckeown1992text, de2021survey}, language translation  \cite{janfaza2012language,al2015translation}  , conversational agents  \cite{lester2004conversational, rubin2010artificially}, Chatbot  \cite{kowsher2024token,satu2015review}, and content summarization  \cite{abualigah2020text,brown2021gaia}. These breakthroughs have redefined the possibilities of artificial intelligence \cite{o2023artificial},  making substantial contributions to academia, industry, and real-world applications  \cite{hadi2023survey,sarker2021machine,sarker2021machine}.

\begin{figure}[htbp]
  \centering
  \includegraphics[width=1.0\textwidth]{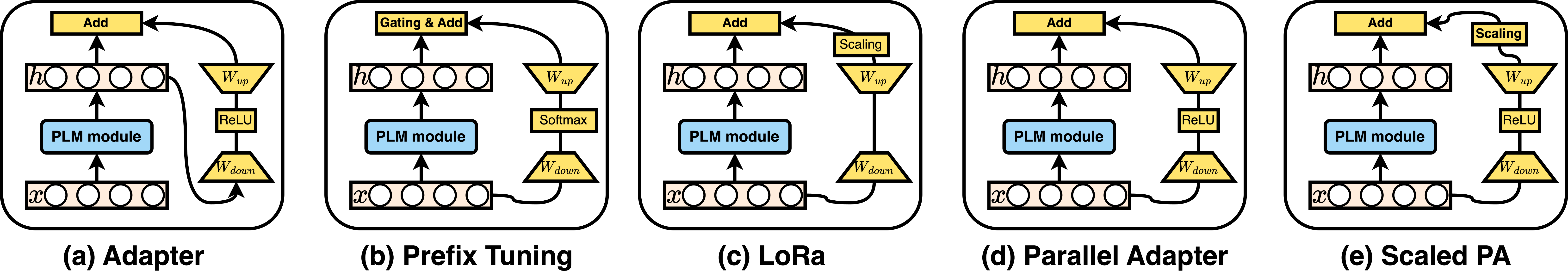}
  \caption{Overview of key PEFT techniques: Adapter, Prefix Tuning, LoRA, Parallel Adapter, and Scaled Parallel Adapter \cite{he2021towards}}.
  \label{fig:peft_overview_of}
\end{figure}

Despite their immense potential, the size and complexity of modern LLMs and Pretrained Language Models (PLMs) \cite{li2024pre,kumar2024large} continue to pose profound challenges to both research and industrial communities  \cite{raiaan2024review, zhao2023survey}. Consider, for example, LLama-3---  \cite{lu2024benchmarking,kumar2024fine}arguably one of the most sophisticated and computationally demanding LLMs available  \cite{dubey2024llama, yu2025your}. Its architecture, featuring on the order of $300$ billion parameters and employing intricate, multi-head attention mechanisms \cite{huang2022classification,feng2021short,baan2019understanding,pecoraro2022local,zhou2023multi} , achieves state-of-the-art benchmarks across a breadth of tasks \cite{pandey2021state,nguyen2023brief}.Yet, diespite theise remtarkable captabilitiees, the fine-tuning process for such a model is nontrivial  \cite{albeverio1997nontrivial,david1985model}. It entails mobilizing immense computational infrastructures, including petabyte-scale storage systems, ultra-high memory bandwidth interfaces, and extensive arrays of cutting-edge GPUs   \cite{lim2024cutting,wael2025accelerating}. For example, effective fine-tuning of LLama-3   \cite{dubey2024llama,mai2024financial} necessitates provisioning compute clusters that may incorporate hundreds to thousands of high-end, data-center-grade GPUs---often NVIDIA A100 or H100 units  \cite{hou2024fine,zheng2024data}. Each of these sophisticated processors comes equipped with tens of gigabytes of high-bandwidth memory (HBM), yet even this generous memory footprint proves insufficient for accommodating the entirety of LLama-3’s parameter set, intermediate activations, and optimizer states on any single device. Multiple GPUs \cite{schaa2009exploring,chen2010dynamic} must thus be aggregated to host the model and its associated training workflow. Achieving the requisite efficiency in this context demands careful orchestration of distributed training paradigms  \cite{mcdonald2010distributed}, including tensor parallelism  \cite{wang2022tesseract,kurita1961tensor,wagenlander2024tenplex}, pipeline parallelism  \cite{huang2019gpipe,thies2007practical,narayanan2019pipedream,yildirim2015application}, and model sharding  \cite{lepikhin2020gshard, li2023survey} all of which must be meticulously tuned to maintain throughput and ensure balanced workload distribution across the GPU ensemble. By necessity, such infrastructural complexity and the corresponding operational overheads place significant resource constraints on the fine-tuning process, effectively limiting the accessibility and deployability of models at this scale  \cite{duan2024efficient}.

As traditional fine-tuning \cite{li2024instruction,swati2019brain} involves updating all model parameters for each new task, which becomes prohibitively expensive as model sizes grow, addressing this knowledge gap is essential for maximizing the potential of LLMs and PLMs  \cite{ swarup2024maximizing} Optimizing their deployment and fine-tuning \cite{tajbakhsh2016convolutional,radenovic2018fine} processes would not only reduce computational demands but also enhance their adaptability to a wide range of tasks, ensuring that these models remain impactful across diverse applications \cite{sharma2020review,wilson2013diverse}. Bridging this gap is crucial for democratizing their use, enabling resource-constrained organizations to harness the power of LLMs like LLama-3 \cite{repede2024llama,lin2024enhanced}  and apply them in emerging fields \cite{nie2009approach,hervas2015clusters}. 

The central research question driving this study is: What are the resource requirements and fine-tuning challenges associated with LLMs and PLMs \cite{yue2024building,shafikuzzaman2024empirical}, and how can they be addressed to optimize their deployment and fine-tuning This question seeks to uncover critical limitations and explore strategies to enhance the efficiency and accessibility of these models.

The aim of this study is to investigate the computational and fine-tuning challenges associated with LLMs, VLMs, and LMMs \cite{ma2024does,rahimi2025user,schoepp2025evolving} and to identify strategies for optimizing their deployment and fine-tuning processes \cite{adenso2006fine,chua2021fine}. Through systematic analysis, this study intends to provide actionable insights to guide researchers and practitioners in overcoming the limitations of these models. 

We hypothesize that LLMs require substantial computational resources and fine-tuning expertise to achieve optimal performance. However, strategies such as parameter-efficient fine-tuning (PEFT)— \cite{peng2024q,tran2024deberta,kim2023memory}which selectively updates only a small subset of model parameters—can significantly reduce resource requirements while maintaining or enhancing performance  \cite{xin2024parameter}. By exploring and validating these approaches, this study aims to contribute to the broader understanding and democratization of LLMs and PLMs, paving the way for their effective use in AI research and applications \cite{sarker2022ai,haleem2019current}. 

PEFT  \cite{lin2024peft}methods offer a promising alternative by significantly reducing the number of trainable parameters \cite{chatelain2022number,thakur2022incremental, kanavati2021partial} making fine-tuning more accessible, scalable, and sustainable. Techniques such as adapter modules, prefix-tuning \cite{mai2023prefix,peng2024deja,chen2022developing,lu2024few} LoRA \cite{andrade2019comprehensive,demetri2019automated,kunz2024train} (Low-Rank Adaptation), BitFit, and prompt tuning have demonstrated strong empirical performance across a variety of benchmarks, often matching or surpassing full fine-tuning with only a fraction of the computational cost. These methods are particularly valuable in real-world scenarios, where practitioners must handle multiple tasks, work within resource constraints, or deploy models on edge devices.

Despite the growing popularity of PEFT, there is still a lack of systematic understanding of the design space, trade-offs, and applicability of these techniques across different modalities. This survey aims to fill that gap by offering a comprehensive review of parameter-efficient fine-tuning methods for both language and vision models \cite{xing2024survey} . We begin by analyzing the computational and memory limitations of standard fine-tuning, followed by a discussion of its inherent drawbacks. We then present a unified taxonomy that categorizes PEFT approaches into five major classes: additive, selective, reparameterized, hybrid, and unified methods. This taxonomy provides a structured lens through which to understand and compare different strategies.

Furthermore, we evaluate the application of PEFT across domains, including NLP  \cite{joshi1991natural,dreisbach2019systematic}, computer vision, multimodal tasks, and generative modeling. We highlight how PEFT methods contribute to improved efficiency, better generalization, and more responsible AI deployment. Lastly, we identify key challenges and open questions in the field, such as interpretability, theoretical foundations, and domain-specific adaptation \cite{chang2019domain,wei2018general}, and we suggest future directions for research.

Through this survey, we aim to provide researchers and practitioners with a clear and comprehensive guide for parameter-efficient fine-tuning \cite{tang2024low,liu2024cpmi}, empowering them to build more efficient and adaptable AI systems.

\section{Main Contributions}
\label{sec:contributions}

To summarize, the main contributions of this survey can be outlined as follows:

\begin{itemize}
    \item \textbf{Comprehensive Resource Analysis}: We examine the computational, memory, and storage demands associated with full fine-tuning of large-scale pre-trained models (PLMs and LLMs), emphasizing practical constraints faced by researchers with limited access to infrastructure.
    
    \item \textbf{Critical Evaluation of Fine-Tuning Limitations}: We discuss the limitations of conventional fine-tuning approaches, such as overfitting on low-resource tasks, catastrophic forgetting in continual learning, redundancy in parameter updates, and scalability bottlenecks.
    
    \item \textbf{Unified Taxonomy of PEFT Methods}: We propose a structured taxonomy categorizing PEFT techniques into five key families—\textit{additive, selective, reparameterized, hybrid, and unified}—to offer a clear lens for comparing design strategies and identifying common patterns.
    
    \item \textbf{Comparison of Representative PEFT Techniques}: We provide a side-by-side evaluation of widely-used methods such as LoRA, adapters, BitFit, prompt tuning, and prefix-tuning, analyzing their parameter efficiency, performance trade-offs, and implementation complexity.
    
    \item \textbf{Cross-Domain Application Survey}: We survey the application of PEFT in diverse domains, including NLP, computer vision, multimodal learning, speech, and generative modeling, highlighting their robustness, transferability, and real-world usability.
    
    \item \textbf{Adaptation in Specialized Settings}: We explore how PEFT methods are applied in emerging areas such as continual learning, federated learning, privacy-preserving fine-tuning, domain adaptation, and low-resource language support.
    
    \item \textbf{Empirical Insights and Trends}: We summarize recent experimental findings and performance benchmarks to uncover trends in PEFT research and identify the conditions under which specific methods excel or fail.
    
    \item \textbf{Open Challenges and Future Directions}: We outline open problems in the field, including scaling PEFT to ultra-large models, enhancing interpretability, improving theoretical understanding, and integrating PEFT with efficient inference strategies.
    
    \item \textbf{Accessible Summary and Practical Guidelines}: We provide an actionable guide to help practitioners choose appropriate PEFT methods based on resource budgets, task types, and model architectures.
\end{itemize}

This paper is organized as follows:

In \textbf{Section~\ref{sec:introduction}}, we introduce the background and motivation for this work, highlighting the rise of large-scale foundation models such as Large Language Models (LLMs), Vision Large Models (VLMs), and Large Multimodal Models (LMMs), and the need for parameter-efficient fine-tuning (PEFT) approaches to mitigate the high computational and resource costs of full fine-tuning.

In \textbf{Section~\ref{sec:contributions}}, we outline the key contributions of this survey, including a systematic taxonomy of PEFT methods, an evaluation of their trade-offs, and an in-depth discussion of their applications and limitations across domains.

In \textbf{Section~\ref{sec:preliminaries}}, we present the necessary preliminaries for understanding PEFT, including attention mechanisms, self-attention, multi-head configurations, transformer architecture, and the inherent inefficiencies of full fine-tuning, supported by complexity and scaling analyses.

In \textbf{Section~\ref{sec:design}}, we detail the key architectural and practical considerations in the design of PEFT strategies, including design goals, quantized decision spaces, task-adaptive routing mechanisms, and optimization strategies for memory, time, and energy efficiency, especially in multimodal contexts.

In \textbf{Section~\ref{sec:peftmethods}}, we present key PEFT methods, including additive fine-tuning with serial and parallel adapters, hybrid adapters for task-specific adaptation, soft prompt tuning, and reparameterized approaches such as LoRA. We also cover scaling behaviors, selective fine-tuning, and emerging hybrid frameworks such as MoE-based PEFT.

In \textbf{Section~\ref{sec:experiments}}, we evaluate the performance of PEFT methods through empirical comparisons on benchmark datasets, including GLUE for NLP tasks and reasoning evaluations on large language models, highlighting parameter-to-performance trade-offs.

In \textbf{Section~\ref{sec:applications}}, we explore the application of PEFT techniques across diverse domains, including natural language processing, computer vision, multimodal learning, and robotics, emphasizing their adaptability and domain-specific benefits.

In \textbf{Section~\ref{sec:complexitypeft}}, we analyze the computational, memory, and scaling complexities associated with different PEFT strategies, offering comparative insights into their theoretical and practical efficiency.

In \textbf{Section~\ref{sec:strengths}}, we summarize the strengths and limitations of PEFT methods, focusing on their parameter efficiency, adaptability, generalization, and constraints in real-world deployment.

In \textbf{Section~\ref{sec:discussion}}, we identify key limitations in current PEFT methods, including heuristic reliance, lack of theory, poor interpretability, and limited standardization—emphasizing the need for semantically aware and architecture-sensitive designs.

In \textbf{Section~\ref{sec:futuredirections}}, we outline promising future research directions, including theoretical modeling of parameter influence, layer-wise tuning strategies, continual learning integration, interpretability, benchmarking, and privacy-aware PEFT.

In \textbf{Section~\ref{sec:conclusion}}, we conclude the paper by reflecting on the role of PEFT in enabling efficient and scalable adaptation of large foundation models, and its significance for the future of resource-aware AI.


\section{PRELIMINARIES}
\label{sec:preliminaries}

\subsection{Attention Mechanisms}

Attention mechanisms  \cite{bahdanau2014neural, vaswani2017attention,kowsher2024does} enable a model to focus on specific parts of an input sequence to produce representations for downstream tasks. Let $\mathbf{X} \in \mathbb{R}^{n \times d_{\text{model}}}$ represent a sequence of $n$ input token embeddings, each of dimension $d_{\text{model}}$. The goal of attention is to combine these token representations into contextualized outputs by weighting their relevance.

To achieve this, the input $\mathbf{X}$ is mapped into three sets of vectors: queries $\mathbf{Q}$, keys $\mathbf{K}$, and values $\mathbf{V}$. Typically, these are given by:
\begin{equation}
\mathbf{Q} = \mathbf{X}\mathbf{W}^Q, \quad \mathbf{K} = \mathbf{X}\mathbf{W}^K, \quad \mathbf{V} = \mathbf{X}\mathbf{W}^V,
\end{equation}
where $\mathbf{W}^Q, \mathbf{W}^K, \mathbf{W}^V \in \mathbb{R}^{d_{\text{model}} \times d_k}$ are trainable projection matrices and $d_k$ is the dimension of each head’s projection space (for single-head attention). These projections enable the computation of pairwise compatibilities between queries and keys, determining how much each token should attend to others.

The core computation of attention is often implemented as scaled dot-product attention. Given $\mathbf{Q}, \mathbf{K}, \mathbf{V}$, we define:
\begin{equation}
\text{Attention}(\mathbf{Q}, \mathbf{K}, \mathbf{V}) = \text{softmax}\left(\frac{\mathbf{Q}\mathbf{K}^\top}{\sqrt{d_k}}\right)\mathbf{V}.
\end{equation}

The dot product $\mathbf{Q}\mathbf{K}^\top \in \mathbb{R}^{n \times n}$ computes the pairwise compatibility between each query vector and each key vector. Without scaling, the magnitude of the dot products increases with dimension $d_k$, potentially affecting training stability. Dividing by $\sqrt{d_k}$ normalizes the variance of the input features, making the softmax distribution less extreme and stabilizing training. Applying the softmax row-wise converts raw alignment scores into a probability distribution, ensuring that attention weights are non-negative and sum to 1. Finally, the output is a weighted combination of the values $\mathbf{V}$, using the attention weights computed by the softmax.

\subsection{Self-Attention}

In self-attention  \cite{guo2022beyond,zhang2019self,gao2020self}, the queries, keys, and values come from the same sequence:
\begin{equation}
\mathbf{Z} = \text{softmax}\left(\frac{\mathbf{X}\mathbf{X}^\top}{\sqrt{d_k}}\right)\mathbf{X}.
\end{equation}

This avoids explicit recurrence or convolution and provides $O(n^2)$ complexity in sequence length $n$, allowing the model to capture long-range dependencies effectively.

\subsection{Multi-Head Attention}

Multi-Head Attention (MHA)  \cite{vaswani2017attention,long2019sentiment,bhojanapalli2020low} generalizes single-head attention by using $H$ parallel attention heads \cite{medina2018parallel,bilonoh2020parallel}. Each head focuses on a different projection of the input, providing richer modeling capacity. Let $\mathbf{Q}_h, \mathbf{K}_h, \mathbf{V}_h$ denote the projections for head $h$:
\begin{align}
\mathbf{Q}_h &= \mathbf{X}\mathbf{W}^Q_h, \quad \mathbf{K}_h = \mathbf{X}\mathbf{W}^K_h, \quad \mathbf{V}_h = \mathbf{X}\mathbf{W}^V_h, \\
&\text{where } \mathbf{Q}_h, \mathbf{K}_h, \mathbf{V}_h \in \mathbb{R}^{n \times d_{\text{head}}}, \; d_{\text{model}} = H \cdot d_{\text{head}}.
\end{align}

Each head computes scaled dot-product attention \cite{bernhard2023alternatives} independently:
\begin{equation}
\mathbf{Z}_h = \text{Attention}(\mathbf{Q}_h, \mathbf{K}_h, \mathbf{V}_h).
\end{equation}

The outputs from all $H$ heads are then concatenated and transformed by an output projection $\mathbf{W}^O \in \mathbb{R}^{(H d_{\text{head}}) \times d_{\text{model}}}$:
\begin{equation}
\mathbf{Z} = [\mathbf{Z}_1; \mathbf{Z}_2; \cdots; \mathbf{Z}_H]\mathbf{W}^O.
\end{equation}

MHA allows the model to jointly attend to information from different representation subspaces, improving its ability to capture complex patterns.

\subsection{Transformer Architecture}

Within the Transformer architecture, a pivotal component is the \textbf{Multi-Head Self-Attention (MHSA)} \cite{leng2021using,liu2021attention} mechanism, which allows the model to attend to different representation subspaces simultaneously. Formally, given an input matrix \( X \in \mathbb{R}^{n \times d_{\text{model}}} \), where \( n \) is the sequence length and \( d_{\text{model}} \) is the dimensionality of the model, the MHSA operates by first linearly projecting \( X \) into three distinct matrices: queries \( Q \), keys \( K \), and values \( V \). These projections are achieved through learnable weight matrices \( W^Q, W^K, W^V \in \mathbb{R}^{d_{\text{model}} \times d_k} \), such that:

\[
Q = XW^Q, \quad K = XW^K, \quad V = XW^V.
\]

Each attention head \( h \) computes scaled dot-product attention as previously defined:

\[
\text{Attention}(Q_h, K_h, V_h) = \text{softmax}\left(\frac{Q_h K_h^\top}{\sqrt{d_k}}\right) V_h.
\]

For \( H \) parallel heads, the outputs are concatenated and projected back to the original model dimension using a weight matrix \( W^O \in \mathbb{R}^{(H \cdot d_k) \times d_{\text{model}}} \):

\begin{align}
\text{MHSA}(X) &= \text{Concat} \Big( \text{Attention}(Q_1, K_1, V_1), \dots, \text{Attention}(Q_H, K_H, V_H) \Big) \times W^O \\
&= \left[ \text{Attention}(Q_1, K_1, V_1) \; || \; \dots \; || \; \text{Attention}(Q_H, K_H, V_H) \right] W^O.
\end{align}

This multi-head approach enables the model to capture diverse aspects of the input by allowing each head to focus on different parts or features of the sequence.

Following the MHSA layer, the Transformer employs a \textbf{Position-Wise Feed-Forward Network (FFN)} \cite{hopfield1987learning,rosay2020feed}, which applies two linear transformations with a non-linear activation function in between. The FFN operates independently on each position in the sequence, enhancing the model's capacity to learn complex patterns. Mathematically, the FFN is defined as:

\[
\text{FFN}(x) = \max(0, xW_1 + b_1) W_2 + b_2,
\]

where \( W_1 \in \mathbb{R}^{d_{\text{model}} \times d_{\text{ff}}} \) and \( W_2 \in \mathbb{R}^{d_{\text{ff}} \times d_{\text{model}}} \) are weight matrices, and \( b_1, b_2 \) are bias vectors. The activation function \( \max(0, \cdot) \) introduces non-linearity, allowing the network to model more complex relationships within the data.

Both the MHSA and FFN  \cite{liu2023efficientvit,cao2023ghostvit}layers are integrated with \textbf{residual connections} and \textbf{layer normalization}  \cite{scholkemper2024residual,kobayashi2021incorporating} to facilitate stable and efficient training. Specifically, the output of each sub-layer is added to its input and then normalized:

\[
\text{Output}_{\text{MHSA}} = \text{LayerNorm}(X + \text{MHSA}(X)),
\]

\[
\text{Output}_{\text{FFN}} = \text{LayerNorm}(\text{Output}_{\text{MHSA}} + \text{FFN}(\text{Output}_{\text{MHSA}})).
\]

These residual connections help in mitigating the vanishing gradient problem, enabling the training of deep Transformer models by allowing gradients to flow more effectively through the network.

\subsection{Pretraining Language Model}

Language models often utilize massive unlabeled corpora for pretraining to develop robust language representations. One prevalent approach is \textbf{Masked Language Modeling (MLM)} \cite{sinha2021masked,nozza2020mask,kawintiranon2021knowledge} where a subset \( M \) of positions within the input sequence is masked, and the model is tasked with predicting these masked tokens. The loss function for MLM is defined as:

\begin{equation}
L_{\text{MLM}} = -\sum_{i \in M} \log P(x_i \mid x_{\setminus i}),
\end{equation}

where \( x_{\setminus i} \) denotes the sequence with the \( i \)-th token masked. This Application encourages the model to understand the context surrounding the masked positions to accurately predict the missing tokens.

Another fundamental approach is \textbf{Autoregressive (AR) Language Modeling} \cite{yu2022non,wu2023ar}, where the model predicts each token based on all preceding tokens in the sequence. The loss function for AR modeling is expressed as:

\begin{equation}
L_{\text{AR}} = -\sum_{i=1}^{n} \log P(x_i \mid x_{<i}).
\end{equation}

In this formulation, \( x_{<i} \) represents all tokens before the \( i \)-th position, allowing the model to generate coherent and contextually relevant sequences. Both MLM and AR Applications contribute to learning representations that are highly general and transferable, enabling the pretrained models to perform effectively across a wide range of downstream tasks.

\subsection{Full Fine-Tuning} 

Full fine-tuning \cite{zhou2017fine} involves updating all parameters of a pre-trained large language model (LLM) to adapt it to a specific downstream task. Let $\boldsymbol{\theta} \in \mathbb{R}^p$ represent the model's parameters, where $p$ typically spans billions. Given a task-specific dataset $\mathcal{D}_{\text{task}} = \{(x_i, y_i)\}_{i=1}^N$, the Application is to determine the optimal parameters $\boldsymbol{\theta}^*$ that minimize the cumulative loss:

\begin{equation}
    \boldsymbol{\theta}^* = \arg \min_{\boldsymbol{\theta}} \sum_{(x, y) \in \mathcal{D}_{\text{task}}} \mathcal{L}(f_{\boldsymbol{\theta}}(x), y),
\end{equation}

where $f_{\boldsymbol{\theta}}(x)$ is the model's prediction for input $x$, and $\mathcal{L}$ is a task-specific loss function, such as cross-entropy for classification.

Optimization typically employs gradient-based algorithms like Adam or AdamW. In each iteration, the parameters are updated as follows:

\begin{equation}
    \boldsymbol{\theta} \leftarrow \boldsymbol{\theta} - \eta \nabla_{\boldsymbol{\theta}} \mathcal{L}(f_{\boldsymbol{\theta}}(x), y),
\end{equation}

where $\eta$ is the learning rate. This comprehensive adaptation leverages the model’s full capacity, enhancing performance on the target task.

\subsection{Limitations and Challenges of Full Fine-Tuning}
However, full fine-tuning incurs significant computational and memory overheads. Storing gradients for all $p$ parameters requires substantial memory, often necessitating techniques like mixed-precision training or gradient checkpointing \cite{wang2024fastpersist,chen2024efficient} to manage resource usage. Additionally, the extensive parameter updates lead to high computational costs, making the training process time-consuming and reliant on specialized hardware such as GPUs or TPUs \cite{wang2019benchmarking}. Maintaining multiple model checkpoints further increases storage requirements, complicating deployment and scalability.

These challenges have driven the development of PEFT methods \cite{zahweh2023empirical} which aim to reduce resource consumption by updating only a subset of parameters or introducing lightweight modules. Despite these alternatives, full fine-tuning remains fundamental due to its ability to fully exploit the model’s expressive power, often resulting in superior task-specific performance.

Full fine-tuning is also prone to overfitting, especially with limited downstream data. To enhance generalization, regularization techniques like weight decay and dropout are commonly employed. Additionally, strategies such as gradual unfreezing—where layers are fine-tuned incrementally—help stabilize training and improve performance.

\subsection{Large Language Models (LLMs)}

LLMs are a class of neural networks characterized by their vast scale, with parameter counts typically ranging from hundreds of millions to hundreds of billions  \cite{hu2023llm, dubey2024llama, bi2024deepseek}. Let \( p \) denote the number of parameters in the model, and \( |D| \) represent the size of the training dataset, measured in terms of the number of tokens. The computational complexity of training such models can be approximated as \( O(|D| \cdot p) \), reflecting the operations required for forward and backward passes during gradient-based optimization. This scaling impacts both the computational cost and the memory footprint, which grows proportionally to \( O(p) \)  \cite{kaplan2020scaling, narayanan2021efficient}, assuming all parameters and activations are stored for gradient computation.

The architecture of LLMs  \cite{kumar2023large,arun2025llms} such as Transformer-based models~ \cite{vaswani2017attention,acheampong2021transformer} enables efficient parallelization via self-attention mechanisms \cite{liu2023self}. However, as \( p \) increases, training and inference require distributed computing strategies to manage memory and computational demands. Typical implementations leverage GPU/TPU clusters, where advanced techniques like mixed-precision arithmetic \cite{micikevicius2017mixed}, gradient checkpointing, and pipeline parallelism optimize performance.

Scaling laws, as empirically demonstrated by Kaplan et al.~ \cite{kaplan2020scaling}, provide a quantitative framework for understanding the relationship between model size, dataset size, and performance. These laws observe that as \( p \) and \( |D| \) increase, the performance of LLMs follows predictable power-law trends. Specifically, the loss \( L \) on a given task is approximately proportional to:
\[
L \propto p^{-\alpha} + |D|^{-\beta},
\]
with \( \alpha, \beta > 0 \) are empirically determined constants. This relationship highlights the diminishing returns of scaling, wherein gains in performance taper off as \( p \) and \( |D| \) grow beyond certain thresholds. 

While larger models exhibit improved flexibility, generalization, and capacity for in-context learning, the resource demands for full fine-tuning scale with \( O(|D| \cdot p) \). Fine-tuning such models for downstream tasks necessitates extensive compute resources, large memory footprints, and long training durations, posing significant barriers to accessibility. Moreover, full fine-tuning modifies all \( p \) parameters, leading to storage inefficiencies when maintaining task-specific variants of a single model.

The impracticality of full fine-tuning for massive LLMs underscores the importance of \textit{PEFT} techniques, which aim to adapt models using significantly fewer parameter updates. By modifying only a subset of \( p \) or introducing lightweight task-specific modules, PEFT enables adaptation with minimal resource overhead while preserving the pre-trained knowledge of the underlying LLM.

\subsection{Transfer Learning}

Transfer learning  \cite{taylor2009transfer,zhou2021transfer,shakil2023pithanet, prottasha2022transfer} is a cornerstone methodology in modern machine learning, designed to harness the knowledge embedded within a large pretrained model to enhance the performance of downstream tasks. At its core, the parameters \( \theta \) of a pretrained LLM encapsulate extensive linguistic and contextual understanding. These parameters can be effectively adapted to task-specific requirements through the fine-tuning of a small subset of parameters or by incorporating lightweight task-specific components. This paradigm demonstrates the remarkable utility of leveraging generalized pretraining to achieve task-specific excellence.

Let us consider a pretrained model \( f_{\theta_0} \) characterized by its parameters \( \theta_0 \), trained on a large corpus. Transfer learning facilitates its adaptation to a downstream task represented by the dataset \( D_{\text{task}} \). The adaptation process aims to produce an optimized model \( f_{\theta^*} \), and the optimization Application is expressed as:
\[
\theta^* = \arg\min_{\theta} \mathcal{L}(f_\theta, D_{\text{task}})
\]
In this formulation, \( \mathcal{L} \) denotes the loss function specific to the downstream task. The pretrained parameters \( \theta_0 \) serve as a robust initialization, which not only enhances the model's generalization capabilities but also acts as a form of regularization. This results in a significant reduction in the need for extensive task-specific training, making the adaptation process computationally efficient and effective.

The principles of transfer learning \cite{clark1985transfer} are integral to the development and application of PEFT methodologies. PEFT techniques—such as adapters, low-rank matrix updates, and prompt tuning—enable the fine-tuning \cite{dong2025low} of only a minimal fraction of the pretrained model's parameters. Alternatively, they introduce lightweight modular adjustments tailored to the task at hand. This strategic approach ensures that task-specific knowledge is seamlessly incorporated, while preserving the comprehensive capabilities of the pretrained model. Consequently, PEFT exemplifies the scalability and resource-efficiency required for modern machine learning applications \cite{ivanovic2015modern}, and transfer learning serves as the foundational framework that underpins its success.

\subsection{Computational Complexity}

The computational and memory demands of training and fine-tuning LLMs are substantial, primarily due to the self-attention mechanism. The \textbf{time complexity} of self-attention is \( \mathbf{O(n^2)} \) in sequence length \( n \) \cite{keles2023computational, kowsher2024does}, as each token in the input sequence attends to every other token. During pretraining, where LLMs process datasets containing billions or trillions of tokens, this quadratic complexity results in operations requiring trillions of floating-point operations (FLOPs) \cite{oberman1996implementing}. Fine-tuning further adds to this computational burden by necessitating the retraining of all model parameters for each downstream task, particularly when handling lengthy input sequences or complex datasets.

The \textbf{memory complexity} of LLMs \cite{jin2024prollm, kowsher2024propulsion} is equally challenging. Storage requirements scale with the number of model parameters, \( p \), and the size of activations, which is proportional to \( \mathbf{O(n \cdot d_{\text{model}})} \), where \( d_{\text{model}} \) represents the dimensionality of the model. During training, memory usage includes storing parameters and their gradients, leading to a total memory requirement of \( \mathbf{2p} \). This results in substantial memory overhead, especially when updating all parameters during full fine-tuning.

PEFT methods address these computational and memory challenges by modifying only a small subset of parameters. Techniques like adding low-rank matrices or lightweight adapter layers \cite{karimi2021compacter, kowsher2024propulsion, prottasha2024parameter} significantly reduce the number of trainable parameters and the associated memory footprint, enabling faster training and deployment on resource-constrained hardware without compromising performance.

\subsection{Overfitting and Generalization}

Although they have a high capacity, LLMs are prone to \textbf{overfitting}~ \cite{jebali2024leveraging,liu2025mitigating} when fine-tuned on small downstream datasets. Overfitting occurs when a model learns to memorize the training data instead of identifying patterns that generalize to unseen examples. This phenomenon is formally characterized by the inequality:

\[
\textbf{Train Error}(\theta) \ll \textbf{Test Error}(\theta),
\]

where \( \theta \) represents the model parameters. Overfitting becomes especially problematic in low-resource settings, where the lack of sufficient training data limits the model's ability to generalize effectively to new tasks or datasets.

The \textbf{bias-variance trade-off}  \cite{faber1999closer} provides a theoretical framework for understanding the generalization capabilities of LLMs. High-capacity models, such as LLMs, inherently exhibit low bias due to their ability to approximate complex functions. However, this flexibility comes at the cost of high variance, which often leads to overfitting on small datasets. 

PEFT methods address overfitting by updating a small, structured subset of parameters, such as in LoRA and adapters \cite{le2024impact}. This implicit regularization reduces variance without adding significant bias, improving generalization in low-resource settings while ensuring computational and memory efficiency.

\section{PEFT Design}
\label{sec:design}

The expansion of LLMs has presented significant challenges in computational resource allocation, necessitating the development of PEFT techniques  \cite{prottasha2022transfer, clarke2024peft, kowsher2024rocoft}. Unlike full fine-tuning, which requires updating all model parameters, PEFT selectively fine-tunes a subset of parameters, maintaining adaptation effectiveness while reducing computational and memory costs  \cite{hu2021lora, zhang2023adalora}. The efficiency of PEFT methods is dictated by multiple factors, including memory footprint, latency, model sparsity, and energy consumption. This section explores innovative efficiency strategies, starting with precision-aware quantization, dynamic task-adaptive routing, memory-optimized fine-tuning, KV-cache optimization, pruning-based efficiency techniques, energy-aware fine-tuning, and multi-modal PEFT adaptations. These approaches collectively enhance PEFT scalability, enabling cost-effective fine-tuning for diverse AI applications.

\subsection{Precision-Aware Quantization}

Quantization serves as a foundational technique for reducing computational complexity and storage requirements in LLM fine-tuning  \cite{micikevicius2017mixed}. Traditional fine-tuning often relies on high-precision floating-point computations, which lead to increased memory usage and slow inference speeds  \cite{kaplan2020scaling}. In contrast, precision-aware quantization strategically reduces numerical precision in model parameters while preserving task-specific performance. Hybrid bit-width quantization assigns lower precision (e.g., 2-bit or 4-bit) to less critical parameters, while preserving higher precision (e.g., 8-bit or 16-bit) for task-sensitive layers, ensuring optimal trade-offs between efficiency and model accuracy  \cite{he2021towards}. Another promising approach is quantization-aware fine-tuning (QAT), where models undergo low-bit adaptation during the fine-tuning process, preventing performance degradation from post-training quantization methods  \cite{narayanan2021efficient}. By integrating adaptive precision scaling, PEFT frameworks achieve efficient inference performance, making them well-suited for edge and mobile deployments.

\subsection{Dynamic Task-Adaptive Routing}

Traditional PEFT methods operate under static tuning architectures, assuming that all tasks require uniform adaptation. However, task complexity varies significantly, necessitating dynamic routing mechanisms that selectively activate fine-tuned modules based on task-specific demands. Attention-based gating enables models to dynamically engage only the necessary fine-tuned adapters, thereby reducing redundant computations and improving adaptation efficiency  \cite{valipour2022dylora}. In multi-task learning, task-specific pathway optimization ensures that different task-related modules remain distinct, preventing interference between independent fine-tuned representations  \cite{he2021towards}. Additionally, self-supervised routing algorithms can learn optimal activation strategies based on data-driven task profiling, further enhancing PEFT scalability across diverse learning Applications  \cite{valipour2022dylora}.

\subsection{Memory-Optimization}

One of the most significant constraints in fine-tuning LLMs is memory consumption, particularly in resource-limited environments  \cite{narayanan2021efficient}. Standard fine-tuning requires the storage of large-scale activations, optimizer states, and gradients, creating high GPU memory overhead  \cite{micikevicius2017mixed}. To alleviate this burden, memory-efficient PEFT strategies employ techniques such as activation checkpointing, where only critical activations are stored during forward passes, and remaining states are recomputed on demand during backpropagation  \cite{narayanan2021efficient}. Gradient offloading further enhances memory efficiency by storing gradients in secondary memory units, reducing the in-memory footprint required for backpropagation. Additionally, reversible fine-tuning architectures eliminate the need for storing intermediate activation states, instead recomputing them as needed, effectively reducing training memory costs  \cite{micikevicius2017mixed}. By integrating these techniques, PEFT models can be fine-tuned on hardware-constrained environments, including mobile devices and low-power AI accelerators.

\subsection{Key-Value (KV) Cache Optimization }

KV-cache management is a critical factor in transformer-based inference efficiency, particularly in auto-regressive generation models  \cite{narayanan2021efficient}. Each new token generation step requires retrieving and updating previous activations, significantly impacting inference latency and memory consumption  \cite{micikevicius2017mixed, narayanan2021efficient}. Inefficient KV-cache handling can lead to fragmentation, slow retrieval speeds, and unnecessary memory bloat. To address these inefficiencies, hierarchical KV-cache storage introduces tiered caching mechanisms, where frequently accessed activations remain in high-speed memory, while longer-term dependencies are stored in low-priority memory pools  \cite{narayanan2021efficient}. Additionally, entropy-based KV-cache pruning ensures that only high-relevance activations are retained, discarding redundant cache states dynamically  \cite{kaplan2020scaling}. Multi-user PEFT deployments further benefit from adaptive cache allocation strategies, which optimize memory distribution based on workload requirements, enabling high-throughput AI systems to function efficiently across various computational environments  \cite{narayanan2021efficient}.

\subsection{Pruning-Based Efficiency}

Pruning has long been recognized as a powerful tool for reducing model size and computational complexity  \cite{he2016deep, liu2022convnet, zhai2019large}. However, unstructured pruning often results in fragmented weight distributions, making weight merging challenging  \cite{xu2023exploring, hu2022sparse, ding2023sparse}. To overcome this limitation, structured PEFT pruning applies task-aware sparsification, ensuring that critical task-relevant layers remain intact, while low-impact parameters are dynamically removed  \cite{yao2024layer, dong2024data, zhao2024apt}. Layer-wise adapter pruning selectively eliminates adapters from lower transformer layers, focusing computational resources on higher-layer fine-tuned representations  \cite{liu2023moelora}. Channel-wise LoRA pruning further refines efficiency by sparsifying LoRA weight matrices (W\_up and W\_down), reducing unnecessary storage and computation  \cite{zhang2023adalora, gu2024light, wang2024roselora}. Additionally, Neural Architecture Search (NAS)-driven pruning integrates automated reinforcement learning techniques that optimize sparsity patterns dynamically, ensuring optimal parameter reduction with minimal impact on task performance  \cite{lawton2023neural, zhou2024autopeft, hu2022sparse}. By implementing structured, sparsity-aware, and automated pruning methodologies, PEFT frameworks can maintain high adaptation accuracy while significantly reducing computational costs.

\subsection{Energy-Aware Tuning}

With increasing concerns over AI energy consumption, sustainable fine-tuning techniques have become essential for reducing the environmental impact of large-scale LLM training  \cite{narayanan2021efficient, luccioni2023estimating, hoffmann2022training}. Gradient-free optimization introduces an alternative approach where fine-tuning is conducted without backpropagation, significantly reducing power consumption  \cite{he2021towards, duan2024efficient, minaee2024large}. Additionally, early convergence monitoring leverages adaptive loss tracking to terminate training once the model achieves optimal adaptation performance, preventing unnecessary computational cycles  \cite{valipour2022dylora, zhang2023adalora, ding2023sparse}. Another key advancement in energy-aware PEFT is low-power computation graph optimization, which restructures transformer execution pathways to minimize redundant processing operations  \cite{micikevicius2017mixed, wang2022tesseract, huang2019gpipe}. These energy-efficient methodologies not only reduce carbon footprints but also enable AI models to operate on energy-constrained devices, making large-scale adaptation more sustainable.

\subsection{Multi-Modal}

While PEFT techniques have predominantly been applied to text-based language models, recent advances demand multi-modal adaptation capabilities, enabling PEFT to function across vision, speech, and multimodal AI systems  \cite{han2024parameter, xin2024parameter, yang2024cross}. Cross-modal parameter sharing introduces a unified fine-tuning approach, where fine-tuned text-based representations are transferred to vision and speech tasks, minimizing redundant adaptation efforts  \cite{lu2023uniadapter, jin2024mv, diao2024unipt}. Furthermore, token-wise sparsity in multi-modal learning ensures that only the most relevant cross-modal embeddings are retained, significantly improving fine-tuning efficiency for vision-language models (VLMs) and multi-sensory AI frameworks  \cite{zhou2024empirical, yin2024lofit, xu2024let}. By integrating multi-modal fine-tuning strategies, PEFT expands beyond traditional NLP tasks, enabling scalable and efficient adaptation across multiple AI disciplines.

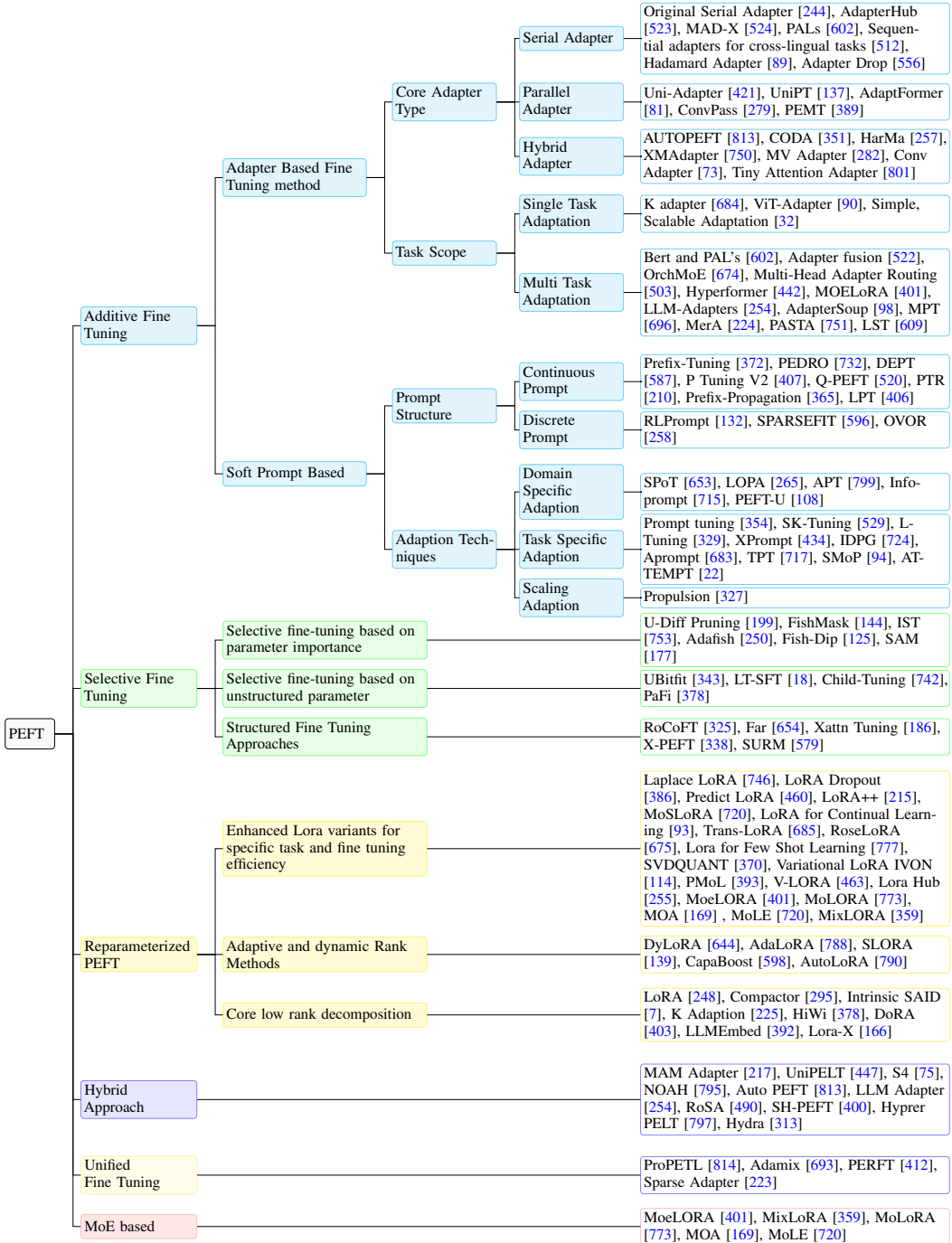
\begin{figure}
    \centering
    \resizebox{\textwidth}{!}{
    \begin{tikzpicture}[
            edge from parent path={
                (\tikzparentnode.east) -- ++(.6cm,0) |- (\tikzchildnode.west)
            },
            every node/.style={
                draw,
                rounded corners=3pt,
                align=left,
                anchor=west
            },
            root/.style={
                text width=1.4cm,
                minimum height=1cm,
                font=\Large
            },
            category/.style={
                text width=3.2cm,
                minimum height=0.8cm,
                font=\Large,
            },
            categoryfirstlevel/.style={
                text width=3.6cm,
                minimum height=0.8cm,
                font=\Large,
            },
            categorysecondlevel/.style={
                text width=4.5cm,
                minimum height=0.8cm,
                font=\Large,
            },
            categorysecondlevelB/.style={
                text width=6.5cm,
                minimum height=0.8cm,
                font=\Large,
            },
            method/.style={
                text width=10cm,
                minimum height=0.6cm,
                font=\Large,
                inner sep=3pt
            },
        ]
        
        \def\selectivefinetune{-2}
        \node[root,fill=gray!5] (peft) at (0,\selectivefinetune-1.5) {PEFT};
        \def\firstlevelblockx{2.5}
        \def\secondlevelblockx{7.2}
        \def\thirdlevelblockx{12.8}
        \def\fourthlevelblockx{17}
        \def\lastblockx{21}
        \def\additiveFT{10}
        \node[categoryfirstlevel,draw=cyan!50, fill=cyan!10] (additive) at (\firstlevelblockx,\additiveFT) {Additive Fine Tuning};
        \node[categoryfirstlevel,draw=green!50, fill=green!10] (selective) at (\firstlevelblockx,\selectivefinetune) {Selective Fine Tuning};
        \def\reparameter{\selectivefinetune-8.8}
        \node[categoryfirstlevel,draw=yellow!70, fill=yellow!20] (reparam) at (\firstlevelblockx,\reparameter) {Reparameterized PEFT};
        \def\hybridy{\reparameter-4.8}
        \node[categoryfirstlevel,draw=blue!50, fill=blue!10] (hybrid) at (\firstlevelblockx,\hybridy) {Hybrid\\ Approach};
        \def\unified{\hybridy-2.5}
        \node[categoryfirstlevel,draw=yellow!50, fill=yellow!10] (unified) at (\firstlevelblockx,\unified) {Unified\\ Fine Tuning};
        \def\moe{\unified-1.7}
        \node[categoryfirstlevel, draw=red!25, fill=red!10] (moe) at (\firstlevelblockx,\moe) {MoE based};
        
        \draw (peft.east) -- ++(0.6cm,0) |- (additive.west);
        \draw (peft.east) -- ++(0.6cm,0) |- (selective.west);
        \draw (peft.east) -- ++(0.6cm,0) |- (reparam.west);
        \draw (peft.east) -- ++(0.6cm,0) |- (hybrid.west);
        \draw (peft.east) -- ++(0.6cm,0) |- (unified.west);
        \draw (peft.east) -- ++(0.6cm,0) |- (moe.west);
        
        \def\adapterbased{\additiveFT+4.9}
        \node[categorysecondlevel, draw=cyan!50, fill=cyan!10] (adapter) at (\secondlevelblockx,\adapterbased) {Adapter Based Fine Tuning method};
        \def\softpromtbased{\additiveFT-4.9}
        \node[categorysecondlevel, draw=cyan!50, fill=cyan!10] (prompt) at (\secondlevelblockx,\softpromtbased) {Soft Prompt Based};
        
        \draw (additive.east) -- ++(0.6cm,0) |- (adapter.west);
        \draw (additive.east) -- ++(0.6cm,0) |- (prompt.west);
        
        \def\coreadapter{\adapterbased+2.5}
        \node[category,draw=cyan!50, fill=cyan!10] (core) at (\thirdlevelblockx,\coreadapter) {Core Adapter Type};
        \def\taskscope{\adapterbased-2.5}
        \node[category,draw=cyan!50, fill=cyan!10] (task) at (\thirdlevelblockx,\taskscope) {Task Scope};
        
        \draw (adapter.east) -- ++(0.6cm,0) |- (core.west);
        \draw (adapter.east) -- ++(0.6cm,0) |- (task.west);
        
        \def\serialadapter{\coreadapter+2.1}
        \node[category,draw=cyan!50, fill=cyan!10] (serial) at (\fourthlevelblockx,\serialadapter) {Serial Adapter};
        \def\paralleladapter{\coreadapter}
        \node[category,draw=cyan!50, fill=cyan!10] (parallel) at (\fourthlevelblockx,\paralleladapter) {Parallel Adapter};
        \def\hybridadapter{\coreadapter-1.8}
        \node[category,draw=cyan!50, fill=cyan!10] (hybrid_adapter) at (\fourthlevelblockx,\hybridadapter) {Hybrid Adapter};
        
        \draw (core.east) -- ++(0.6cm,0) |- (serial.west);
        \draw (core.east) -- ++(0.6cm,0) |- (parallel.west);
        \draw (core.east) -- ++(0.6cm,0) |- (hybrid_adapter.west);
        
        \def\singletaskadapt{\taskscope+1.3}
        \node[category,draw=cyan!50, fill=cyan!10] (single) at (\fourthlevelblockx,\singletaskadapt) {Single Task Adaptation};
        \def\multitaskadapt{\taskscope-1.3}
        \node[category,draw=cyan!50, fill=cyan!10] (multi) at (\fourthlevelblockx,\multitaskadapt) {Multi Task Adaptation};
        
        \draw (task.east) -- ++(0.6cm,0) |- (single.west);
        \draw (task.east) -- ++(0.6cm,0) |- (multi.west);
        
        \def\promptstruc{\softpromtbased+2.25}
        \node[category,draw=cyan!50, fill=cyan!10] (structure) at (\thirdlevelblockx,\promptstruc) {Prompt\\ Structure};
        \def\adaptationtech{\softpromtbased-2.5}
        \node[category,draw=cyan!50, fill=cyan!10] (adaption) at (\thirdlevelblockx,\adaptationtech) {Adaption Techniques};
        
        \draw (prompt.east) -- ++(0.6cm,0) |- (structure.west);
        \draw (prompt.east) -- ++(0.6cm,0) |- (adaption.west);
        
        \def\discreteprompt{\promptstruc-0.8}
        \node[category,draw=cyan!50, fill=cyan!10] (discrete) at (\fourthlevelblockx,\discreteprompt) {Discrete Prompt};
        \def\continuousprompt{\promptstruc+0.8}
        \node[category,draw=cyan!50, fill=cyan!10] (continuous) at (\fourthlevelblockx,\continuousprompt) {Continuous Prompt};
        
        \draw (structure.east) -- ++(0.6cm,0) |- (discrete.west);
        \draw (structure.east) -- ++(0.6cm,0) |- (continuous.west);
        
        \def\domainspecadapt{\adaptationtech+1.9}
        \node[category,draw=cyan!50, fill=cyan!10] (domain) at (\fourthlevelblockx,\domainspecadapt) {Domain\\ Specific\\ Adaption};
        \def\taskspec{\adaptationtech}
        \node[category,draw=cyan!50, fill=cyan!10] (task_spec) at (\fourthlevelblockx,\taskspec) {Task Specific Adaption};
        \def\scalingadapt{\adaptationtech-1.6}
        \node[category,draw=cyan!50, fill=cyan!10] (scaling) at (\fourthlevelblockx,\scalingadapt) {Scaling\\ Adaption};
        
        \draw (adaption.east) -- ++(0.6cm,0) |- (domain.west);
        \draw (adaption.east) -- ++(0.6cm,0) |- (task_spec.west);
        \draw (adaption.east) -- ++(0.6cm,0) |- (scaling.west);
        
        \def\Selectivefinetuningparameterimport{\selectivefinetune+1.6}
        \node[categorysecondlevelB,draw=green!50, fill=green!10] (importance) at (\secondlevelblockx,\Selectivefinetuningparameterimport) {Selective fine-tuning based on\\ parameter importance};
        \def\Selectivefinetuningunstructuredparameter{\selectivefinetune}
        \node[categorysecondlevelB,draw=green!50, fill=green!10] (unstructured) at (\secondlevelblockx,\Selectivefinetuningunstructuredparameter) {Selective fine-tuning based on\\ unstructured parameter};
        \def\StructuredFineTuning{\selectivefinetune-1.6}
        \node[categorysecondlevelB,draw=green!50, fill=green!10] (structured) at (\secondlevelblockx,\StructuredFineTuning) {Structured Fine Tuning\\ Approaches};
        
        \draw (selective.east) -- ++(0.6cm,0) |- (importance.west);
        \draw (selective.east) -- ++(0.6cm,0) |- (unstructured.west);
        \draw (selective.east) -- ++(0.6cm,0) |- (structured.west);
        
        \def\Corelowrank{\reparameter-2}
        \node[categorysecondlevelB,draw=yellow!70, fill=yellow!20] (core_low) at (\secondlevelblockx,\Corelowrank) {Core low rank decomposition};
        \def\Adaptiveanddynamic{\reparameter}
        \node[categorysecondlevelB,draw=yellow!70, fill=yellow!20] (adaptive) at (\secondlevelblockx,\Adaptiveanddynamic) {Adaptive and dynamic Rank Methods};
        \def\enhancedrepara{\reparameter+3.5}
        \node[categorysecondlevelB,draw=yellow!70, fill=yellow!20] (enhanced) at (\secondlevelblockx,\enhancedrepara) {Enhanced Lora variants for specific task and fine tuning efficiency};
        
        \draw (reparam.east) -- ++(0.6cm,0) |- (core_low.west);
        \draw (reparam.east) -- ++(0.6cm,0) |- (adaptive.west);
        \draw (reparam.east) -- ++(0.6cm,0) |- (enhanced.west);
        
        \node[method,draw=blue!50] (hybrid_methods) at (\lastblockx,\hybridy) {MAM Adapter  \cite{he2021towards}, UniPELT  \cite{mao2021unipelt}, S4  \cite{chen2023parameter}, NOAH  \cite{zhang2024neural}, Auto PEFT  \cite{zhou2024autopeft}, LLM Adapter  \cite{hu2023llm}, RoSA  \cite{nikdan2024rosa}, SH-PEFT  \cite{liu2024sparsity}, Hyprer PELT  \cite{zhang2022hyperpelt}, Hydra  \cite{kim2024hydra}};
        
        \draw (hybrid.east) -- ++(0.6cm,0) |- (hybrid_methods.west);
        
        \node[method,draw=blue!50] (unified_methods) at (\lastblockx,\unified) {ProPETL  \cite{zhouprogressive}, Adamix   \cite{wang2022adamix}, PERFT \cite{liu2024perft}, Sparse Adapter \cite{he2022sparseadapter}};
        
        \draw (unified.east) -- ++(0.6cm,0) |- (unified_methods.west);
        
        \node[method,draw=red!25] (moe_methods) at (\lastblockx,\moe) {MoeLORA  \cite{liu2023moelora}, MixLoRA  \cite{li2024mixlora}, MoLoRA  \cite{zadouri2023pushing}, MOA  \cite{feng2024mixture}, MoLE  \cite{wu2024mixture}};
        
        \draw (moe.east) -- ++(0.6cm,0) |- (moe_methods.west);
        
        \node[method,draw=green!50] (importance_methods) at (\lastblockx,\Selectivefinetuningparameterimport) {U-Diff Pruning  \cite{guo2020parameter}, FishMask  \cite{dong2024targeted}, IST  \cite{yao2024layer}, Adafish  \cite{hu2024adafish}, Fish-Dip  \cite{das2023unified}, SAM  \cite{fu2023effectiveness}};
        \node[method,draw=green!50] (unstructured_methods) at (\lastblockx,\Selectivefinetuningunstructuredparameter) {UBitfit  \cite{lawton2023neural}, LT-SFT  \cite{ansell2021composable}, Child-Tuning  \cite{xu2021raise}, PaFi  \cite{liao2023parameter} };
        \node[method,draw=green!50] (structured_methods) at (\lastblockx,\StructuredFineTuning) {RoCoFT  \cite{kowsher2024rocoft}, Far  \cite{vucetic2022efficient}, Xattn Tuning  \cite{gheini2021cross}, X-PEFT  \cite{kwak2024x}, SURM  \cite{sehanobish2024structured}};
        
        \node[method,draw=yellow!70] (core_low_methods) at (\lastblockx,\Corelowrank) {LoRA  \cite{hu2022lora}, Compactor  \cite{karimi2021compacter}, Intrinsic SAID \cite{aghajanyan2020intrinsic}, K Adaption  \cite{he2023parameter}, HiWi  \cite{liao2023parameter}, DoRA  \cite{liu2024dora}, LLMEmbed  \cite{liu2024llmembed}, Lora-X  \cite{farhadzadeh2025lora}};
        \node[method,draw=yellow!70] (adaptive_methods) at (\lastblockx,\Adaptiveanddynamic) {DyLoRA  \cite{valipour2022dylora}, AdaLoRA  \cite{zhang2023adalora}, SLORA  \cite{ding2023sparse}, CapaBoost  \cite{song2024increasing}, AutoLoRA  \cite{zhang2024autolora}};
        \node[method,draw=yellow!70] (enhanced_methods) at (\lastblockx,\enhancedrepara) {Laplace LoRA  \cite{yang2023bayesian}, LoRA Dropout  \cite{lin2024lora}, Predict LoRA  \cite{meng2024periodiclora}, LoRA++  \cite{hayou2024lora+}, MoSLoRA  \cite{wu2024mixture}, LoRA for Continual Learning  \cite{chitale2023task}, Trans-LoRA  \cite{wang2024trans}, RoseLoRA  \cite{wang2024roselora}, Lora for Few Shot Learning  \cite{zanella2024low}, SVDQUANT  \cite{li2024svdqunat}, Variational LoRA IVON  \cite{cong2024variational}, PMoL  \cite{liu2024pmol}, V-LORA  \cite{mi2024v}, Lora Hub  \cite{huang2023lorahub}, MoeLORA  \cite{liu2023moelora}, MoLORA  \cite{zadouri2023pushing}, MOA  \cite{feng2024mixture} , MoLE  \cite{wu2024mixture}, MixLORA  \cite{li2024mixlora}};
        
        \draw (importance.east) -- ++(0.6cm,0) |- (importance_methods.west);
        \draw (unstructured.east) -- ++(0.6cm,0) |- (unstructured_methods.west);
        \draw (structured.east) -- ++(0.6cm,0) |- (structured_methods.west);
        
        \draw (core_low.east) -- ++(0.6cm,0) |- (core_low_methods.west);
        \draw (adaptive.east) -- ++(0.6cm,0) |- (adaptive_methods.west);
        \draw (enhanced.east) -- ++(0.6cm,0) |- (enhanced_methods.west);
        
        \node[method,draw=cyan!50] (serial_methods) at (\lastblockx,\serialadapter) {Original Serial Adapter  \cite{houlsby2019parameter}, AdapterHub  \cite{pfeiffer2020adapterhub}, MAD-X  \cite{pfeiffer2020mad}, PALs  \cite{stickland2019bert}, Sequential adapters for cross-lingual tasks  \cite{parovic2023cross}, Hadamard Adapter  \cite{chen2023hadamard}, Adapter Drop  \cite{ruckle2020adapterdrop}};
        \node[method,draw=cyan!50] (parallel_methods) at (\lastblockx,\paralleladapter) {Uni-Adapter  \cite{lu2023uniadapter}, UniPT  \cite{diao2024unipt}, AdaptFormer  \cite{chen2022adaptformer}, ConvPass  \cite{jie2024convolutional}, PEMT  \cite{lin2024pemt}};
        \node[method,draw=cyan!50] (hybrid_adapter_methods) at (\lastblockx,\hybridadapter) {AUTOPEFT  \cite{zhou2024autopeft}, CODA  \cite{lei2023conditional}, HarMa  \cite{huang2024efficient}, XMAdapter  \cite{yang2024cross}, MV Adapter  \cite{jin2024mv}, Conv Adapter  \cite{chen2024conv}, Tiny Attention Adapter \cite{zhao2022tiny}};
        
        \node[method,draw=cyan!50] (single_methods) at (\lastblockx,\singletaskadapt) {K adapter  \cite{wang2020k}, ViT-Adapter  \cite{chen2022vision}, Simple, Scalable Adaptation  \cite{bapna2019simple}};
        \node[method,draw=cyan!50] (multi_methods) at (\lastblockx,\multitaskadapt) {Bert and PAL’s  \cite{stickland2019bert}, Adapter fusion  \cite{pfeiffer2020adapterfusion}, OrchMoE  \cite{wang2024orchmoe}, Multi-Head Adapter Routing  \cite{page2023multi}, Hyperformer  \cite{mahabadi2021parameter}, MOELoRA  \cite{liu2023moelora}, LLM-Adapters  \cite{hu2023llm}, AdapterSoup  \cite{chronopoulou2023adaptersoup}, MPT  \cite{wang2023multitask}, MerA  \cite{he2023mera}, PASTA \cite{yang2022parameter}, LST  \cite{sung2022lst}};
        
        \node[method,draw=cyan!50] (discrete_methods) at (\lastblockx,\discreteprompt) {RLPrompt  \cite{deng2022rlprompt}, SPARSEFIT  \cite{solano2023sparsefit}, OVOR  \cite{huang2024ovor}};
        \node[method,draw=cyan!50] (continuous_methods) at (\lastblockx,\continuousprompt) {Prefix-Tuning  \cite{li2021prefix}, PEDRO  \cite{xie2024pedro}, DEPT  \cite{shi2023dept}, P Tuning V2  \cite{liu2021p}, Q-PEFT \cite{peng2024q}, PTR  \cite{han2022ptr}, Prefix-Propagation  \cite{li2023prefix}, LPT  \cite{liu2022late}};
        \node[method,draw=cyan!50] (domain_methods) at (\lastblockx,\domainspecadapt) {SPoT  \cite{vu2021spot}, LOPA  \cite{jain2024prompt}, APT  \cite{zhao2024apt}, Infoprompt  \cite{wu2023infoprompt}, PEFT-U  \cite{clarke2024peft}};
        \node[method,draw=cyan!50] (task_spec_methods) at (\lastblockx,\taskspec) {Prompt tuning  \cite{lester2021power}, SK-Tuning  \cite{prottasha2024parameter}, L-Tuning  \cite{kowsher2023tuning}, XPrompt  \cite{ma2022xprompt}, IDPG  \cite{wu2022idpg}, Aprompt  \cite{wang2023aprompt}, TPT  \cite{wu2023parameter}, SMoP  \cite{choi2023smop}, ATTEMPT  \cite{asai2022attempt}};
        \node[method,draw=cyan!50] (scaling_methods) at (\lastblockx,\scalingadapt) {Propulsion \cite{kowsher2024propulsion}};
        
        \draw (serial.east) -- ++(0.6cm,0) |- (serial_methods.west);
        \draw (parallel.east) -- ++(0.6cm,0) |- (parallel_methods.west);
        \draw (hybrid_adapter.east) -- ++(0.6cm,0) |- (hybrid_adapter_methods.west);
        
        \draw (single.east) -- ++(0.6cm,0) |- (single_methods.west);
        \draw (multi.east) -- ++(0.6cm,0) |- (multi_methods.west);
        
        \draw (discrete.east) -- ++(0.6cm,0) |- (discrete_methods.west);
        \draw (continuous.east) -- ++(0.6cm,0) |- (continuous_methods.west);
        
        \draw (domain.east) -- ++(0.6cm,0) |- (domain_methods.west);
        \draw (task_spec.east) -- ++(0.6cm,0) |- (task_spec_methods.west);
        \draw (scaling.east) -- ++(0.6cm,0) |- (scaling_methods.west);
        
        \end{tikzpicture}
    }
    \caption{PEFT Categorized. A comprehensive taxonomy of Parameter-Efficient Fine-Tuning (PEFT) methods. The diagram illustrates the hierarchical organization of PEFT techniques into five major branches: Additive Fine Tuning (with Adapter-Based and Soft Prompt-Based methods), Selective Fine Tuning (parameter-based, unstructured parameter-based, and structured approaches), Reparameterized PEFT (including low-rank decomposition, adaptive rank methods, and Lora variants), Hybrid Approach, and MoE-based methods. Each branch further subdivides into specific implementation strategies and variants. The taxonomy highlights the diverse approaches to achieving parameter efficiency while maintaining model performance across various adaptation scenarios.}
    \label{fig:peft_taxonomy_catagory}
\end{figure}

\section{PEFT Methods}
\label{sec:peftmethods}

With a foundation established in the underlying principles of transfer learning and fine-tuning for large-scale neural networks, we now delve into \textit{PEFT}, a transformative paradigm for adapting LLMs  \cite{han2024parameter}. Traditional full fine-tuning involves updating all parameters $\theta \in \mathbb{R}^p$, where $p$ can scale to billions for modern LLMs  \cite{brown2020language}. While this approach achieves state-of-the-art performance in task-specific scenarios, it poses core difficulties, including high computational costs, substantial storage overhead, and inefficiencies in multi-task learning  \cite{hoffmann2022training}. These issues become particularly pronounced when deploying and maintaining task-specific versions of LLMs at scale  \cite{minaee2024large}. PEFT techniques address these limitations by rethinking the fine-tuning process, focusing on updating only a small subset of parameters or introducing lightweight task-specific modules, thereby significantly reducing computational and memory overhead  \cite{hu2021lora}. The primary Applications of PEFT methods are to reduce trainable parameters, minimize computational demands, and preserve or enhance model performance despite fewer updates. These techniques are particularly well-suited for resource-efficient adaptation of LLMs to new tasks and domains, enabling practical deployment scenarios. 
PEFT strategies can be broadly classified into five distinct categories, each tailored to optimize the fine-tuning process while minimizing computational and memory overhead.

\textbf{Additive Fine-Tuning} enhances the adaptability of pre-trained models by introducing new, trainable modules or parameters into the existing architecture  \cite{houlsby2019parameter}. These modules, such as adapters or low-rank projections, integrate task-specific information without modifying the frozen parameters of the pre-trained model. This approach maintains the original model's generalizability while efficiently encoding task-specific features, making it a resource-effective solution. \textbf{Selective Fine-Tuning} focuses on updating only a subset of the model's parameters, targeting components most relevant to the task at hand  \cite{zaken2021bitfit}. This method significantly reduces the computational requirements of fine-tuning while retaining task-specific effectiveness. Strategies like LoRA (Low-Rank Adaptation) and BitFit selectively adjust specific layers or modules, offering a balance between computational efficiency and performance. \textbf{Reparameterized PEFT} transforms the model parameters into a lower-dimensional representation during training to facilitate efficient optimization  \cite{hu2021lora}. These reparameterized forms are later mapped back to the original parameter space during inference, ensuring the model's full capacity and expressiveness are preserved. Techniques such as tensor decomposition, low-rank matrix factorization, and singular value decomposition exemplify this approach, making it particularly valuable for large-scale models. \textbf{Hybrid Approach} combines elements from multiple PEFT strategies, creating a unified framework that leverages their complementary strengths  \cite{zhang2022hyperpelt}. For example, hybrid methods may integrate additive modules with selective fine-tuning to optimize both modularity and task-specific performance. This approach provides flexibility and adaptability, enabling tailored solutions for complex tasks with varying resource constraints. \textbf{MoE-Based PEFT} (Mixture-of-Experts) leverages sparsely activated architectures where only specific subsets of parameters, or experts, are utilized for a given task  \cite{liu2023moelora}. Dynamic gating mechanisms determine which experts to activate during inference, ensuring task relevance while reducing unnecessary computation. This strategy excels in multi-task and large-scale systems by dynamically allocating resources to achieve efficiency and specialization. Collectively, these strategies present a robust and versatile framework for adapting pre-trained models to diverse tasks, offering significant computational savings while preserving or enhancing task performance. An overview of different PEFT algorithms is summarized below. In Figure~\ref{fig:peft_taxonomy_catagory}, we present a detailed categorization of PEFT techniques.

\begin{figure}[htbp]
\centering
\includegraphics[width=0.99\textwidth]{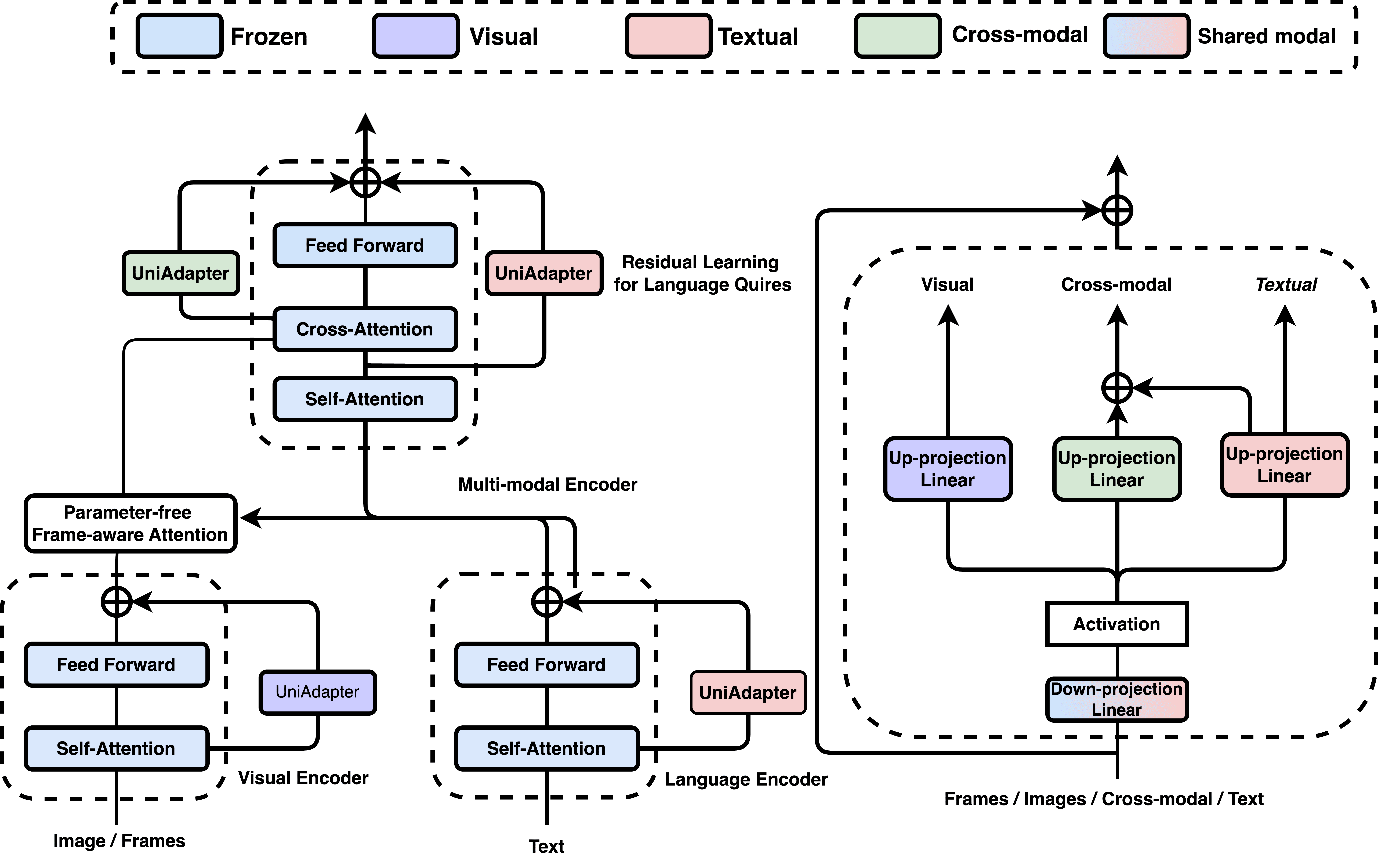}
\caption{Left: Parallel adapter implementation across visual and language encoders with cross-modal connections. Right: Unified adapter structure with modality-specific up-projections feeding into a shared down-projection pathway.}
\label{fig:adapter}
\end{figure}

\subsection{Additive Fine-tuning}

\textbf{Additive fine-tuning} has emerged as a transformative approach in the field of artificial intelligence, offering an efficient and scalable way to customize large-scale pre-trained models for diverse downstream applications. Unlike traditional fine-tuning, which requires extensive updates to all model parameters, \textbf{additive fine-tuning} introduces modular components known as adapters. These adapters provide a lightweight mechanism for integrating task-specific knowledge while preserving the integrity of the frozen parameters of the pre-trained model  \cite{houlsby2019parameter, hu2021lora}. By significantly reducing computational demands and memory requirements, this approach has become a cornerstone in the development of adaptable and flexible models. Additive fine-tuning encompasses three primary architectures—\textbf{serial adapters}  \cite{houlsby2019parameter, pfeiffer2020adapterhub}, \textbf{parallel adapters}  \cite{he2021towards}, and \textbf{hybrid adapters}  \cite{hu2023llm, kim2024hydra} —each designed to address distinct computational and application-specific challenges. Furthermore, these architectures are applied across two major adaptation task scopes: \textbf{single-task adaptation}  \cite{hu2021lora, li2021prefix} and \textbf{multi-task adaptation}  \cite{pfeiffer2020adapterfusion, huang2023lorahub}, broadening their applicability to a variety of practical contexts. In Figure~\ref{fig:adapter}, we illustrate the parallel adapter and unified adapter structures.

\begin{figure}[htbp]
\centering
\includegraphics[width=0.95\textwidth]{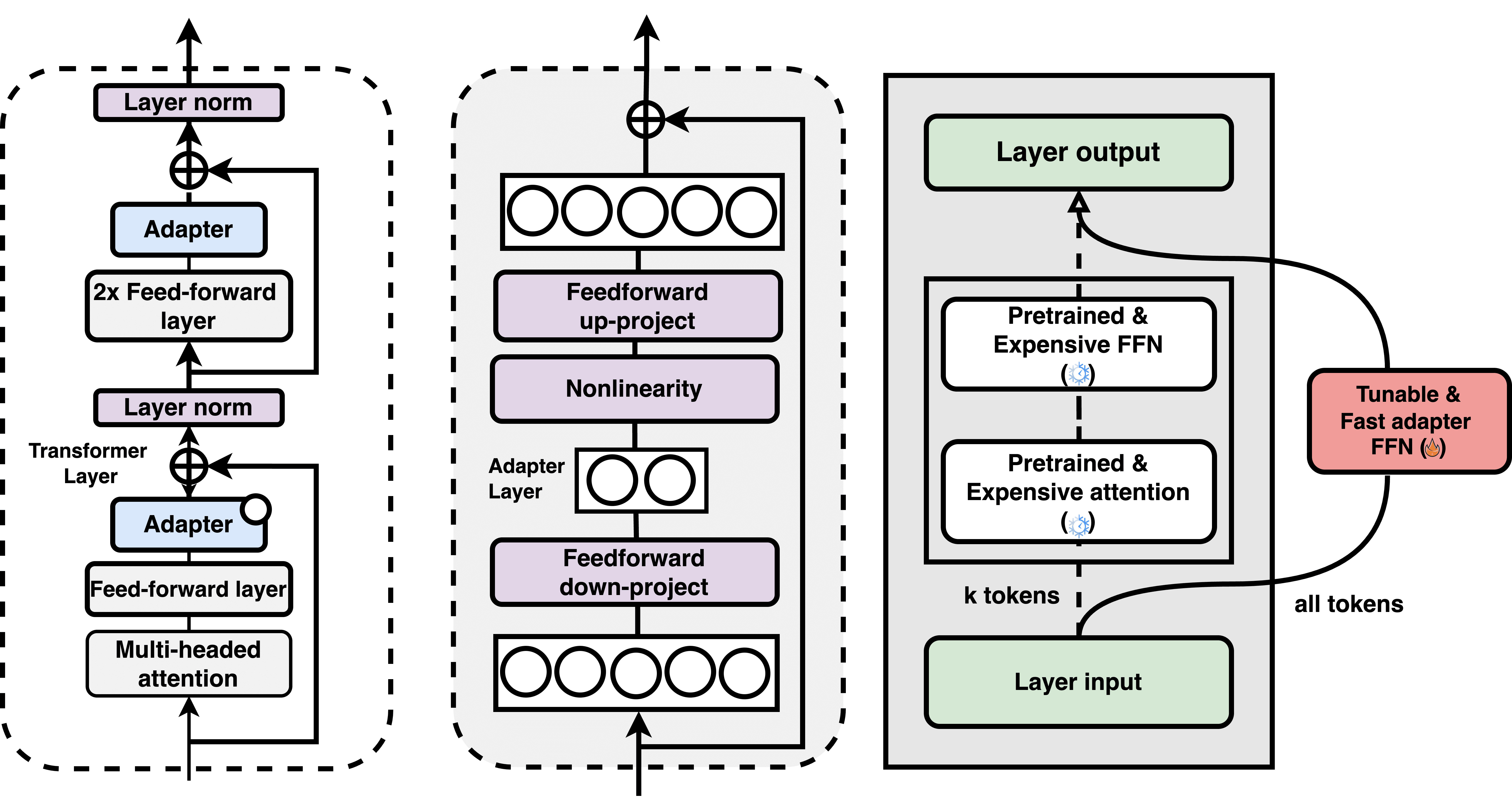}
\caption{Comparison of serial adapter integration in Transformer architecture (left) and adapter layer structure (middle) and Hybrid adapter architecture(right)}.
\label{fig:serial_hyb}
\end{figure}

\subsubsection{Serial adapters} Serial adapters represent one of the earliest and most straightforward approaches to additive fine-tuning. These adapters are integrated sequentially into each layer of the pre-trained model, transforming intermediate representations to embed task-specific features while leaving the original model parameters unaltered. Architecturally, \textbf{serial adapters} employ a bottleneck structure, including a down-projection layer to reduce dimensionality, a non-linear activation function for task-specific transformations, and an up-projection layer to restore the original dimensionality  \cite{houlsby2019parameter}. Notable implementations include the \textbf{Original Serial Adapter}  \cite{houlsby2019parameter}, \textbf{AdapterHub}  \cite{pfeiffer2020adapterhub}, \textbf{MAD-X}  \cite{pfeiffer2020mad}, and \textbf{PALs}  \cite{stickland2019bert}. For instance, \textbf{AdapterHub}  \cite{pfeiffer2020adapterhub} provides a modular framework that facilitates the deployment and reuse of adapters across a variety of domains, enhancing both scalability and adaptability. Meanwhile, \textbf{MAD-X (Modular Adapter Exchange)}  \cite{pfeiffer2020mad} extends the capabilities of \textbf{serial adapters}  \cite{houlsby2019parameter} to multilingual and cross-lingual tasks by integrating task-specific and language-specific adapters. Mathematically, the transformations in a \textbf{serial adapter}  \cite{houlsby2019parameter} are described as follows:
\[
h_{\text{down}} = \text{ReLU}(W_{\text{down}} h_{\text{in}} + b_{\text{down}})
\]
\[
h_{\text{up}} = W_{\text{up}} h_{\text{down}} + b_{\text{up}}
\]
where \( W_{\text{down}} \) and \( W_{\text{up}} \) are projection matrices, and \( b_{\text{down}} \), \( b_{\text{up}} \) are biases. The final output is computed as:
\[
h_{\text{out}} = h_{\text{in}} + h_{\text{up}}
\]

This residual structure ensures that task-specific features are added without disrupting the foundational knowledge encoded in the pre-trained model. These methods, classified as Additive Tuning, are illustrated in Figure~\ref{fig:serial_hyb} showcasing their sequential integration within model architectures.

\subsubsection{Parallel adapters} \textbf{Parallel adapters} offer an alternative design, introducing task-specific modules that operate concurrently with the primary layers of the pre-trained model  \cite{he2021towards}. Unlike \textbf{serial adapters}  \cite{houlsby2019parameter}, which modify intermediate representations directly, \textbf{parallel adapters}  \cite{he2021towards} process task-specific representations alongside the model's primary computations, reducing interference while maintaining independent pathways for task-specific learning. Examples include \textbf{Uni-Adapter}  \cite{lu2023uniadapter}, \textbf{UniPT}  \cite{diao2024unipt}, \textbf{AdaptFormer}  \cite{chen2022adaptformer}, \textbf{ConvPass}  \cite{chen2024conv}, and \textbf{PEMT}  \cite{lin2024pemt}. For instance, \textbf{AdaptFormer}  \cite{chen2022adaptformer} embeds \textbf{parallel adapters}  \cite{he2021towards} within transformer-based architectures to improve adaptability in multi-task contexts, while \textbf{ConvPass}  \cite{chen2024conv} uses convolutional modules for enhanced performance in vision-oriented tasks. The operation of a \textbf{parallel adapter}  \cite{he2021towards} can be expressed mathematically as:
\[h_{\text{parallel}} = W_{\text{parallel}} h_{\text{in}} + b_{\text{parallel}}
\]
where \( W_{\text{parallel}} \) and \( b_{\text{parallel}} \) represent learnable parameters. The final output combines the primary model’s representation \( h_{\text{main}} \) with the adapter’s output:
\[
h_{\text{out}} = h_{\text{main}} + \alpha h_{\text{parallel}}
\]

with \( \alpha \) is a scaling factor that adjusts the influence of the adapter's contribution. These designs, categorized under Additive Tuning, are illustrated in Figure~\ref{fig:adapter}, highlighting the parallel integration of UniAdapters across visual, textual, and cross-modal pathways.

\subsection{Hybrid adapters} \textbf{Hybrid adapters} synthesize the benefits of both \textbf{serial adapters}  \cite{houlsby2019parameter} and \textbf{parallel adapters}  \cite{he2021towards}, offering a unified framework that balances computational efficiency with adaptability to complex tasks. By combining sequential pathways for feature extraction with parallel modules for task-specific encoding, \textbf{hybrid adapters}  \cite{zhang2022hyperpelt} address scenarios such as multi-modal learning and domain-specific applications. Key implementations include \textbf{AUTOPET}  \cite{zhou2024autopeft}, \textbf{CODA}  \cite{lei2023conditional}, \textbf{HarMa}  \cite{he2023mera}, \textbf{XMAdapter}  \cite{yang2024cross}, \textbf{MV Adapter}  \cite{jin2024mv}, and \textbf{Conv Adapter}  \cite{chen2024conv}. For example, \textbf{XMAdapter}  \cite{yang2024cross} effectively blends \textbf{serial}  \cite{houlsby2019parameter} and \textbf{parallel components}  \cite{he2021towards} to adapt models for vision-language tasks, while \textbf{AUTOPET}  \cite{zhou2024autopeft} dynamically adjusts the architecture based on task complexity, optimizing both performance and resource use. The mathematical formulation for \textbf{hybrid adapters}  \cite{zhang2022hyperpelt} integrates outputs from serial and parallel components:
\[
h_{\text{out}} = \beta h_{\text{serial}} + \gamma h_{\text{parallel}}
\]
where \( \beta \) and \( \gamma \) are coefficients that balance the contributions of the two pathways. These versatile approaches, classified as Additive Tuning, are depicted in Figure~\ref{fig:serial_hyb}, illustrating their capability to handle diverse and complex tasks.

Beyond the core architectures, additive fine-tuning is applied to two primary adaptation scenarios from Task Scope: \textbf{single-task adaptation}  \cite{hu2021lora} and \textbf{multi-task adaptation}  \cite{pfeiffer2020adapterfusion}.

\subsubsection{Single-task adaptation}
\begin{figure}[htbp]
\centering
\includegraphics[width=0.85\textwidth]{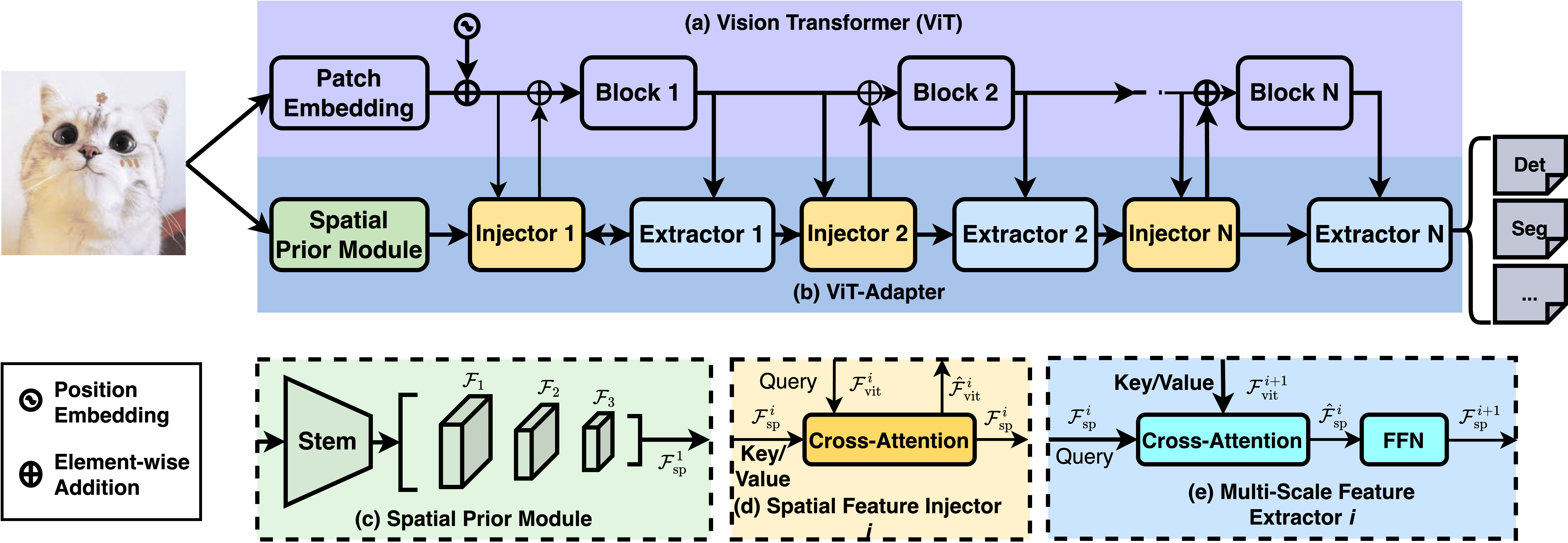}
\caption{Vision Transformer (ViT) with adapter modules: (a) Standard ViT architecture, (b) ViT-Adapter framework with injector-extractor modules, (c) Spatial Prior Module, (d) Spatial Feature Injector with cross-attention, and (e) Multi-Scale Feature Extractor. The design supports single-task applications including detection and segmentation.}
\label{fig:single_task_adaption}
\end{figure}

\textbf{Single-task adaptation} focuses on fine-tuning models for specific applications by employing highly tailored adapters  \cite{hu2021lora}. Examples include the \textbf{K-Adapter}  \cite{wang2020k}, \textbf{ViT-Adapter}  \cite{chen2022vision}, and methods for neural machine translation. The \textbf{K-Adapter}  \cite{wang2020k} integrates external knowledge into pre-trained systems, enabling them to excel in knowledge-intensive tasks, while the \textbf{ViT-Adapter}  \cite{chen2022vision} adapts Vision Transformers for visual tasks such as object detection and segmentation. This approach incorporates spatial prior modules, feature injectors, and extractors to embed task-specific knowledge. Mathematically, the integration of external features \( F_{\text{sp}} \) into a layer representation \( F_i \) is performed via cross-attention:
\[
\hat{F}_i = \text{CrossAttention}(F_i, F_{\text{sp}})
\]
These methods, categorized under Additive Tuning, are illustrated in Figure~\ref{fig:single_task_adaption}, showcasing the integration of spatial prior modules, feature injectors, and extractors for single-task adaptation.

\subsubsection{Multi-task adaptation} \textbf{Multi-task adaptation}, on the other hand, enables a single model to handle multiple applications simultaneously by maintaining task-specific representations while leveraging shared pre-trained parameters  \cite{pfeiffer2020adapterfusion}. Notable implementations include \textbf{AdapterFusion}  \cite{pfeiffer2020adapterfusion}, \textbf{Hyperformer}  \cite{mahabadi2021parameter}, \textbf{AdapterSoup}  \cite{chronopoulou2023adaptersoup}, \textbf{OrchMoE}, and \textbf{MOEoRA}  \cite{liu2024moe}. For instance, \textbf{AdapterFusion}  \cite{pfeiffer2020adapterfusion} integrates multiple adapters dynamically to optimize performance across different tasks, while 
\textbf{AdapterSoup}  \cite{chronopoulou2023adaptersoup} aggregates and selects relevant adapters during inference for enhanced task generalization.

\begin{figure}[htbp]
  \centering
  \includegraphics[width=0.85\textwidth]{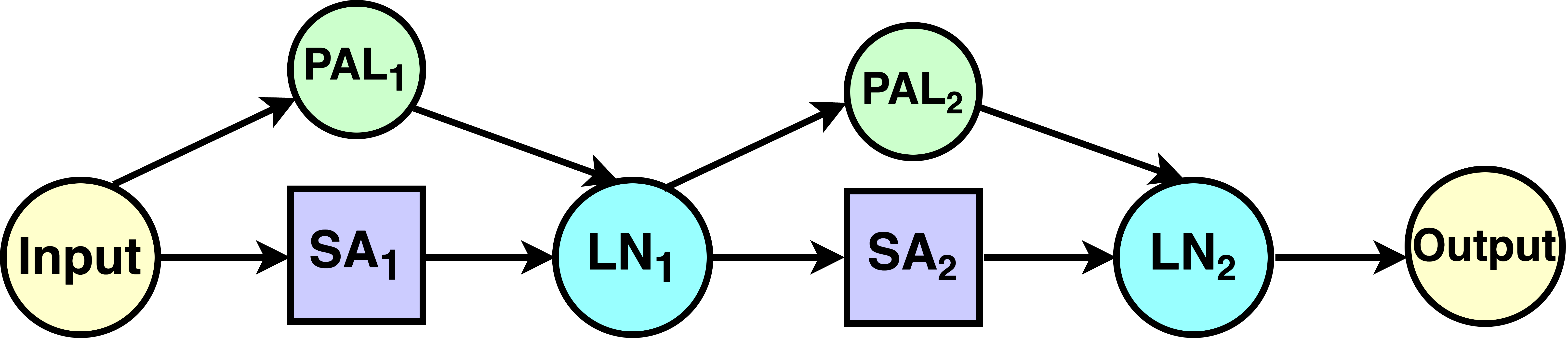}
  \caption{An illustration of a multi-task adaptation architecture integrating Parallel Adapter Layers (PALs). The input passes through a sequence of self-attention (SA) and layer normalization (LN) blocks, while task-specific PAL modules inject parallel residual pathways into the backbone model. The PALs at each layer facilitate task-specific adaptation while maintaining the shared structure.}
  \label{fig:multi_task_adaption}
\end{figure}

Mathematically, for a specific task \( t \), the adapter processes the input as:
\[
h_{\text{task-specific}} = W_t h_{\text{in}} + b_t
\]
Fusion mechanisms, such as those employed in AdapterSoup, dynamically combine the outputs of multiple adapters, ensuring effective performance across diverse tasks. These strategies, classified under Additive Tuning, are depicted in Figure~\ref{fig:multi_task_adaption}, showcasing their versatility in multi-task environments.

\subsection{Soft Prompt PEFT} 
\begin{figure}[htbp]
  \centering
  \includegraphics[width=1\textwidth]{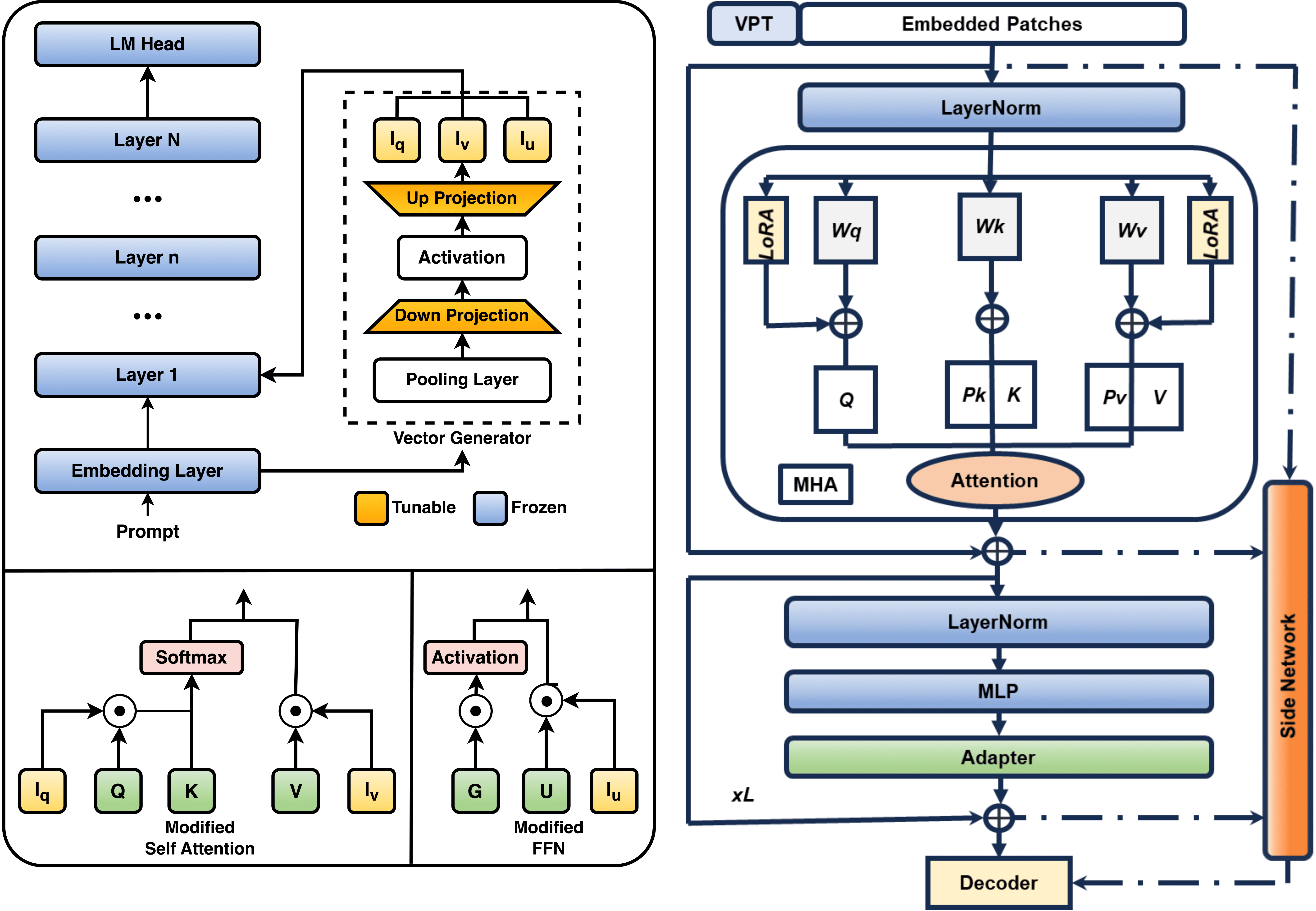}
  \caption{Prompt-based parameter-efficient fine-tuning architecture. (Left) Top: Model structure with frozen LM layers and tunable Vector Generator. Bottom: Modified Self-Attention and FFN mechanisms showing interaction between generated vectors (yellow) and existing components (green). (Right) Architecture of a RL Prompt}
  \label{fig:rl_prompt}
\end{figure}

\textbf{Soft prompt-based} fine-tuning has revolutionized how pre-trained models are adapted for specific tasks  \cite{lester2021power}. By leveraging lightweight prompts—either learnable or fixed—it offers a modular and efficient alternative to traditional fine-tuning approaches. This framework revolves around two fundamental components: \textbf{prompt structures}  \cite{liu2021p}  and \textbf{adaptation Techniques}  \cite{li2021prefix}, both supported by precise mathematical formulations. Each section below includes correctly represented mathematical expressions for a deeper understanding. At the core of this methodology are the \textbf{prompt structures}, which determine how prompts are represented and incorporated into the model architecture  \cite{lester2021power}. These structures are divided into two principal categories: \textbf{continuous prompts}  \cite{li2021prefix} and \textbf{discrete prompts}  \cite{han2022ptr}. \textbf{Continuous prompts} are learnable embeddings, optimized during the fine-tuning process to capture intricate task-specific patterns. Techniques such as \textbf{Prefix-Tuning}  \cite{li2021prefix}, \textbf{P-TUNING v2}  \cite{liu2021p}, \textbf{Q-PEFT}  \cite{peng2024q}, \textbf{PTR}  \cite{han2022ptr}, \textbf{Prefix-Propagation}  \cite{li2023prefix}, and \textbf{LPT}  \cite{liu2022late} showcase the adaptability of this approach. For instance, \textbf{Prefix-Tuning} appends trainable embeddings, referred to as prefixes, to the input sequence, augmenting the attention mechanism to emphasize task-relevant features. This adjustment is mathematically expressed as:
\begin{equation}
\text{Attention}(Q, [P; K], [P; V]) = \text{softmax}\left(\frac{Q[P; K]^T}{\sqrt{d}}\right)[P; V],
\end{equation}
where $P$ represents the prefix, and $Q, K, V$ are the query, key, and value matrices, respectively. Expanding on this, \textbf{P-TUNING v2}  \cite{liu2021p} incorporates continuous prompts into multiple transformer layers, enabling deeper task-specific generalization. Other advancements, such as \textbf{Q-PEFT}  \cite{peng2024q}, employ quantized embeddings to enhance memory efficiency, while \textbf{PTR}  \cite{han2022ptr} facilitates the transfer of learned prompts across related domains. \textbf{Figure~\ref{fig:rl_prompt}} (Left) illustrates how continuous prompts are integrated into transformer architectures, dynamically adjusting the model's focus on task-critical elements.

Discrete prompts, by contrast, utilize fixed, tokenized sequences—such as phrases or pre-defined linguistic structures—that provide explicit guidance to the model. These prompts are static and do not involve learnable embeddings. Techniques such as \textbf{RLPrompt}  \cite{deng2022rlprompt}, \textbf{SPARSETT}  \cite{solano2023sparsefit}, and \textbf{OVOR}  \cite{huang2024ovor} demonstrate the utility of discrete prompting. For example, \textbf{RLPrompt} (Figure \ref{fig:rl_prompt} (Right) optimizes tokenized prompts through reinforcement learning to maximize task performance. The optimization process is mathematically represented as:

\begin{equation}
\theta = \theta + \eta \nabla_\theta \mathbb{E}[R(P)],
\end{equation}
where $\theta$ denotes the parameters of the prompt policy, $\eta$ is the learning rate, and $R(P)$ represents the reward function tailored to task-specific Applications. Similarly, \textbf{SPARSETT}  \cite{solano2023sparsefit} employs sparse optimization techniques to retain only the most relevant tokens, ensuring computational efficiency. This refinement process is expressed as:
\begin{equation}
P_{\text{sparse}} = \underset{P}{\text{argmin}} \|P\|_0 \quad \text{subject to } R(P) \geq R_{\text{threshold}}.
\end{equation}
The role of discrete prompts is visually depicted in \textbf{Figure \ref{fig:rl_prompt} (Right)}, highlighting their integration into the input sequence to guide model outputs effectively. Building on these foundational structures, \textbf{adaptation strategies} refine how prompts are used to tailor pre-trained models for specific applications. \textbf{Task-specific adaptation} focuses on tailoring prompts for individual tasks to achieve high accuracy and efficiency. Techniques such as \textbf{Prompt Tuning}  \cite{ma2022xprompt}, \textbf{SK-Tuning}  \cite{prottasha2024parameter}, \textbf{L-Tuning}  \cite{kowsher2023tuning}, \textbf{XPrompt}  \cite{ma2022xprompt}, \textbf{IDPG}  \cite{wu2022idpg}, \textbf{Arprompt} \cite{wang2023aprompt}, \textbf{TPT} \cite{wu2023parameter}, and \textbf{SMoP} \cite{choi2023smop} fall into this category. For instance, \textbf{Prompt Tuning} introduces task-specific embeddings to the input sequence, enabling precise modulation of the model's outputs. This can be mathematically expressed as:
\begin{equation}
y = f(x, P_{\text{task-specific}}),
\end{equation}
with $x$ is the input, and $P_{\text{task-specific}}$ represents the learned prompt for the specific task. Iterative approaches like \textbf{IDPG (Iterative Dual Prompt Generation)}  \cite{wu2022idpg} refine prompts iteratively over multiple steps, expressed as:
\begin{equation}
P^{(t+1)} = P^{(t)} - \alpha \nabla_P \mathcal{L}(f(x, P^{(t)})),
\end{equation}
where $t$ represents the iteration step, and $\alpha$ is the learning rate.

\subsection{Scaling PEFT}  
\begin{figure}[htbp]
\centering
\includegraphics[width=0.9\textwidth]{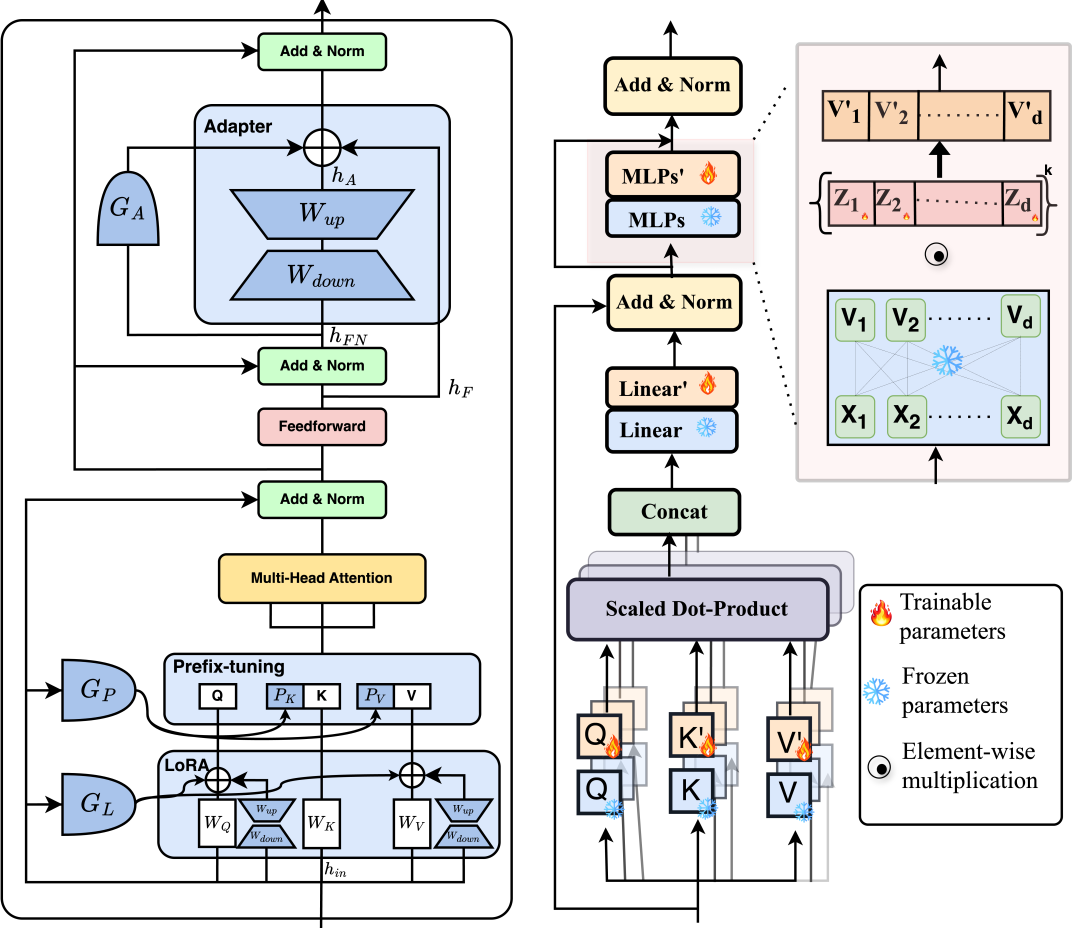}
\caption{Architecture of a unified hybrid PEFT approach integrating multiple parameter-efficient fine-tuning strategies within a transformer block (left). Propulsion architecture showing trainable and frozen components. The design features scaled dot-product attention with selectively tunable parameters (orange) and frozen weights (blue), enabling efficient fine-tuning through element-wise multiplication of scaled vectors (right).}\label{fig:propulsion}
\end{figure}

\textbf{Scaling PEFT} extends the utility of prompts to handle broader and more complex contexts. A notable advancement in this area is the \textbf{Propulsion concept}  \cite{kowsher2024propulsion}, which incorporates polynomial scaling to dynamically adjust the influence of prompt parameters. This mechanism allows for granular control over the model’s sensitivity to input features and is mathematically defined as:
\begin{equation}
V'_i = [v_1 \odot z_1^k, v_2 \odot z_2^k, \dots, v_s \odot z_s^k],
\end{equation}
where $v_i$ represents the input features, $z_i$ are scaling parameters, $k$ is the scaling exponent, and $\odot$ denotes element-wise multiplication. The Propulsion method's architecture is illustrated in \textbf{Figure \ref{fig:propulsion}} (right), showing its attention mechanism modification approach with selective parameter tuning.

\subsection{Selective fine-tuning} \textbf{Selective fine-tuning} is an advanced model optimization technique designed to adapt pre-trained models to specific tasks by modifying only a carefully chosen subset of parameters while keeping the rest unchanged  \cite{kowsher2024rocoft, zaken2021bitfit, hu2024adafish}. Unlike traditional fine-tuning, which updates all parameters in the model, selective fine-tuning focuses on parameters that are most relevant to the task. This targeted approach reduces computational costs, mitigates overfitting, and preserves the general-purpose knowledge embedded in the pre-trained model. By relying on principles such as parameter importance, sparsity, and structural organization, selective fine-tuning achieves a
balance between efficiency and adaptability, making it an invaluable tool in modern machine learning.

The parameters of a pre-trained model, denoted as \( \theta \), are divided into two subsets: \( \theta_s \), the parameters selected for fine-tuning, and \( \theta_f \), the parameters that remain fixed. The selection of \( \theta_s \) is guided by a criterion \( C(\cdot) \), which evaluates the relevance of each parameter to the task. Parameters with relevance scores exceeding a threshold \( \tau \) are included in \( \theta_s \), while the rest are assigned to \( \theta_f \). Mathematically, this can be expressed as:
\[
\theta_s = \{ \theta_i \, | \, C(\theta_i) \geq \tau \}, \quad \theta_f = \theta \setminus \theta_s.
\]
The optimization process then focuses on \( \theta_s \), while \( \theta_f \) remains unchanged to preserve the pre-trained knowledge. This process is formalized as:
\[
\underset{\theta_s}{\text{argmin}} \, \mathcal{L}(f(x; \theta_s \cup \theta_f), y),
\]
where \( f(x; \theta) \) represents the model's output for input \( x \), and \( \mathcal{L} \) is the task-specific loss function. Selective fine-tuning often identifies important parameters based on their contribution to the task. Techniques such as Fisher information, gradient magnitudes, or sensitivity analysis are commonly used to measure parameter importance. Methods like \textbf{FishMask}  \cite{dong2024data}, \textbf{Adafish}  \cite{hu2024adafish}, \textbf{IST}  \cite{yao2024layer}, and \textbf{ U-Diff Pruning }  \cite{guo2020parameter} exemplify this approach.

\begin{figure}[t!]
\centering
\includegraphics[width=0.56\textwidth]{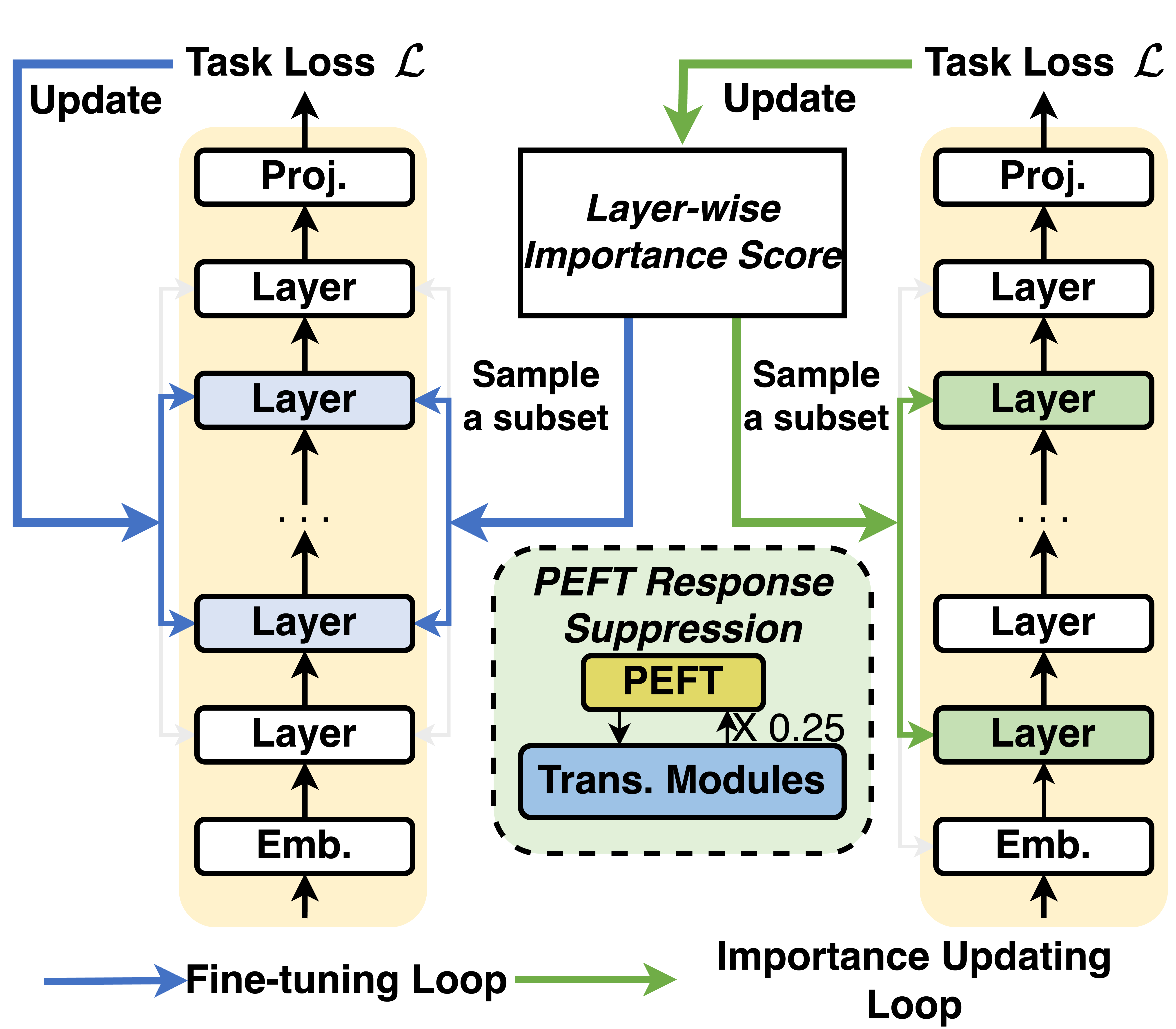}
\caption{Parameter importance selection framework with dual optimization loops. Left: Fine-tuning loop updating model parameters based on task loss. Center: Layer-wise importance scoring mechanism with PEFT response suppression. Right: Importance updating loop that samples subset layers for targeted optimization based on their calculated importance scores.}
\label{fig:ist}
\end{figure}

For example, \textbf{FishMask}  \cite{dong2024data} uses Fisher information to evaluate the importance of each parameter. The Fisher information for parameter \( \theta_i \) is defined as:
\[
\mathcal{I}(\theta_i) = \mathbb{E} \left[ \left( \frac{\partial \log p(x; \theta)}{\partial \theta_i} \right)^2 \right],
\]
where \( \mathcal{I}(\theta_i) \) quantifies the influence of \( \theta_i \) on the model's predictions, and \( \log p(x; \theta) \) is the log-likelihood. Parameters with high Fisher information are included in \( \theta_s \), as they have a significant impact on task performance. Similarly, \textbf{Adafish}  \cite{hu2024adafish} dynamically adjusts the selection criteria during training by analyzing gradient magnitudes, allowing the model to focus on parameters that are most relevant to the evolving task. As illustrated in \textbf{Figure~\ref{fig:ist}}, this approach utilizes a dual-loop architecture to iteratively identify and select important parameters based on Fisher information, enabling more efficient task-specific adaptation.

Unstructured parameter selection focuses on selecting individual parameters independently of their groupings or positions within the model. This approach is commonly employed in sparsity-based techniques, such as \textbf{UBitFit}  \cite{lawton2023neural}, \textbf{LT-SFT}  \cite{ansell2021composable}, \textbf{Child-Tuning}  \cite{xu2021raise}, and \textbf{PaFi}  \cite{liao2023parameter}.

\begin{figure}[htbp]
   \centering
   \includegraphics[width=0.76\textwidth]{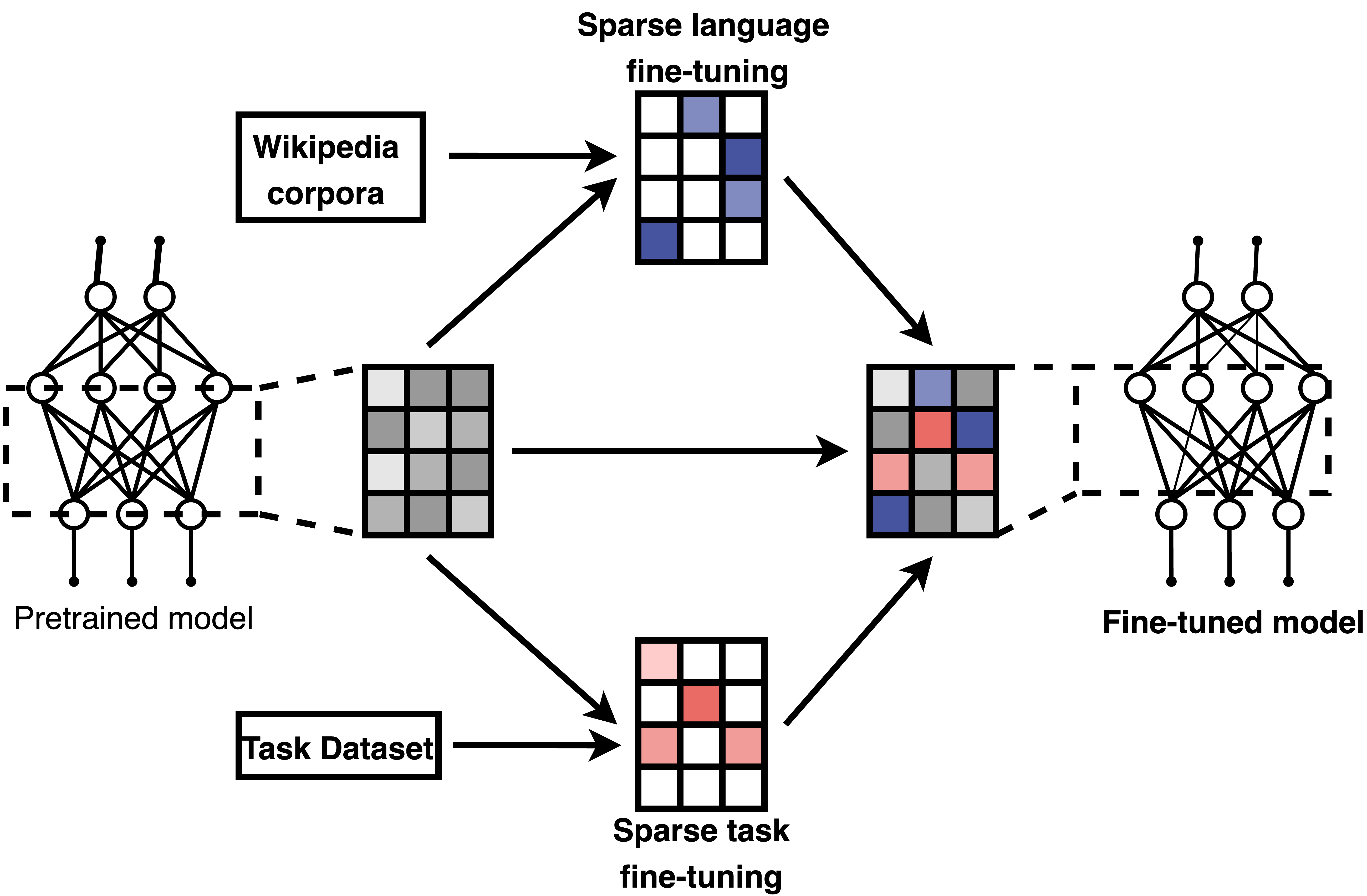}
   \caption{Sparse Fine-tuning Framework. This diagram illustrates the process of sparse parameter fine-tuning for language models. Starting with a pretrained model (left), the framework utilizes both Wikipedia corpora for sparse language fine-tuning (top path) and task-specific datasets for sparse task fine-tuning (bottom path). The selective parameter updates, represented by colored cells in the matrices, allow the fine-tuned model (right) to maintain general capabilities while adapting to specific tasks with minimal parameter changes. The blue cells represent language-related parameters, while red cells indicate task-specific parameters selected for updating.}
   \label{fig:LT_SFT}
\end{figure}

In \textbf{Child-Tuning}, for instance, the relevance of each parameter is determined using gradient norms:
\[
C(\theta_i) = \left\| \frac{\partial \mathcal{L}}{\partial \theta_i} \right\|,
\]
where $C(\theta_i)$ is the importance score for parameter $\theta_i$, and parameters with scores above a threshold $\tau$ are included in $\theta_s$. This approach ensures that only the parameters that contribute significantly to the task are updated, enhancing both efficiency and performance. \textbf{Figure~\ref{fig:LT_SFT}} illustrates the unstructured parameter selection process, showing how individual parameters are identified and updated to achieve task-specific optimization without disrupting the overall structure of the model.

Structured fine-tuning focuses on updating coherent groups of parameters, such as layers, attention heads, or blocks, rather than individual parameters. This approach is particularly effective for modular architectures like transformers, where parameters are hierarchically organized. Methods such as \textbf{RoCoFT}  \cite{kowsher2024rocoft}, \textbf{Far}  \cite{vucetic2022efficient}, \textbf{Xattn Tuning}  \cite{gheini2021cross}, \textbf{X-PEFT with Hard Masking}  \cite{kwak2024x}, and \textbf{SURM}  \cite{sehanobish2024structured} adopt this strategy.

\begin{figure}[htbp]
  \centering
  \includegraphics[width=0.70\textwidth]{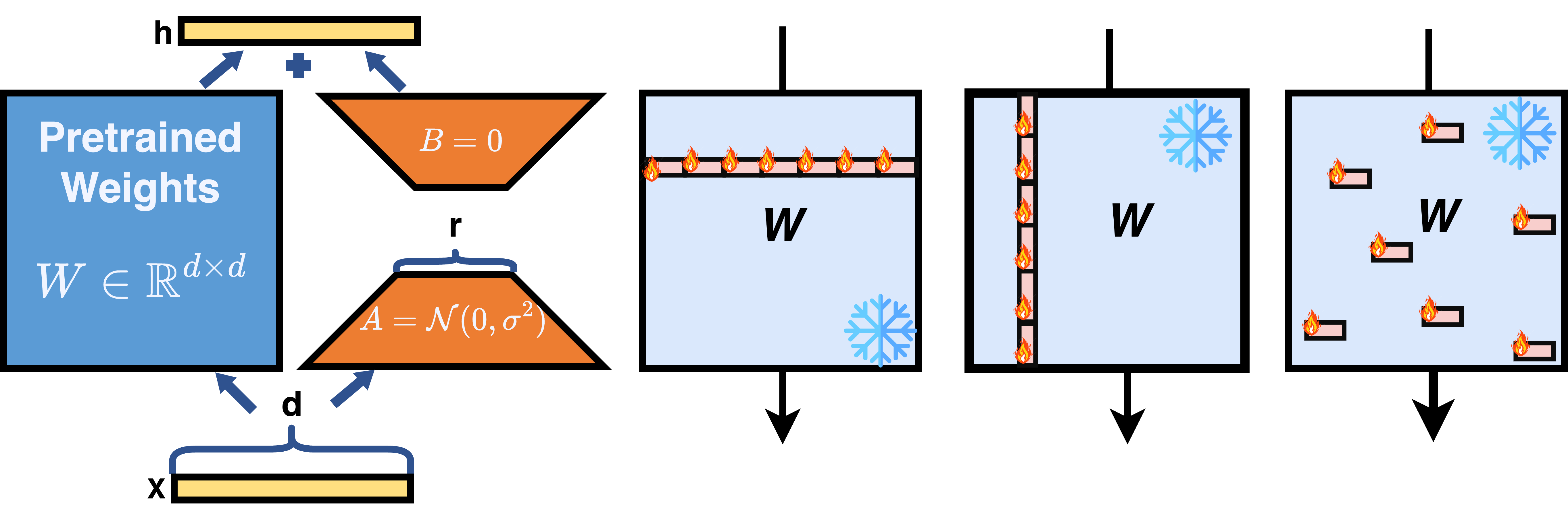}
  \caption{Visualization of LoRA and RoCoFT architectures. The first figure (left) illustrates the architecture of LoRA. The remaining figures depict the RoCoFT variants, including row-wise updates, column-wise updates, and random updates.}
  \label{fig:rocoft_lora_patterns}
\end{figure}

In \textbf{X-PEFT with Hard Masking}, a binary mask \( M \) is applied to enforce structural constraints during fine-tuning:
\[
\theta_s = M \odot \theta,
\]
where \( M \in \{0, 1\}^d \) represents the mask, \( \odot \) denotes element-wise multiplication, and \( d \) is the number of parameters. This ensures that only critical components, such as specific layers or blocks, are updated, while the remaining parameters remain fixed. Similarly, \textbf{SURM}  \cite{sehanobish2024structured} applies domain-specific masking strategies to align fine-tuning with the structural requirements of the task. \textbf{Figure \ref{fig:rocoft_lora_patterns} (right)} illustrates different parameter update patterns in the \textbf{RoCoFT}  \cite{kowsher2024rocoft} approach, showing how selective modification of rows or columns within weight matrices enables efficient fine-tuning while preserving model coherence.

\subsection{Reparameterized PEFT} \textbf{Reparameterized PEFT} methods aim to optimize large-scale pre-trained models by introducing efficient, low-rank transformations that reduce trainable parameters while preserving task-specific performance  \cite{hu2021lora, liu2024dora, zhang2023adalora, gao2024dlora}. These methods can be broadly categorized into three groups: core low-rank decomposition techniques, adaptive and dynamic rank methods, and enhanced LoRA variants for specific tasks and fine-tuning efficiency. This paper provides a detailed overview of these methods and their applications in large-scale machine learning models. Reparameterized PEFT addresses the computational and memory constraints of fine-tuning large-scale models by introducing low-rank parameterization techniques. These approaches focus on reparameterizing the delta weight matrix ($\Delta W$) into a low-dimensional form, significantly reducing the number of trainable parameters. The techniques can be classified into three main categories: \textbf{core low-rank decomposition}, \textbf{adaptive and dynamic rank methods}, and \textbf{enhanced LoRA variants} tailored for specific tasks.
\subsubsection{Low-Rank Decomposition}
The foundation of reparameterized PEFT lies in \textbf{low-rank decomposition}, where the parameter update matrix $\Delta W \in \mathbb{R}^{d \times d}$ is approximated as the product of two low-rank matrices, $A \in \mathbb{R}^{d \times r}$ and $B \in \mathbb{R}^{r \times d}$, with $r \ll d$. Mathematically, this can be expressed as:
\begin{equation}
    \Delta W \approx AB
\end{equation}

This decomposition reduces the number of trainable parameters from $d^2$ to $2dr$, significantly lowering computational requirements. Methods such as \textbf{LoRA}  \cite{hu2021lora}, \textbf{Compactor}  \cite{karimi2021compacter}, \textbf{Intrinsic SAID}  \cite{aghajanyan2020intrinsic}, \textbf{K Adaption}  \cite{he2023parameter}, \textbf{DoRA}  \cite{liu2024dora}, and \textbf{LLEmbed}  \cite{liu2024llmembed} build on this concept. For instance, LoRA constrains the updates to a low-rank subspace, while Compactor introduces sparsity-inducing priors for further efficiency. Intrinsic SAID optimizes updates using intrinsic dimensionality principles, and K Adaption dynamically tunes the rank $r$ to align with task-specific requirements, enhancing flexibility.  
\textbf{Figure \ref{fig:rocoft_lora_patterns}} illustrates the low-rank adaptation approach used in LoRA, demonstrating how the pretrained weight matrix $W \in \mathbb{R}^{d \times d}$ is combined with low-rank matrices $A$ and $B$ to form the final weight matrix $h = W\cdot x + BA \cdot x$, where $B$ is initialized to zero and $A$ is sampled from a Gaussian distribution.

This approach is particularly effective for large language models where the full fine-tuning of all parameters would be prohibitively expensive. By focusing training exclusively on the low-rank matrices $A$ and $B$, LoRA achieves comparable performance to full fine-tuning while requiring only a fraction of the computational resources and storage requirements. The rank $r$ serves as a hyperparameter that controls the trade-off between model capacity and training efficiency.

\subsubsection{Dynamic Rank Methods}
\begin{figure}[htbp]
  \centering
  \includegraphics[width=0.99999\textwidth]{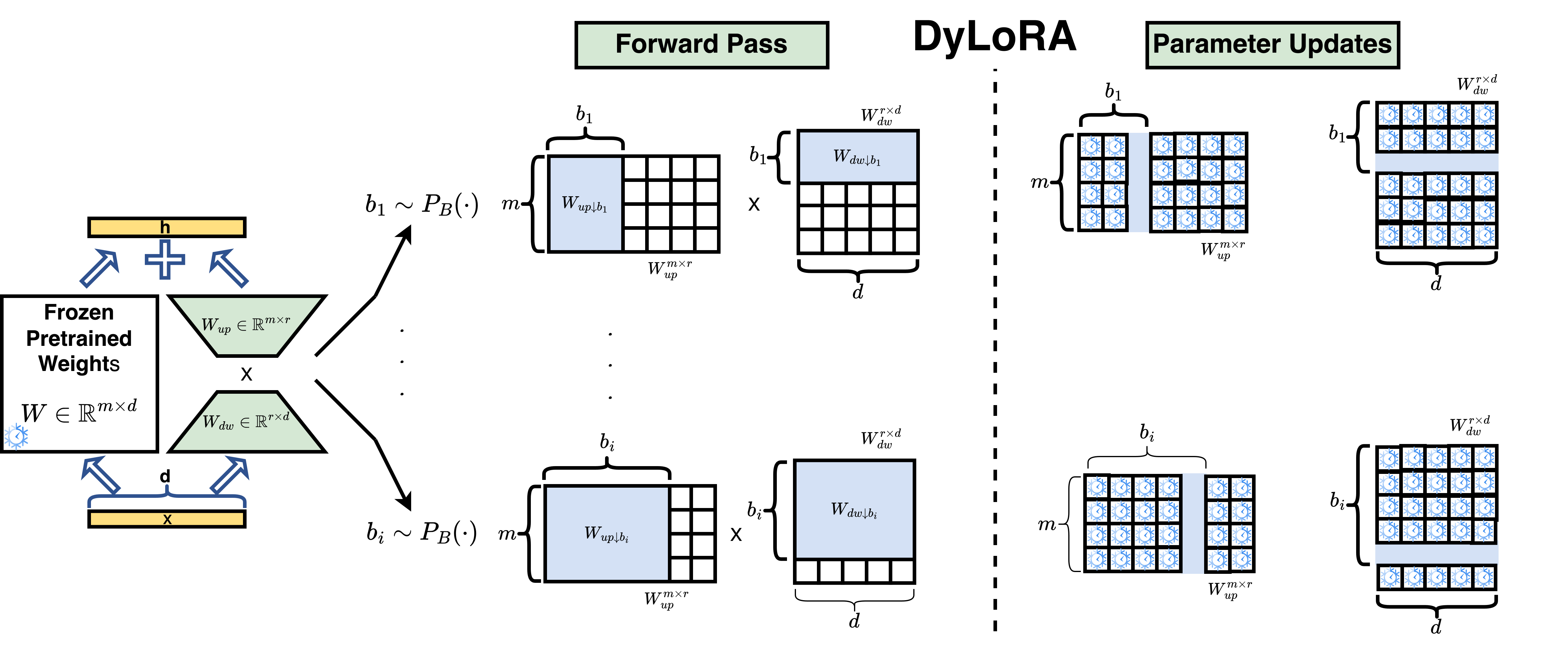}
  \caption{Visualization of DyLoRA (Dynamic Low-Rank Adaptation), which enhances standard LoRA by introducing dynamically sampled low-rank blocks. Left: Frozen pretrained weight matrix $W \in \mathbb{R}^{m \times d}$ with blocks $W_{qp}$ and $W_{kv}$. Center: Forward pass showing how blocks $b_i \sim P_B(\cdot)$ are sampled and multiplied with corresponding weight matrices. Right: Parameter updates with dynamic allocation across matrix blocks, enabling more efficient fine-tuning by focusing updates where they provide the greatest benefit.}
  \label{fig:dylora_approach}
\end{figure}

While core low-rank decomposition uses a fixed rank $r$, \textbf{adaptive and dynamic rank methods} adjust the rank during training to optimize performance and resource usage. Techniques such as \textbf{DyLoRA}  \cite{valipour2022dylora} and \textbf{AdaLoRA}  \cite{zhang2023adalora} dynamically modify $r$ based on gradient information or layer sensitivity:
\begin{equation}
    r_t = f(\|\nabla_t\|), \quad t \in \{1, \dots, T\}
\end{equation}
where $f(\cdot)$ is a function of the gradient norm $\|\nabla_t\|$, and $T$ is the number of training steps. Similarly, \textbf{SLORA}  \cite{ding2023sparse} employs layer-wise rank scheduling, while \textbf{CapaBoost}  \cite{song2024increasing} and \textbf{AutoLoRA}  \cite{zhang2024autolora} automate rank selection using task-specific metrics. These methods introduce adaptability, ensuring efficient resource allocation and improved task performance.  
\textbf{Figure \ref{fig:dylora_approach}} illustrates the DyLoRA approach that dynamically adjusts low-rank updates through block-wise decomposition patterns, showcasing how parameter updates propagate through the model architecture while maintaining efficiency.

DyLoRA extends the standard LoRA framework by introducing block-wise dynamic parameter allocation, where update resources are distributed based on importance. Rather than applying uniform low-rank decomposition across all weight matrices, DyLoRA samples blocks $b_i$ from a probability distribution $P_B(\cdot)$ and focuses on parameter updates in these regions. This targeted approach allows the model to concentrate computational resources where they will be most impactful, further reducing training overhead while maintaining or even improving adaptation quality compared to static low-rank methods.

\subsubsection{LoRA Variants}
Building on the foundational low-rank framework, enhanced variants of LoRA address domain-specific challenges and improve fine-tuning efficiency. Methods like \textbf{Laplace LoRA}  \cite{yang2023bayesian}, \textbf{LoRA Dropout}  \cite{lin2024lora}, and \textbf{Predict LoRA}  \cite{meng2024periodiclora} introduce regularization and dropout techniques to mitigate overfitting. For example, Laplace LoRA augments the decomposition with a regularization term:
\begin{equation}
    \Delta W \approx AB + \lambda I, \quad \lambda > 0
\end{equation}
where $\lambda$ controls the regularization strength. \textbf{LoRA++}  \cite{hayou2024lora+}, \textbf{MoSLoRA}  \cite{wu2024mixture}, and \textbf{LoRA for Continual Learning}  \cite{chitale2023task} are tailored to sequential learning tasks, effectively preventing catastrophic forgetting. On the other hand, \textbf{Trans-LoRA}  \cite{wang2024trans} and \textbf{RoseLoRA}  \cite{wang2024roselora} extend LoRA to transfer learning scenarios, adapting pre-trained models to new domains through task-specific subspaces.

\begin{figure}[htbp]
  \centering
  \includegraphics[width=0.66\textwidth]{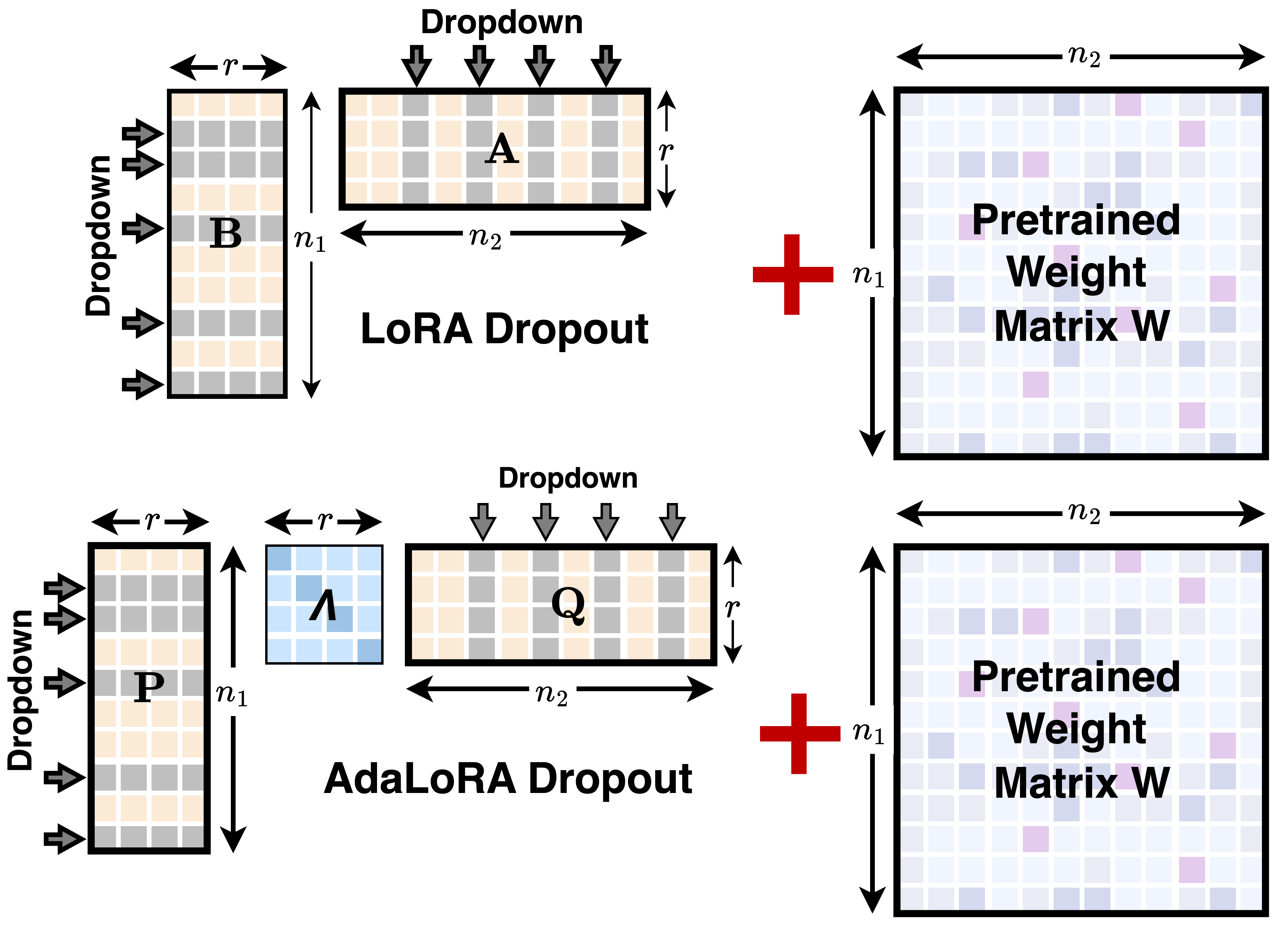}
  \caption{Comparison of dropout regularization strategies for low-rank adaptation methods: Top - LoRA Dropout applies structured dropout to low-rank matrices $A \in \mathbb{R}^{r \times n_2}$ and $B \in \mathbb{R}^{n_1 \times r}$ during training, randomly dropping elements along both row and column dimensions; Bottom - AdaLoRA Dropout extends this concept with matrices $P$, $\Lambda$, and $Q$, providing more flexible regularization patterns while maintaining the efficiency benefits of low-rank decomposition. Both approaches preserve the pretrained weight matrix $W \in \mathbb{R}^{n_1 \times n_2}$ while selectively regularizing the update components.}
  \label{fig:lora_dropout_fig}
\end{figure}

Further innovations address fine-tuning challenges in low-resource settings. \textbf{LoRA for Few-Shot Learning}  \cite{zanella2024low}, \textbf{SVDQUANT}  \cite{li2024svdqunat}, and \textbf{Variational LoRA IVON}  \cite{cong2024variational} enhance efficiency through quantization and probabilistic modeling. For example, SVDQUANT performs singular value decomposition (SVD) followed by quantization, while Variational LoRA incorporates Bayesian principles to account for uncertainty:
\begin{equation}
    p(W | \mathcal{D}) \propto p(\mathcal{D} | W) p(W), \quad W = W_0 + AB
\end{equation}
where $W_0$ is the original weight matrix, and $AB$ is the low-rank update. Ensemble methods, such as \textbf{MoeLoRA}  \cite{liu2023moelora}, \textbf{MoLoRA}  \cite{zadouri2023pushing}, and \textbf{MixLoRA}  \cite{li2024mixlora}, integrate multiple low-rank models to improve robustness and generalization. Finally, \textbf{LoRA Hub}  \cite{huang2023lorahub} consolidates diverse PEFT strategies into a unified framework, facilitating their application across varied tasks.  
\textbf{Figure \ref{fig:lora_dropout_fig}} illustrates dropout regularization techniques applied to LoRA and AdaLoRA, showcasing how selective parameter dropping during training enhances model robustness and prevents overfitting.

These dropout techniques strategically disable portions of the low-rank matrices during training, which serves multiple purposes: preventing co-adaptation of weight updates, improving generalization by creating an implicit ensemble effect, and further reducing computational demands. In LoRA Dropout, elements in matrices $A$ and $B$ are randomly zeroed according to dropout patterns along both dimensions, while AdaLoRA Dropout implements a more sophisticated approach with its three-matrix decomposition. These regularization methods are particularly valuable for scenarios where the fine-tuning dataset is limited, as they help prevent the model from simply memorizing training examples while maintaining the parameter efficiency that makes low-rank adaptation methods attractive.

\subsection{Hybrid PEFT}
Hybrid approaches in PEFT combine multiple fine-tuning strategies, such as \textbf{LoRA}, \textbf{adapters}, and \textbf{prompt-tuning}, into a unified framework to leverage the strengths of each method. By integrating these techniques, hybrid approaches provide flexibility, adaptability, and robustness across diverse tasks. These methods dynamically determine the most suitable combination of strategies to optimize performance while maintaining efficiency.

For example, the \textbf{MAM Adapter}  \cite{he2021towards} incorporates memory components into adapters, allowing task-specific information to be stored and retrieved, thereby enhancing the model’s ability to specialize in different tasks. Similarly, \textbf{UniPELT}  \cite{mao2021unipelt} (Unified Parameter-Efficient Language Tuning) integrates \textbf{LoRA}, \textbf{prefix-tuning}, and \textbf{adapters} within a single framework, enabling the model to switch dynamically between strategies depending on the task. Another prominent method, \textbf{RoSA} (Rank-Ordered Subspace Adaptation)  \cite{nikdan2024rosa}, prioritizes the most significant subspaces of parameters for fine-tuning. This is achieved by rank-ordering parameters and selecting the top-ranked ones for updates:
\[
\theta_s = \{ \theta_i \, | \, \text{rank}(\theta_i) \leq k \},
\]
where \( k \) is the threshold for the top-ranked parameters. 

Hybrid approaches often use a weighted combination of parameter updates:
\[
\theta_{\text{hybrid}} = \sum_{i=1}^n \alpha_i \theta_i,
\]
where \( \alpha_i \) represents the dynamically adjusted weight for the \( i \)-th strategy, and \( n \) is the number of integrated strategies. This framework allows hybrid methods to balance computational efficiency with task-specific adaptability. 

Additional methods such as \textbf{S4}   \cite{chen2023parameter}, \textbf{NOAH}  \cite{zhang2024neural}, \textbf{Auto PEFT}  \cite{zhou2024autopeft}, \textbf{LLM Adapter}  \cite{hu2023llm}, \textbf{SH-PEFT}  \cite{liu2024sparsity}, \textbf{Hyper PELT}  \cite{zhang2022hyperpelt}, and \textbf{Hydra}  \cite{kim2024hydra} extend the versatility of hybrid approaches by automating strategy selection, focusing on structured sparsity, or incorporating multi-headed designs for enhanced flexibility. \textbf{Figure \ref{fig:propulsion} (left)} illustrates the architecture of a unified hybrid approach, demonstrating how multiple parameter-efficient fine-tuning methods can be integrated within a single transformer block.

This unified architecture elegantly combines the strengths of multiple PEFT approaches: Adapters provide sequential transformation through bottleneck architectures, Prefix-tuning prepends learnable vectors to modify attention patterns, and LoRA applies low-rank updates to weight matrices. The inclusion of gating mechanisms ($G_A$, $G_P$, $G_L$) enables the model to dynamically weight the contribution of each method based on task requirements. This hybrid design achieves superior performance by leveraging complementary benefits: Adapters excel at capturing task-specific transformations, Prefix-tuning provides efficient context modification, and LoRA delivers parameter-efficient weight adjustments. The unified approach not only improves task performance but also enhances transfer learning capabilities across diverse domains while maintaining the parameter efficiency that makes PEFT methods attractive for resource-constrained environments.

\subsubsection{MoE-Based}  
\begin{figure}[htbp]
  \centering
  \includegraphics[width=0.5\textwidth]{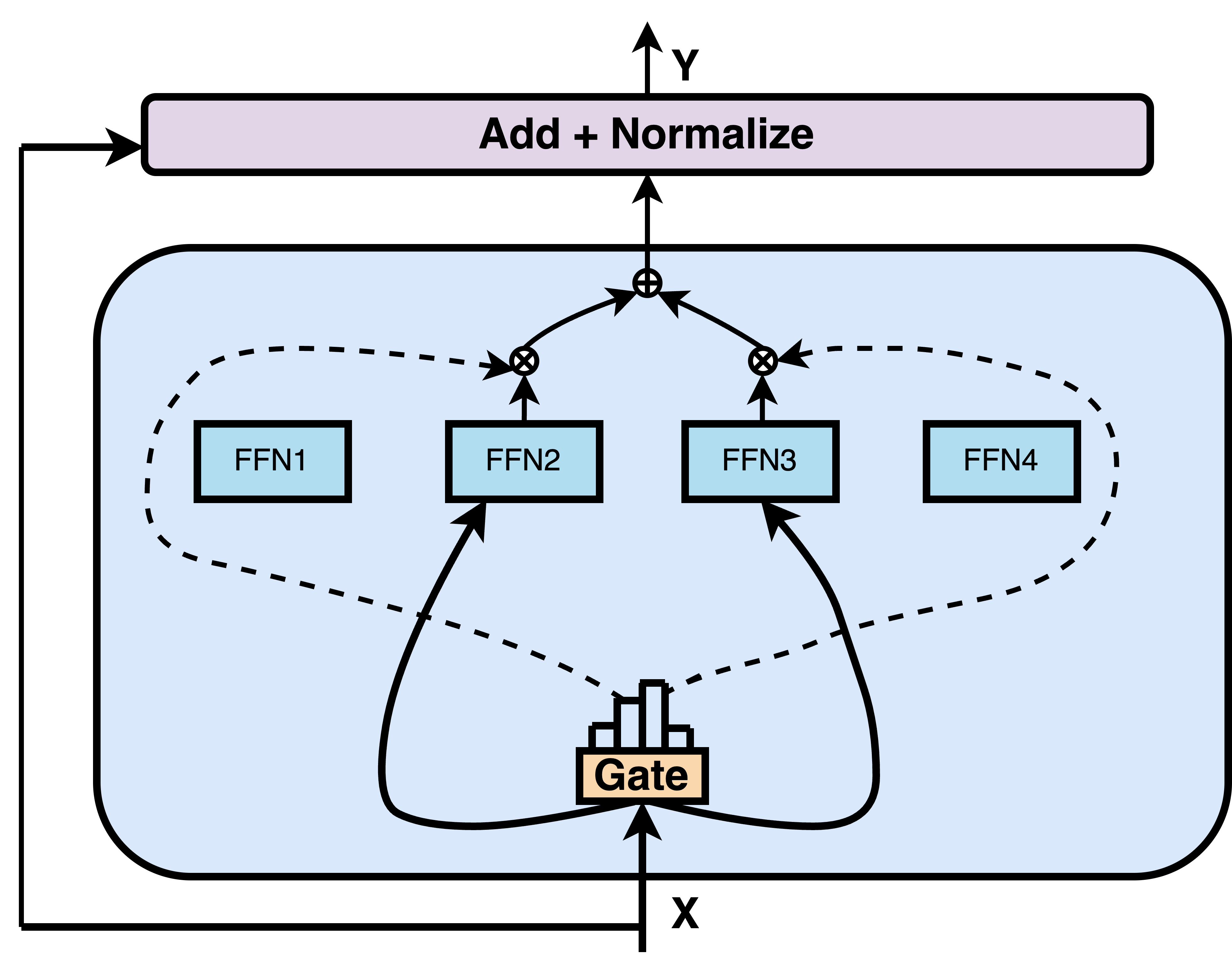}
  \caption{Illustration of a Mixture-of-Experts (MoE) feedforward network layer with gated expert selection. The input $X$ is routed to multiple feedforward sub-networks (experts), labeled as FFN1 through FFN4. A gating mechanism computes routing weights to determine which subset of experts to activate for a given input. In this example, two experts (FFN2 and FFN3) are selected and their outputs are weighted and combined. The result is added to the residual connection and passed through a normalization layer to produce the final output $Y$. This structure enables conditional computation, enhancing model capacity while maintaining computational efficiency.}
  \label{fig:MoE}
\end{figure}

An emerging class of \textbf{Mixture-of-Experts (MoE)-based} PEFT methods extends low-rank adaptation by incorporating expert routing mechanisms that dynamically select or combine multiple low-rank modules during training or inference. These methods aim to improve model specialization and generalization across diverse tasks while maintaining parameter efficiency. Formally, the update matrix $\Delta W$ is expressed as a weighted sum of expert-specific low-rank transformations:  
\begin{equation}
    \Delta W = \sum_{i=1}^{n} \alpha_i A_i B_i, \quad \sum_{i=1}^{n} \alpha_i = 1
\end{equation}
where $A_i B_i$ represents the $i$-th low-rank expert, and $\alpha_i$ is a gating coefficient determined by the MoE router. This formulation enables input-dependent specialization by activating only the most relevant subset of experts, reducing computational overhead while enhancing adaptability. Several MoE-based methods have been proposed to leverage this framework. \textbf{MoE LoRA}~ \cite{liu2023moelora} introduces a learned gating mechanism to select among multiple LoRA experts, facilitating dynamic specialization across inputs. \textbf{MixLoRA}~ \cite{li2024mixlora} combines several LoRA modules through task-aware mixture weights, improving robustness and domain generalization. \textbf{MoLoRA}~ \cite{zadouri2023pushing} routes tokens to different LoRA experts at each transformer layer, enabling fine-grained control over parameter updates. \textbf{MOA} (Mixture of Adaptations)~ \cite{feng2024mixture} generalizes this idea by integrating multiple adaptation strategies—such as LoRA, adapters, and prefix tuning—within a unified routing framework. Finally, \textbf{MoLE} (Mixture of Low-rank Experts)~ \cite{wu2024mixture} consolidates several low-rank experts and selects them dynamically based on input features, enhancing scalability and performance in multi-task and low-resource settings. \textbf{Figure \ref{fig:MoE}} illustrates the taxonomy and relationships among MoE-based PEFT methods, highlighting how they extend traditional low-rank approaches with modular, expert-driven architectures to support efficient, task-adaptive fine-tuning.

\section{Experiments}
\label{sec:experiments}

\subsection{GLUE Benchmark Performance Comparison: } 

\begin{table*}[ht]
\centering
\scalebox{.70}{
\setlength{\tabcolsep}{6.0pt}
\begin{tabular}{lccccccccc}
\hline
\rowcolor{gray!20}
\textbf{PEFT Method} & \textbf{\# TTPs} & \textbf{CoLA} & \textbf{SST2} & \textbf{MRPC} & \textbf{STS-B} & \textbf{QQP} & \textbf{MNLI} & \textbf{QNLI} & \textbf{RTE} \\
\hline

\rowcolor[HTML]{EAF3FA}\multicolumn{10}{c}{\textbf{RoBERTa\textsubscript{Base}}} \\ \hline
\rowcolor[HTML]{EAF3FA}
FT & 124.6M & 59.84 & 92.89 & 85.24/88.18 & 90.48/90.16 & 90.18/87.02 & 86.27 & 91.17 & 72.43 \\

\rowcolor[HTML]{EAF3FA}
Adapter\textsuperscript{S} & 7.41M & 61.53 & 94.11 & 89.81/90.85 & 90.25/90.09 & 89.81/86.90 & 86.27 & 92.06 & 73.56 \\

\rowcolor[HTML]{EAF3FA}
Prompt tuning & 0.61M & 49.37 & 92.09 & 70.83/81.72 & 82.44/83.11 & 82.99/78.35 & 80.57 & 80.03 & 58.12 \\

\rowcolor[HTML]{EAF3FA}
Prefix-tuning & 0.96M & 59.31 & 93.81 & 84.25/85.03 & 88.48/88.32 & 87.75/84.09 & 85.21 & 90.77 & 54.51 \\

\rowcolor[HTML]{EAF3FA}
(IA)\textsuperscript{3} & 0.66M & 58.58 & 93.92 & 83.00/85.52 & 90.30/90.32 & 87.99/84.10 & 83.95 & 90.88 & 71.12 \\

\rowcolor[HTML]{EAF3FA}
BitFit & 0.083M & 61.32 & 93.12 & 87.22/88.41 & 90.34/90.27 & 88.12/84.11 & 84.64 & 91.09 & 77.98 \\

\rowcolor[HTML]{EAF3FA}
LoRA & 0.89M & 60.09 & 93.31 & 86.50/88.68 & 90.66/90.47 & 88.83/85.21 & 86.54 & 92.02 & 74.92 \\

\rowcolor[HTML]{EAF3FA}
AdaLoRA & 1.03M & 59.82 & 93.92 & 86.49/88.03 & 90.83/90.73 & 88.58/84.98 & 86.26 & 91.43 & 76.04 \\

\rowcolor[HTML]{EAF3FA}
MAM Adapter & 1.78M & 58.34 & 94.24 & 87.31/88.21 & 90.74/90.42 & 88.31/83.20 & 86.63 & 90.19 & 72.62 \\

\rowcolor[HTML]{EAF3FA}
PROPETL\textsubscript{Adapter} & 1.87M & 64.24 & 93.85 & 87.15/87.82 & 90.33/90.64 & 89.22/85.79 & 86.49 & 91.56 & 75.54 \\

\rowcolor[HTML]{EAF3FA}
PROPETL\textsubscript{Prefix} & 10.49M & 60.11 & 93.63 & 86.73/87.98 & 90.30/90.19 & 88.54/85.05 & 86.22 & 91.51 & 63.31 \\

\rowcolor[HTML]{EAF3FA}
PROPETL\textsubscript{LoRA} & 1.77M & 57.94 & 94.11 & 87.42/88.87 & 90.66/90.35 & 88.90/85.55 & 86.83 & 92.04 & 67.39 \\ 

\rowcolor[HTML]{EAF3FA}
MoSLoRA & 1.67M & 60.57 & 93.95 & 86.74/87.98 & 90.05/89.43 & 88.76/85.62 & 87.84 & 90.60 & 75.10 \\

\hline

\rowcolor[HTML]{F8EDEB}
LoRA-XS & 0.26M & 58.49 & 93.19 & 86.65/87.49 & 89.60/89.33 & 87.13/84.31 & 85.34 & 90.42 & 76.24 \\

\rowcolor[HTML]{F8EDEB}
VeRA & 0.043M & 60.35 & 93.89 & 86.01/87.88 & 89.27/89.41 & 87.88/85.65 & 85.64 & 90.22 & 75.32 \\

\rowcolor[HTML]{F8EDEB}
LoRAFA & 0.44M & 60.49 & 93.65 & 88.18/89.98 & 90.70/90.66 & 88.90/85.46 & 86.11 & 91.42 & 76.11 \\

\rowcolor[HTML]{F8EDEB}
SFT & 0.90M & 64.45 & 94.28 & 87.74/88.64 & 89.37/89.12 & 87.24/85.11 & 86.64 & 92.11 & 78.42 \\

\rowcolor[HTML]{F8EDEB}
Diff Pruning & 1.24M & 62.45 & 93.77 & 88.00/89.21 & 89.72/90.02 & 88.62/85.54 & 85.32 & 92.14 & 77.90 \\

\rowcolor[HTML]{F8EDEB}
RoCoFT\textsubscript{1-Row} & 0.083M & 60.18 & 94.06 & 87.74/88.48 & 90.70/90.47 & 88.49/85.35 & 85.23 & 90.70 & 76.61 \\

\rowcolor[HTML]{F8EDEB}
RoCoFT\textsubscript{3-Row} & 0.249M & 63.53 & 94.92 & 87.71/88.74 & 90.89/90.49 & 88.97/85.80 & 86.73 & 92.12 & 78.31 \\

\rowcolor[HTML]{F8EDEB}
RoCoFT\textsubscript{1-Column} & 0.083M & 60.32 & 93.88 & 88.38/89.78 & 90.23/90.14 & 88.46/85.84 & 85.35 & 90.58 & 76.74 \\

\rowcolor[HTML]{F8EDEB}
RoCoFT\textsubscript{3-Column} & 0.249M & 62.95 & 94.69 & 89.18/90.94 & 90.85/90.45 & 88.86/85.38 & 86.76 & 91.89 & 75.21 \\

\rowcolor[HTML]{F8EDEB}
Propulsion\textsubscript{1-Row} & 0.086M & 61.76 & 93.18 & 89.34/85.99 & 90.37/89.92 & 89.11/86.53 & 86.41 & 91.79 & 75.66 \\

\rowcolor[HTML]{F8EDEB}
Propulsion\textsubscript{3-Row} & 0.258M & 63.21 & 94.35 & 87.28/86.12 & 90.29/90.04 & 89.42/86.84 & 87.12 & 91.56 & 76.92 \\

\rowcolor[HTML]{F8EDEB}
Propulsion\textsubscript{Attn} & 0.028M & 58.51 & 92.03 & 89.01/85.14 & 89.36/89.96 & 86.73/84.80 & 85.13 & 89.89 & 75.02 \\

\rowcolor[HTML]{F8EDEB}
SK-Tuning (Prompt) & 0.60M & 60.21 & 93.88 & 89.73/92.47 & 91.30/90.19 & 87.83/85.82 & 86.24 & 92.60 & 76.91 \\

\rowcolor[HTML]{F8EDEB}
SK-Tuning (Prefix) & 0.84M & 61.83 & 93.72 & 87.21/88.04 & 90.11/89.92 & 88.67/87.12 & 85.83 & 92.09 & 75.32 \\

\hline
\rowcolor[HTML]{ECEEFF}\multicolumn{10}{c}{\textbf{RoBERTa\textsubscript{Large}}} \\ \hline

\rowcolor[HTML]{ECEEFF}
FT & 355.3M & 65.78 & 95.50 & 92.22/94.28 & 91.74/91.96 & 90.83/88.68 & 89.21 & 93.19 & 81.40 \\

\rowcolor[HTML]{ECEEFF}
Adapter\textsuperscript{S} & 19.77M & 65.33 & 96.37 & 89.88/90.23 & 92.58/92.42 & 91.19/87.11 & 91.00 & 94.31 & 85.25 \\

\rowcolor[HTML]{ECEEFF}
Prompt-tuning & 1.07M & 61.13 & 94.61 & 73.04/76.29 & 78.51/78.99 & 80.74/75.16 & 68.15 & 89.13 & 60.29 \\

\rowcolor[HTML]{ECEEFF}
Prefix-tuning & 2.03M & 59.01 & 95.76 & 88.24/89.37 & 90.92/91.07 & 88.88/85.45 & 89.30 & 93.32 & 74.01 \\

\rowcolor[HTML]{ECEEFF}
(IA)\textsuperscript{3} & 1.22M & 61.15 & 94.61 & 86.45/87.53 & 92.22/86.25 & 89.45/86.25 & 88.63 & 94.25 & 81.23 \\

\rowcolor[HTML]{ECEEFF}
BitFit & 0.222M & 67.01 & 96.10 & 90.93/92.13 & 91.93/93.38 & 89.48/86.43 & 89.98 & 94.47 & 87.73 \\

\rowcolor[HTML]{ECEEFF}
LoRA & 1.84M & 64.47 & 96.67 & 87.50/88.19 & 91.66/91.44 & 90.15/86.91 & 90.76 & 95.00 & 79.78 \\

\rowcolor[HTML]{ECEEFF}
AdaLoRA & 2.23M & 65.85 & 94.95 & 89.46/90.34 & 92.05/91.80 & 89.60/86.30 & 90.36 & 94.62 & 77.98 \\

\rowcolor[HTML]{ECEEFF}
MAM Adapter & 4.20M & 67.39 & 95.81 & 90.12/92.07 & 92.44/92.18 & 90.87/86.65 & 90.62 & 94.31 & 86.62 \\

\rowcolor[HTML]{ECEEFF}
PROPETL\textsubscript{Adapter} & 5.40M & 65.55 & 96.27 & 89.71/91.15 & 91.92/91.67 & 90.67/87.74 & 91.37 & 94.80 & 87.69 \\

\rowcolor[HTML]{ECEEFF}
PROPETL\textsubscript{Prefix} & 26.85M & 62.24 & 96.17 & 90.04/91.92 & 90.70/90.49 & 89.30/86.30 & 90.33 & 94.73 & 79.71 \\

\rowcolor[HTML]{ECEEFF}
PROPETL\textsubscript{LoRA} & 4.19M & 61.90 & 95.93 & 87.31/89.87 & 91.66/91.38 & 90.93/88.05 & 90.53 & 94.93 & 83.57 \\

\rowcolor[HTML]{ECEEFF}
MoSLoRA & 3.23M & 67.27 & 96.17 & 89.96/92.67 & 90.97/91.72 & 90.12/87.68 & 90.29 & 94.73 & 82.41 \\
\hline

\rowcolor[HTML]{E8F5E9}
RoCoFT\textsubscript{1-Row} & 0.222M & 65.70 & 96.63 & 89.97/90.79 & 91.81/92.07 & 90.17/86.15 & 90.73 & 94.20 & 85.31 \\

\rowcolor[HTML]{E8F5E9}
RoCoFT\textsubscript{3-Row} & 0.666M & 67.39 & 96.69 & 91.05/92.19 & 92.10/92.10 & 90.82/86.11 & 90.98 & 94.85 & 87.83 \\

\rowcolor[HTML]{E8F5E9}
RoCoFT\textsubscript{1-Column} & 0.222M & 64.89 & 96.60 & 89.12/90.24 & 91.96/92.10 & 90.17/85.83 & 90.81 & 94.17 & 85.71 \\

\rowcolor[HTML]{E8F5E9}
RoCoFT\textsubscript{3-Column} & 0.666M & 67.18 & 96.67 & 89.88/91.47 & 92.52/92.31 & 91.38/87.12 & 91.13 & 94.85 & 87.82 \\

\rowcolor[HTML]{E8F5E9}
Propulsion\textsubscript{1-Row} & 0.225M & 64.53 & 95.10 & 90.47/88.85 & 91.78/91.58 & 92.26/88.91 & 90.52 & 95.34 & 85.30 \\

\rowcolor[HTML]{E8F5E9}
Propulsion\textsubscript{3-Row} & 0.675M & 67.12 & 96.68 & 91.15/92.07 & 91.68/91.81 & 91.96/87.84 & 91.42 & 95.12 & 88.28 \\

\rowcolor[HTML]{E8F5E9}
Propulsion\textsubscript{Attn} & 0.073M & 62.31 & 94.02 & 89.78/87.95 & 90.16/90.86 & 88.02/86.19 & 89.54 & 94.00 & 83.07 \\

\rowcolor[HTML]{E8F5E9}
SK-Tuning (Prompt) & 1.02M & 67.13 & 96.43 & 91.10/93.22 & 90.54/90.11 & 92.10/88.73 & 90.42 & 95.42 & 87.11 \\

\rowcolor[HTML]{E8F5E9}
SK-Tuning (Prefix) & 1.94M & 66.33 & 96.08 & 90.96/93.09 & 91.87/90.68 & 90.23/87.93 & 89.97 & 96.10 & 86.99 \\

\hline

\end{tabular}
}
\caption{RoBERTa models performance on GLUE tasks: Metrics used are MCC for CoLA, accuracy for SST-2, accuracy/F1 score for MRPC and QQP, Pearson/Spearman correlations for STS-B, and accuracy for MNLI, QNLI, and RTE.}
\label{tab:roberta_glue}
\end{table*}

We conducted a comprehensive evaluation of various PEFT methods across the General Language Understanding Evaluation (GLUE) benchmark  \cite{wang2018glue} using both RoBERTa\textsubscript{Base} and RoBERTa\textsubscript{Large} models  \cite{liu2019roberta}. The GLUE tasks include a diverse range of linguistic challenges, such as single-sentence classification (CoLA and SST-2), sentence-pair classification (MRPC, QQP, MNLI, QNLI, and RTE), and regression-based semantic similarity (STS-B). For evaluation, we employed standard metrics: Matthews Correlation Coefficient (MCC) for CoLA, accuracy for SST-2, accuracy and F1 score for MRPC and QQP, Pearson and Spearman correlations for STS-B, and accuracy for MNLI, QNLI, and RTE. This analysis aimed to determine the trade-off between model performance and parameter efficiency across established and novel PEFT techniques, including the recently introduced SK-Tuning method.

Table~\ref{tab:roberta_glue} presents a comprehensive evaluation of various parameter-efficient fine-tuning (PEFT) methods on the GLUE benchmark using RoBERTa\textsubscript{Base} and RoBERTa\textsubscript{Large} models. Full fine-tuning (FT), which updates all model parameters, consistently yields strong performance across all tasks but incurs high computational and memory costs. It serves as a performance upper bound for assessing the efficiency of PEFT techniques.

Among the PEFT baselines, \textbf{Adapter\textsuperscript{S}}, \textbf{BitFit}, and \textbf{LoRA} perform remarkably well. For example, BitFit (0.083M parameters) achieves 77.98\% on RTE and 93.12\% on SST-2, rivaling full fine-tuning. LoRA (0.89M) consistently outperforms most early PEFT methods and even FT in certain tasks, such as MNLI and QNLI. Adapter\textsuperscript{S} also demonstrates strong performance, particularly with RoBERTa\textsubscript{Large}, scoring 96.37\% on SST-2 and 85.25\% on RTE.

\textbf{Prompt-tuning} and \textbf{Prefix-tuning}, while highly parameter-efficient (under 1M parameters), generally underperform on tasks requiring fine-grained semantic understanding, such as MRPC, STS-B, and RTE. This highlights their limited expressive capacity despite their minimal footprint.

\textbf{Advanced LoRA-based methods} such as \textbf{AdaLoRA}, \textbf{MoSLoRA}, and \textbf{LoRAFA} improve performance further. AdaLoRA, for instance, achieves 76.04\% on RTE and 90.83\% on STS-B with RoBERTa\textsubscript{Base}, indicating the benefit of adaptive low-rank decompositions. MoSLoRA (1.67M) performs particularly well on MNLI (87.84\%) and QNLI (90.60\%), suggesting it captures diverse token-level information more effectively.

The \textbf{RoCoFT} and \textbf{Propulsion} families deliver better results among compact methods. RoCoFT\textsubscript{3-Row} and RoCoFT\textsubscript{3-Column} attain scores close to or exceeding FT on several tasks. Notably, RoCoFT\textsubscript{3-Row} reaches 78.31\% on RTE and 94.92\% on SST-2, with only 0.249M parameters. Similarly, Propulsion\textsubscript{3-Row} matches or surpasses strong baselines, achieving 76.92\% on RTE and 94.35\% on SST-2 with just 0.258M parameters. Even ultra-light versions like Propulsion\textsubscript{Attn} (0.028M) score competitively on tasks like STS-B and MRPC.

\textbf{SK-Tuning}, a recent method that integrates semantic knowledge into prompt and prefix tuning, demonstrates robust performance. SK-Tuning (Prompt) with 0.60M parameters achieves 92.60\% on QNLI and 76.91\% on RTE, outperforming traditional prompt-based approaches. Its prefix variant also performs well across all tasks, suggesting that semantically-aware prompting offers a powerful alternative for low-resource fine-tuning.

Finally, comparing across model sizes, PEFT methods applied to RoBERTa\textsubscript{Large} typically outperform their RoBERTa\textsubscript{Base} counterparts by a significant margin. For instance, RoCoFT\textsubscript{3-Row} achieves 87.83\% on RTE with RoBERTa\textsubscript{Large}, compared to 78.31\% with RoBERTa\textsubscript{Base}, highlighting the scaling benefits of PEFT with larger backbones.

In summary, modern PEFT methods—particularly LoRA-based variants, RoCoFT, Propulsion, and SK-Tuning—approach or even surpass full fine-tuning performance on many GLUE tasks while drastically reducing the number of updated parameters. This makes them highly attractive for efficient and scalable deployment of large language models in both academic and production settings.

\subsection{LLM Reasoning PEFT Comparison : } 

\begin{table*}[htbp]
\centering
\scalebox{.70}{
\setlength{\tabcolsep}{3.4pt}
\begin{tabular}{c|ccccccccc|ccccc}
\rowcolor{gray!20}
\hline
\textbf{Method} &  \textbf{\# TTPs} &\textbf{BoolQ} & \textbf{PIQA} & \textbf{SIQA} & \textbf{H.Sw.} & \textbf{W.Gra.} & \textbf{ARCe} & \textbf{ARCc} & \textbf{OBQA} & \textbf{M.Ar.} & \textbf{G.8K} & \textbf{A.S.} & \textbf{Sing.Eq} & \textbf{S.MP}\\
\hline
\rowcolor[HTML]{FFF9C4}\multicolumn{15}{c}{\textbf{BLOOM\textsubscript{7B}}} \\ \hline
\rowcolor[HTML]{FFF9C4}
 Prefix & 33.37M & 58.53 & 62.24 & 65.41 & 48.32 & 66.63 & 68.13 & 49.32 & 63.51 & 78.41 & 66.45 & 67.52 & 66.94 & 49.10 \\
\rowcolor[HTML]{FFF9C4}
 AdaLoRA & 24.88M & 66.94 & 74.68 & 72.49 & 55.89 & 68.30 & 73.21 & 56.59 & 72.85 & 79.43 & 70.25 & 68.93 & 70.93 & 53.89 \\
\rowcolor[HTML]{FFF9C4}
 (IA)\textsuperscript{3} & 19.34M & 63.30 & 73.33 & 71.01 & 52.50 & 71.60 & 69.45 & 54.14 & 68.60 & 78.90 & 71.17 & 70.33 & 70.84 & 53.95 \\
\rowcolor[HTML]{FFF9C4}
 LoRA & 24.22M & 65.89 & 73.92 & 73.33 & 56.65 & 71.39 & 73.46 & 57.15 & 72.31 & 79.50 & 70.93 & 70.90 & 70.59 & 54.85 \\
\rowcolor[HTML]{FFF9C4}
 RoCoFT\textsubscript{3-Row} & 13.37M & 66.33 & 74.53 & 73.56 & 56.60 & 72.14 & 73.29 & 57.48 & 72.92 & 79.76 & 70.94 & 70.95 & 70.90 & 54.42 \\
\rowcolor[HTML]{FFF9C4}
 RoCoFT\textsubscript{3-Column} & 13.37M & 66.34 & 74.64 & 71.12 & 55.93 & 72.50 & 73.11 & 57.19 & 72.90 & 79.72 & 71.05 & 70.88 & 70.76 & 54.38 \\
\rowcolor[HTML]{FFF9C4}
 Propulsion & 13.37M & 66.38 & 74.63 & 73.62 & 57.25 & 72.33 & 73.09 & 57.61 & 73.12 & 79.36 & 70.95 & 70.92 & 71.22 & 53.52 \\
\hline

\rowcolor[HTML]{E0F7FA}\multicolumn{15}{c}{\textbf{GPT-J\textsubscript{6B}}} \\ \hline
\rowcolor[HTML]{E0F7FA}
 Prefix & 27.83M & 62.28 & 65.04 & 67.72 & 44.15 & 63.71 & 63.59 & 46.47 & 58.31 & 83.12 & 67.44 & 75.25 & 78.46 & 49.12 \\
\rowcolor[HTML]{E0F7FA}
 AdaLoRA & 20.77M & 65.19 & 67.58 & 71.22 & 45.16 & 66.03 & 64.10 & 47.75 & 63.92 & 88.51 & 73.45 & 80.21 & 83.03 & 56.14 \\
\rowcolor[HTML]{E0F7FA}
 (IA)\textsuperscript{3} & 16.61M & 63.17 & 68.51 & 68.97 & 45.79 & 66.06 & 62.42 & 45.32 & 65.42 & 89.51 & 72.04 & 80.50 & 81.50 & 55.43 \\
\rowcolor[HTML]{E0F7FA}
 LoRA & 20.02M & 65.50 & 68.63 & 69.46 & 45.60 & 66.80 & 65.56 & 46.81 & 63.82 & 88.30 & 72.82 & 80.60 & 81.24 & 56.73 \\
\rowcolor[HTML]{E0F7FA}
 RoCoFT\textsubscript{3-Row} & 11.62M & 65.92 & 68.53 & 69.90 & 45.97 & 66.87 & 64.91 & 45.12 & 65.07 & 89.45 & 72.80 & 80.45 & 82.12 & 56.79 \\
\rowcolor[HTML]{E0F7FA}
 RoCoFT\textsubscript{3-Column} & 11.62M & 65.12 & 68.22 & 69.96 & 45.98 & 66.78 & 64.89 & 45.70 & 64.81 & 89.74 & 72.24 & 80.23 & 82.61 & 56.70 \\
\rowcolor[HTML]{E0F7FA}
 Propulsion & 11.62M & 65.97 & 68.05 & 69.96 & 45.99 & 66.18 & 64.45 & 46.95 & 64.56 & 89.19 & 72.82 & 81.41 & 81.42 & 56.68 \\
\hline

\rowcolor[HTML]{F1F8E9}\multicolumn{15}{c}{\textbf{LLama-2\textsubscript{7B}}} \\ \hline
\rowcolor[HTML]{F1F8E9}
 Prefix & 33.53M & 67.33 & 79.46 & 75.80 & 76.04 & 72.11 & 71.67 & 57.33 & 69.98 & 84.18 & 68.47 & 81.04 & 80.00 & 52.17 \\
\rowcolor[HTML]{F1F8E9}
 AdaLoRA & 24.90M & 67.03 & 80.69 & 76.06 & 88.85 & 76.47 & 76.50 & 61.36 & 74.22 & 89.81 & 77.07 & 86.70 & 83.01 & 60.25 \\
\rowcolor[HTML]{F1F8E9}
 (IA)\textsuperscript{3} & 19.42M & 69.02 & 78.10 & 78.00 & 87.57 & 76.78 & 75.48 & 60.54 & 74.02 & 90.20 & 76.13 & 86.55 & 83.70 & 59.16 \\
\rowcolor[HTML]{F1F8E9}
 LoRA & 24.30M & 69.89 & 79.37 & 76.15 & 88.86 & 77.54 & 76.54 & 60.55 & 74.63 & 90.13 & 75.68 & 84.67 & 82.14 & 59.94 \\
\rowcolor[HTML]{F1F8E9}
 RoCoFT\textsubscript{3-Row} & 13.47M & 69.36 & 80.01 & 78.09 & 87.28 & 76.73 & 76.46 & 60.55 & 75.55 & 90.37 & 76.12 & 86.66 & 82.75 & 59.92 \\
\rowcolor[HTML]{F1F8E9}
 RoCoFT\textsubscript{3-Column} & 13.47M & 69.32 & 80.08 & 77.99 & 87.46 & 76.41 & 76.46 & 60.59 & 74.90 & 90.42 & 77.35 & 86.16 & 82.48 & 60.35 \\
\rowcolor[HTML]{F1F8E9}
 Propulsion & 13.47M & 68.99 & 79.47 & 77.02 & 76.73 & 76.06 & 76.64 & 61.29 & 74.76 & 90.21 & 77.57 & 85.63 & 82.60 & 60.51 \\
\hline

\rowcolor[HTML]{FFF3E0}\multicolumn{15}{c}{\textbf{LLama-2\textsubscript{13B}}} \\ \hline
\rowcolor[HTML]{FFF3E0}
 Prefix & 61.97M & 68.38 & 80.99 & 77.80 & 80.00 & 76.35 & 77.62 & 61.32 & 72.94 & 87.22 & 71.09 & 84.09 & 81.28 & 58.25 \\
\rowcolor[HTML]{FFF3E0}
 AdaLoRA & 45.04M & 71.71 & 82.55 & 78.88 & 91.60 & 83.01 & 83.04 & 67.33 & 81.76 & 90.55 & 80.19 & 87.00 & 87.10 & 66.03 \\
\rowcolor[HTML]{FFF3E0}
 (IA)\textsuperscript{3} & 36.02M & 71.39 & 83.33 & 78.32 & 92.40 & 83.24 & 83.34 & 66.43 & 80.99 & 91.88 & 79.24 & 88.16 & 87.08 & 67.03 \\
\rowcolor[HTML]{FFF3E0}
 LoRA & 44.94M & 71.19 & 83.99 & 79.15 & 91.86 & 83.24 & 83.35 & 67.05 & 81.37 & 91.27 & 78.90 & 86.89 & 86.07 & 65.85 \\
\rowcolor[HTML]{FFF3E0}
 RoCoFT\textsubscript{3-Row} & 24.88M & 71.46 & 83.32 & 79.54 & 91.86 & 83.22 & 83.65 & 67.12 & 81.54 & 90.69 & 79.70 & 88.24 & 87.28 & 66.60 \\
\rowcolor[HTML]{FFF3E0}
 RoCoFT\textsubscript{3-Column} & 24.88M & 71.44 & 83.52 & 79.50 & 91.84 & 83.20 & 83.39 & 67.06 & 81.73 & 91.46 & 79.63 & 88.11 & 87.58 & 66.63 \\
\rowcolor[HTML]{FFF3E0}
 Propulsion & 24.88M & 71.93 & 83.12 & 79.01 & 90.73 & 83.60 & 83.44 & 67.64 & 81.38 & 90.91 & 78.71 & 87.64 & 87.11 & 66.67 \\
\hline
\end{tabular}
}
\caption{Accuracy comparison of commonsense and mathematical reasoning performance across different PEFT methods using LLMs.}
\label{table:LLM_results}
\end{table*}

Table~\ref{table:LLM_results} provides a detailed comparison of PEFT methods on a diverse set of reasoning tasks—including commonsense (BoolQ  \cite{clark2019boolq}, PIQA  \cite{bisk2020piqa}, SIQA  \cite{sap2019socialiqa}, HellaSwag  \cite{zellers2019hellaswag}, WinoGrande  \cite{sakaguchi2021winogrande}, ARCe  \cite{clark2018think}, ARCc  \cite{clark2018think}, OBQA  \cite{mihaylov2018can}) and mathematical/logical reasoning (MathQA  \cite{amini2019mathqa}, GSM8K  \cite{cobbe2021training}, Arithmetic Sequence (A.S.)  \cite{hosseini2014learning}, SVAMP  \cite{patel2021nlp}, and SingleEq  \cite{koncel2015parsing}). The analysis spans four large language models (LLMs): BLOOMZ\textsubscript{7B}  \cite{workshop2022bloom}, GPT-J\textsubscript{6B}  \cite{mesh-transformer-jax}, LLaMA-2\textsubscript{7B}  \cite{touvron2023llama}, and LLaMA-2\textsubscript{13B}  \cite{touvron2023llama}. We observe consistent patterns in performance improvements across PEFT methods and models.

Across all LLMs, full prefix tuning serves as a baseline and generally underperforms compared to more advanced PEFT methods, despite using a relatively large number of trainable parameters (e.g., 61.97M for LLaMA-2\textsubscript{13B}). In contrast, \textbf{AdaLoRA}, \textbf{(IA)\textsuperscript{3}}, and \textbf{LoRA} deliver substantial gains in reasoning benchmarks while reducing the parameter budget by 25–40\%. Notably, AdaLoRA achieves robust results across most tasks, particularly with LLaMA-2\textsubscript{13B}, scoring 91.60\% on HellaSwag, 83.01\% on WinoGrande, and 66.03\% on SingleEq.

\textbf{RoCoFT} and \textbf{Propulsion}, both low-rank, structure-aware fine-tuning strategies, consistently match or outperform other PEFT baselines with significantly fewer trainable parameters. For example, RoCoFT\textsubscript{3-Row} and Propulsion, each with only 13.37M parameters on BLOOMZ\textsubscript{7B}, outperform both AdaLoRA and LoRA on multiple tasks such as PIQA (74.63\%) and SIQA (73.62\%), while maintaining comparable scores on MathQA, GSM8K, and SVAMP. On GPT-J\textsubscript{6B}, RoCoFT and Propulsion similarly demonstrate improvements over LoRA, especially on arithmetic and symbolic reasoning benchmarks like AquaRat and SVAMP, reflecting their potential to capture deeper reasoning patterns with minimal parameter cost.

With LLaMA-2\textsubscript{7B}, performance increases across the board. LoRA and RoCoFT\textsubscript{3-Row} show strong results on difficult commonsense tasks such as HellaSwag (88.86\% and 87.28\%) and ARCc (60.55\% for both). Meanwhile, Propulsion achieves near-competitive results (e.g., 77.57\% on GSM8K) while maintaining efficiency. This further supports that structural PEFT methods can scale well to larger models without compromising generalization ability.

The LLaMA-2\textsubscript{13B} model yields the highest overall accuracy, with all PEFT methods outperforming their smaller model counterparts. RoCoFT\textsubscript{3-Row} and Propulsion reach peak performance on SIQA (79.54\%), HellaSwag (91.86\%), and OBQA (81.54\%), matching or exceeding AdaLoRA despite requiring nearly half the trainable parameters. For mathematical reasoning tasks like GSM8K and AquaRat, (IA)\textsuperscript{3} and Propulsion offer strong performance, indicating that selective structural adaptation helps retain precision in arithmetic operations and symbolic pattern generalization.

In summary, while classical methods such as LoRA and AdaLoRA continue to perform strongly, newer PEFT techniques like RoCoFT and Propulsion demonstrate impressive performance-per-parameter efficiency across a wide range of reasoning tasks and model sizes. These approaches not only reduce computational costs but also scale robustly with model size, making them ideal for fine-tuning large LLMs on complex reasoning domains in real-world applications.

\section{Applications} 
\label{sec:applications}

\begin{figure}[htbp]
\centering
\includegraphics[width=0.85\textwidth]{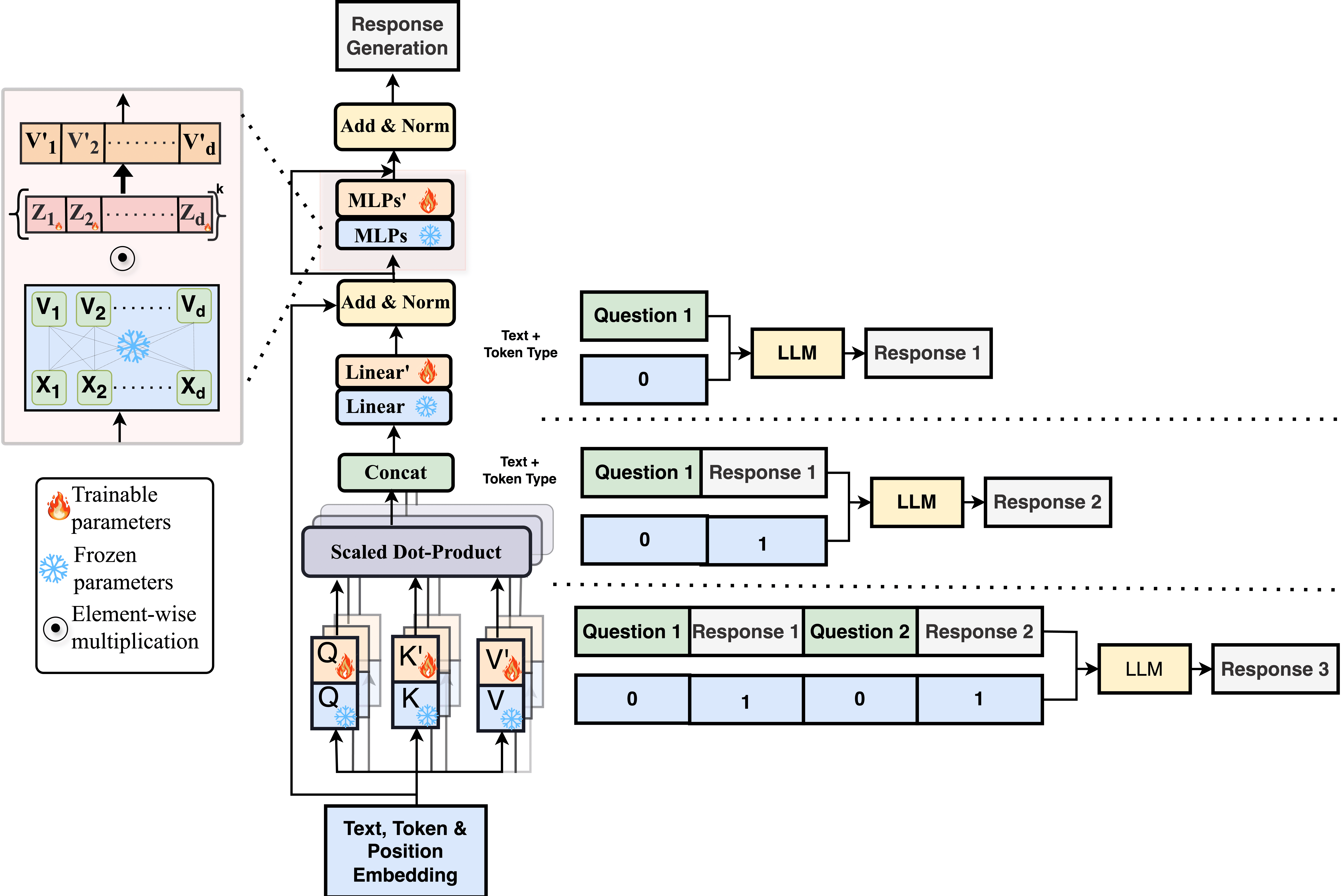}
\caption{The figure depicts a Propulsion into ChatLLM framework. Token-type embeddings handle multi-turn dialogue, while a reparameterization mechanism modulates activations via element-wise multiplication for efficient adaptation in tasks like instruction tuning and personalization.}\label{fig:NLP}
\end{figure}

\subsection{PEFT in NLP}

PEFT techniques have been widely adopted across a range of NLP applications, offering an efficient way to adapt large language models (LLMs) to task-specific Applications without incurring the high cost of full fine-tuning. In text classification tasks such as sentiment analysis  \cite{wankhade2022survey, medhat2014sentiment,taboada2016sentiment, mejova2009sentiment, cambria2017practical, hussein2018survey, liu2010sentiment, devika2016sentiment}, spam detection   \cite{crawford2015survey, jindal2007review, markines2009social, wang2010don, xie2012review, wang2011social}, and topic categorization   \cite{sriurai2011improving, kanavos2015topic, zhou2009text,
zhou2007topic, qu2012evaluation}, PEFT methods allow models to be fine-tuned on relatively small labeled datasets while retaining strong performance, particularly in low-resource or domain-specific settings. For sequence generation tasks like text summarization   \cite{liu2019text, widyassari2022review, el2021automatic, tas2007survey, gambhir2017recent, hovy2005text, abualigah2019text}, information extraction  \cite{prottasha2025user}, and machine translation   \cite{wang2022progress, lopez2008statistical,
koehn2009statistical, hutchins1995machine, kenny2018machine, wu2016google}, PEFT enables the model to adapt to domain-specific vocabulary and style, achieving competitive results with a fraction of the training parameters.

In dialogue systems, especially in multi-turn chat applications   \cite{li2021dialogue, zhang2024beyond, kwan2024mt, kao2019model, wang2019multi}, PEFT plays a crucial role by enabling LLMs to handle evolving context and intent across conversation turns. Notably, PEFT has been integrated into frameworks like ChatLLM   \cite{kowsher2024token, hao2025chatllm, rovnyagin2024optimizing}, where it supports efficient training and deployment of chat models by modifying only selected parameters—such as adapters or token embeddings—while keeping the core model frozen. This allows for rapid customization to different user personas, use-cases, or industries (e.g., healthcare, customer support) without retraining the entire model.

PEFT is also instrumental in instruction tuning and prompt-based learning, where models are aligned to follow specific instructions or exhibit desired behavioral traits. In few-shot and zero-shot scenarios, PEFT enables effective model adaptation with very limited examples. Furthermore, in multi-task and cross-lingual NLP setups   \cite{chen2018multi, 
rovnyagin2024optimizing, hao2025chatllm, conneau2019cross}, PEFT allows a single LLM to be fine-tuned for multiple tasks or languages using task-specific adapters, thus promoting parameter sharing and memory efficiency.

An overview of this architecture is shown in \textbf{Figure~\ref{fig:NLP}}, where the Propulsion method  \cite{kowsher2024propulsion} is integrated into the ChatLLM framework  \cite{kowsher2024token, hao2025chatllm} for training chat models with LLMs.

Additionally, we present a comprehensive summary of PEFT methods applied in NLP tasks across various LLMs and datasets in Tables~\ref{tab:dataset-overview}, \ref{tab:dataset-overview1}, \ref{tab:dataset-overview2}, \ref{tab:peft_methods_nlp1}, \ref{tab:peft_methods_nlp}, \ref{tab:peft_methods_nlp_combined2}, \ref{tab:peft_methods_nlp_combined3}, and  \ref{tab:peft_methods_nlp4}.

Overall, PEFT has become a cornerstone of practical NLP development, making the deployment of powerful LLMs feasible for a wide variety of real-world applications and resource-constrained environments.

\begin{figure}[htbp]
\centering
\includegraphics[width=0.8\textwidth]{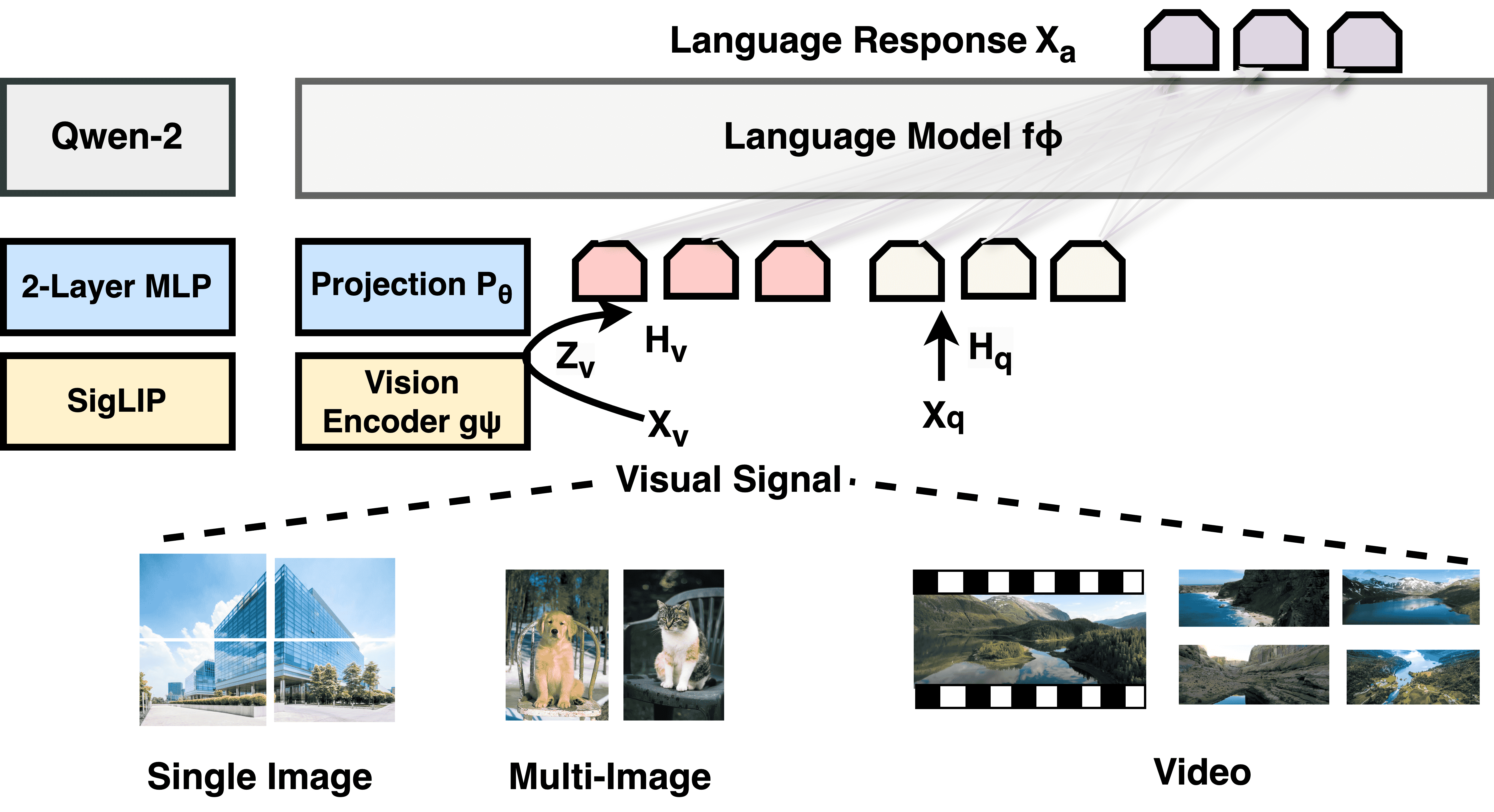}
\caption{Diagram Description: The figure demonstrates a pipeline where vision embeddings (from SigLIP) are projected into the LLM input space via trainable layers. The model processes varying visual contexts (e.g., a cat or a natural scene) to produce structured language outputs, suitable for captioning, summarization, and vision-grounded reasoning.}\label{fig:Vision}
\end{figure}

\subsection{PEFT in Vision}

PEFT has also gained significant traction in computer vision, where large vision models—such as Vision Transformers (ViTs)  \cite{dosovitskiy2020image} and large-scale convolutional neural networks (CNNs)  \cite{maggiori2016convolutional, kowsher2024token, chauhan2018convolutional} —are increasingly being adapted to diverse downstream tasks. In image classification  \cite{wang2019metamorphic, lu2007survey, chen2021review,  rawat2017deep}, PEFT enables efficient domain adaptation by allowing models pretrained on large datasets like ImageNet   \cite{wang2019learning} to be fine-tuned on task-specific or domain-specific datasets (e.g., medical imaging  \cite{kasban2015comparative, hounsfield1980computed, shung2012principles, beutel2000handbook, ganguly2010medical, suzuki2017overview}, satellite imagery  \cite{borra2019satellite, basu2015deepsat, dare2005shadow, jean2016combining, reed1994measuring}) with minimal additional parameters. This is particularly valuable in cases where data is scarce or computational resources are limited.

In object detection  \cite{zou2023object, zhao2019object, amit2021object, 
papageorgiou2000trainable} and instance segmentation  \cite{ke2021deep}, PEFT techniques such as adapter-based tuning or bias tuning have been applied to integrate task-specific knowledge into large vision backbones like DETR and Mask R-CNN  \cite{girshick2014rich, he2017mask, bharati2020deep}. This allows the base detection models to be repurposed for new object categories or specialized detection tasks (e.g., autonomous driving  \cite{yurtsever2020survey, levinson2011towards,maurer2016autonomous, huang2018apolloscape, caesar2020nuscenes, campbell2010autonomous}, surveillance  \cite{valera2005intelligent, klaucke1988guidelines, elharrouss2021review,
emori1991national, cucchiara2005multimedia, 
mokdad2009behavioral}) without the need to retrain all parameters.

PEFT has also shown promise in vision-language tasks such as image captioning  \cite{ke2019reflective, you2016image, hossain2019comprehensive}, visual question answering (VQA)  \cite{kafle2017visual, antol2015vqa, wu2017visual, goyal2017making}, and referring expression comprehension, where it helps adapt multimodal models like BLIP  \cite{li2022blip, li2023blip, cabero2019blip}, Flamingo  \cite{alayrac2022flamingo, childress2008flamingo, mccarthy2023flamingo}, and CLIP  \cite{dai2022enabling, hafner2021clip, shen2021much, sun2023eva} to specific domains or tasks. In such multimodal setups, PEFT modules can be injected into either the visual encoder, the language decoder, or their cross-attention layers to steer the joint representation learning efficiently.

Furthermore, PEFT facilitates continual learning in vision  \cite{xin2024parameter}, enabling models to incorporate new classes or tasks incrementally without catastrophic forgetting. In few-shot and zero-shot image classification scenarios, PEFT makes it feasible to quickly adapt models with very limited supervision  \cite{kilminster2000effective, loganbill1982supervision, holloway1995supervision,
hogan1964issues, oliva2004supervision, wiles1975supervision}.

As an example, PEFT techniques have enabled the integration of frozen visual encoders, such as SigLIP  \cite{zhai2023sigmoid, tschannen2025siglip}, with pretrained LLMs for language-guided visual reasoning tasks. Similar to the X2L framework shown in \textbf{Figure \ref{fig:Vision}}, the key innovation lies in the use of a lightweight two-layer multilayer perceptron (MLP) and a projection matrix $P_\theta$, trained to convert visual features into a token-compatible format that can be understood by the LLM. This visual adapter module performs the necessary transformation without altering the vision encoder or language model, significantly reducing the number of parameters requiring training. This parameter-efficient approach supports diverse input formats---including static images, video frames, and multi-image sequences---making it highly suitable for tasks like temporal visual reasoning, multi-image comparison, and descriptive captioning. By grounding image features into a language-aligned semantic space, this PEFT-driven architecture ensures generalizability across domains and tasks without necessitating re-training of foundational models.

We present a comprehensive summary of PEFT methods applied in vision tasks across various LLMs and datasets in Tables~\ref{tab:dataset-overview}, \ref{tab:dataset-overview1}, \ref{tab:dataset-overview2}, \ref{tab:peft_methods_vision5}, and \ref{tab:vision-peft6}.

\begin{figure*}[htbp]
\centering
\includegraphics[width=0.8\textwidth]{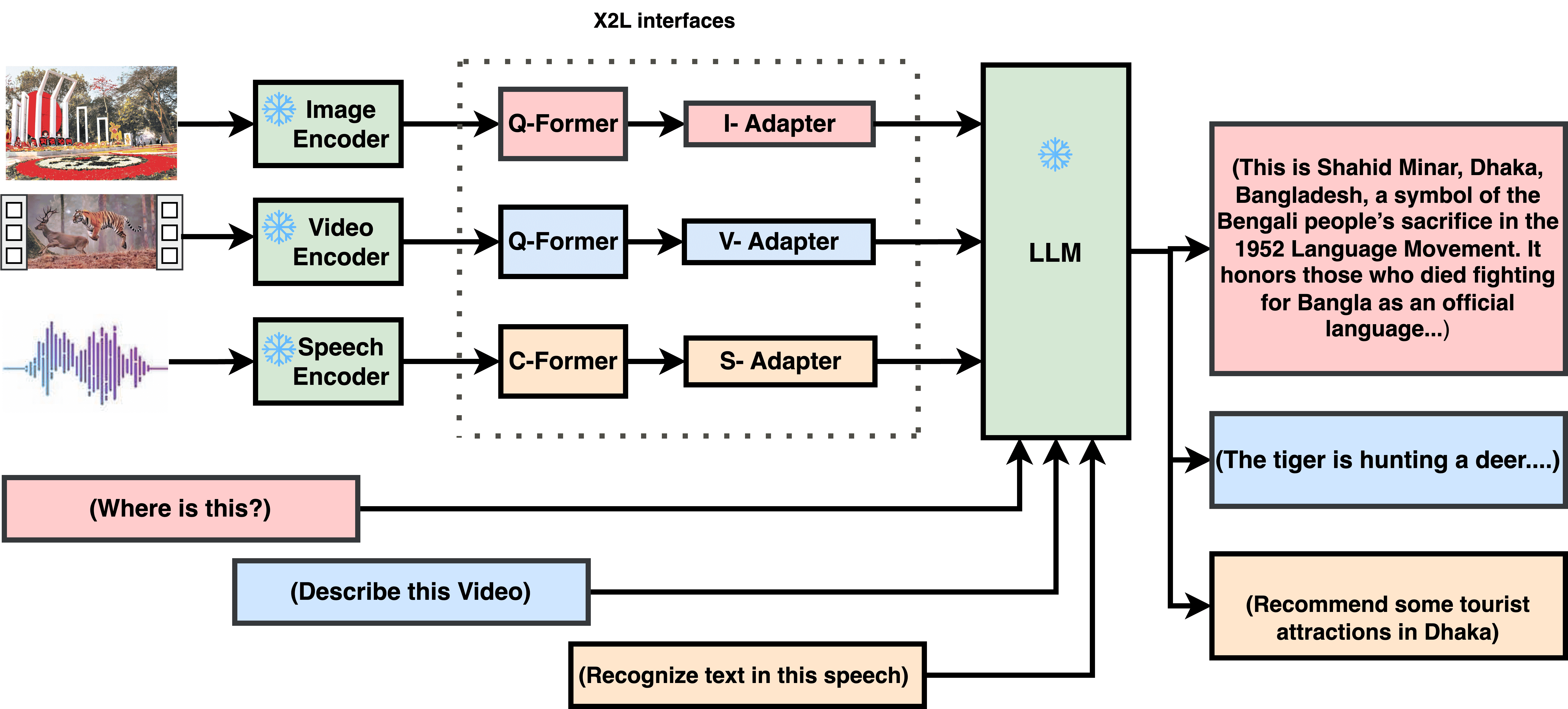}
\caption{This figure outlines an X2L (Cross-modal to Language) framework where modality-specific encoders feed into lightweight adapters, which in turn align with a frozen LLM. Example outputs demonstrate cross-lingual generation grounded in images, videos, and speech, showcasing real-world applicability in multilingual and multimedia environments.}\label{fig:Multimodel}
\end{figure*}

\subsection{PEFT in Multimodal Learning}

PEFT has become increasingly important in multimodal learning, where models process and integrate information from multiple input modalities—such as vision  \cite{sagi1985and, vision1996demystifying, avant1965vision, witkin1983role,
devalois1990spatial, 
fry2012vision, yu2022metaformer}, language  \cite{cohen2000data, wolff2003integrating, 
hartsuiker2008language, liang2024survey, 
crawford1987integrating}, audio  \cite{chen1998audio, benoit2000audio,
foote1999overview, crocco2016audio, 
kamuni2024advancing, cobos2022overview, wang2022deep}, and video  \cite{ramanishka2016multimodal,
snoek2005multimodal, 
li2024llava, wang2018m3, 
hori2017attention, luo2020univl, wang2024multimodal}. Modern multimodal architectures, like CLIP  \cite{
dai2022enabling, hafner2021clip, shen2021much, sun2023eva}, Flamingo  \cite{alayrac2022flamingo, childress2008flamingo, mccarthy2023flamingo}, BLIP  \cite{li2022blip, li2023blip, cabero2019blip}, PaLI  \cite{chen2022pali}, and Video-LLaMA  \cite{zhang2023video}, typically consist of large pretrained encoders and decoders spanning both visual and textual domains. Fine-tuning these models entirely is computationally expensive and memory-intensive, especially when adapting to new modalities, tasks, or domains. PEFT addresses this challenge by introducing lightweight, task-specific modules—such as adapters, low-rank matrices, or reparameterized prompts—into selected parts of the multimodal pipeline, allowing efficient and scalable adaptation.

In vision-language tasks like image captioning  \cite{ke2019reflective,you2016image, vinyals2016show, yao2017boosting, rennie2017self}, visual question answering (VQA)  \cite{goyal2017making, ke2019reflective, you2016image, hossain2019comprehensive}, and cross-modal retrieval  \cite{zhen2019deep, wang2016comprehensive, 
wang2017adversarial, wang2025cross, 
wei2016cross, feng2014cross, jing2021cross}, PEFT modules are often injected into the cross-attention layers between vision and text components, enabling the model to learn task-specific alignments without modifying the full backbone. Similarly, in video-language models used for tasks such as video QA  \cite{yang2003videoqa, 
zeng2017leveraging, 
yu2019activitynet, choi2021dramaqa,
khurana2021video, lei2018tvqa}, temporal grounding  \cite{chen2021end, lin2023univtg, mun2020local,
lei2019tvqa+, wang2022negative, zhang2020does, gu2024context}, and action recognition  \cite{kong2022human, jhuang2013towards, sun2022human, wang2013action, ji20123d,
herath2017going, carreira2017quo}, PEFT enables efficient fine-tuning on long video sequences by adapting only certain projection or fusion layers while freezing the majority of the vision encoder and language decoder.

Multimodal instruction tuning is another growing area where PEFT is heavily used, especially for aligning models to follow visual and language instructions together. In models like InstructBLIP  \cite{panagopoulou2023x, wang2023instructta} and MiniGPT-4  \cite{zhu2023minigpt, yao2024minicpm, yuan2023artgpt, ataallah2024minigpt4, 
azizi2024minigpt}, PEFT techniques allow fast customization to downstream multimodal tasks such as referring expression comprehension, image editing via text commands  \cite{brooks2023instructpix2pix, kopec1990editing, faltings2020text, 
laput2013pixeltone, saund2003perceptually}, and multimodal dialogue  \cite{johnston2002match, trung2006multimodal, 
liao2018knowledge, jovanovic2018multimodal, walker2004generation}, all with limited supervision. Moreover, in low-resource or domain-specific settings (e.g., medical image–report generation or surveillance video QA), PEFT allows multimodal models to generalize effectively by training only a small subset of parameters.

As demonstrated in \textbf{Figure \ref{fig:Multimodel}}, each modality---whether vision, audio, or text---is processed by a frozen encoder, such as SigLIP  \cite{zhai2023sigmoid, tschannen2025siglip} for images or pretrained audio models for speech. These encoders generate modality-specific embeddings, which are aligned into a shared latent space using transformer modules like Q-Former for visual features and C-Former for speech. Instead of re-training these components, PEFT introduces modality-specific lightweight adapters---namely, I-Adapter (image), V-Adapter (video), and S-Adapter (speech)---which serve as narrow bottleneck modules for mapping each modality's features into a unified token stream consumable by a frozen LLM. This approach localizes the adaptation to specific components, enabling the model to support cross-modal reasoning and generation tasks, such as bilingual captioning or voice-command interpretation, without degrading performance on prior capabilities. The decoupled design makes it possible to incrementally expand the system to new modalities with minimal overhead, exemplifying PEFT's utility in extensible and memory-efficient architectures.

\begin{figure}[htbp]
\centering
\includegraphics[width=0.8\textwidth]{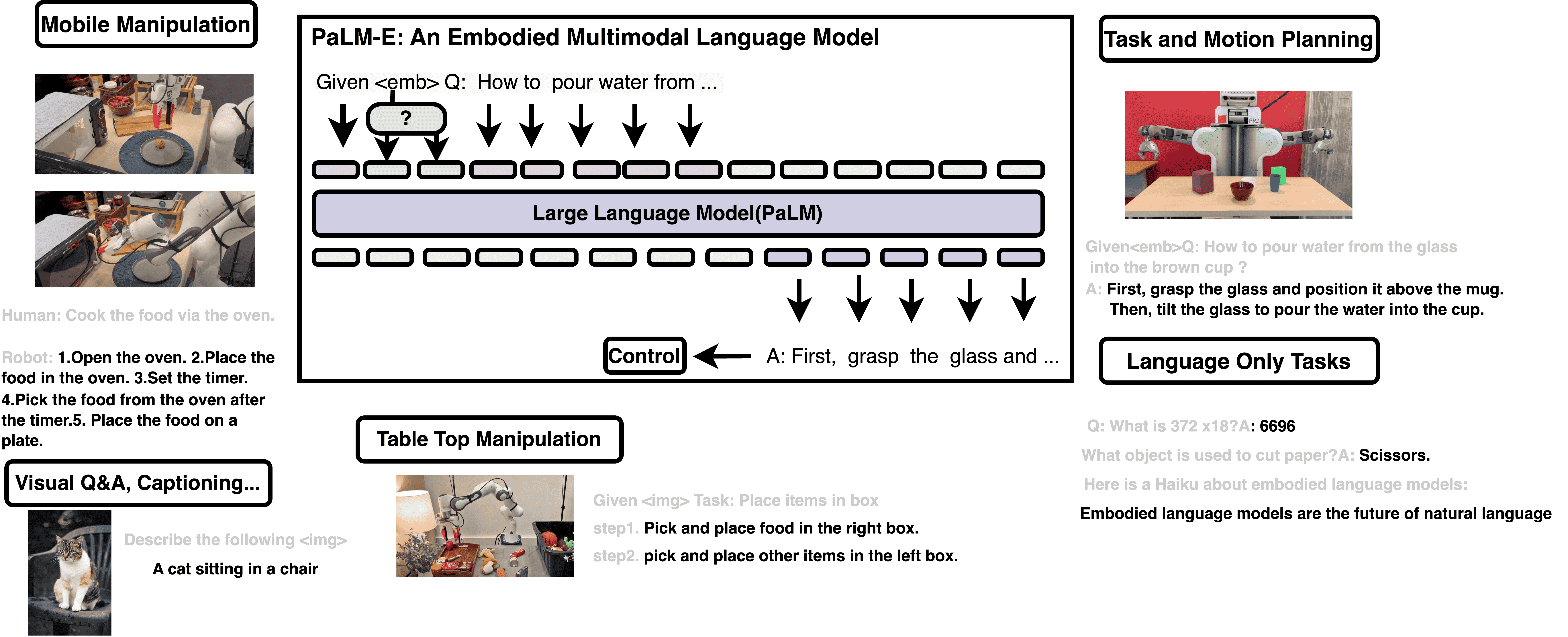}
\caption{The figure presents various real-world robotic use cases, such as cooking, object sorting, and visual Q \& A, all mediated through a unified LLM that integrates perception and control. By interpreting complex queries and translating them into actionable steps, the model exemplifies the future of grounded language understanding in robotics.}\label{fig:Robotic}
\end{figure}

\subsection{PEFT in Robotics}

In robotics, PEFT is increasingly being used to adapt large pretrained models—particularly vision-language and policy models—for control   \cite{luo2023controlling, saxena2023multi, 
zhao2024vlmpc}, perception   \cite{xie2003fundamentals, 
ollero2004control, traver2010review, 
navarro2021proximity, yan2014survey, wulfmeier2021representation, sanders1999perception}, and decision-making tasks   \cite{kaupp2010human, agostini2017efficient, 
mojtahedzadeh2015support,tsymbal2019adaptive, unhelkar2020decision, mihaylova2003comparison} in physical environments. Robotics systems often require integrating visual input, natural language instructions, and low-level control signals to perform complex tasks in real-world settings. However, fine-tuning large multimodal models for each specific robot, environment, or task is often impractical due to computational and data constraints. PEFT provides a practical solution by allowing targeted fine-tuning of small subsets of parameters, such as adapters or low-rank projections, while keeping the majority of the base model frozen.

For instance, in vision-language-action models used in imitation learning and instruction following, PEFT enables adaptation to new environments or unseen tasks with minimal retraining. Techniques like LoRA   \cite{hu2021lora, devalal2018lora, sundaram2019survey, bor2016lora} or adapter tuning   \cite{lu2023uniadapter, he2021effectiveness, zhang2023llama, le2021lightweight} have been successfully applied in models such as RT-2   \cite{brohan2023rt, zitkovich2023rt} and SayCan   \cite{ahn2022can}, where robots are guided by high-level language commands grounded in visual context. PEFT also facilitates domain adaptation—e.g., transferring policies trained in simulation to the real world (sim-to-real)—by fine-tuning lightweight modules on real-world data without the need to retrain the entire policy network   \cite{enroth2011policy, dowding1995model, atkinson1992policy, peterson2003policy,
wright1988policy, klijn1997policy,
rhodes2006policy}.

In embodied AI, where agents interact with their surroundings via sensors and actuators (e.g., navigation, manipulation, object fetching), PEFT allows for task- or goal-specific adaptation by introducing small trainable components into large pretrained transformers or diffusion policies   \cite{hou2024diffusion, gupta2024pre, 
dasari2024ingredients, reuss2023multimodal, reuss2024multimodal}. These approaches help maintain generalization across environments while allowing fast adaptation to new robotic skills with limited data. Moreover, in multi-robot or multi-task scenarios, PEFT promotes modularity and parameter sharing, enabling efficient scalability across hardware platforms and task sets.

 As illustrated in \textbf{Figure \ref{fig:Robotic}}, although current implementations like PaLM-E   \cite{driess2023palm} are not explicitly PEFT-based, their architecture---a frozen LLM processing tokenized streams of language prompts, visual inputs, and proprioceptive data---lends itself well to PEFT augmentation. The figure demonstrates various real-world robotic applications including mobile manipulation, tabletop tasks, and motion planning, all unified through a central language model. In such systems, PEFT can be applied through LoRA (Low-Rank Adaptation) modules designed for specific output heads, or by introducing lightweight adapters that facilitate fusion of proprioceptive information. These adaptations would allow the robot to learn task-specific motor commands, such as ``grasp,'' ``rotate,'' or ``navigate,'' with high efficiency and without retraining the entire network. This is particularly advantageous in robotic environments, where data is sparse and tasks are dynamic, requiring continual learning without compromising previously acquired behaviors. Through the integration of PEFT modules, robotics systems can achieve lifelong learning capabilities, extending their utility across diverse operational contexts with minimal retraining cost.

\begin{table*}[htbp]
\centering
\resizebox{\textwidth}{!}{
\begin{tabular}{llcccc}
\toprule
\textbf{Category} & \textbf{Method} & \textbf{Space Complexity} & \textbf{Time Complexity} & \textbf{TTPs} & \textbf{APs} \\
\midrule
\textbf{Full Fine-Tuning} & FT (Full Fine-Tuning) & $O(d \times d)$ & $O(d \times d)$ & $d^2$ & 0 \\
\midrule
\textbf{Adapter-Based Fine-Tuning} & (IA)$^3$ & $O(l_k + l_v + l_{ff})$ & $O(d_k + d_v + d_{ff})$ & $3d$ & $3d$ \\
\midrule
\textbf{Soft Prompt-Based} & Prompt & $O(d \times l_p)$ & $O(d \times l_p)$ & $l_p d$ & $l_p d$ \\
\textbf{Soft Prompt-Based} & Prefix & $O(L \times d \times l_p)$ & $O(L \times d \times l_p)$ & $L l_p d$ & $L l_p d$ \\
\midrule
\textbf{Structured Fine-Tuning} & LoRA & $O((d + d) \times r)$ & $O((d + d) \times r)$ & $2dr$ & $dr$ \\
\textbf{Structured Fine-Tuning} & LoRA-FA & $O((d + d) \times r)$ & $O((d + d) \times r)$ & $dr$ & $2dr$ \\
\midrule
\textbf{Adaptive Rank Methods} & AdaLoRA & $O((d + d + r) \times r)$ & $O((d + d + r) \times r)$ & $2dr + r^2$ & $2dr + r^2$ \\
\midrule
\textbf{Hybrid Approach} & LoHA & $O(2r \times (d + d))$ & $O(2r \times (d + d))$ & $4dr$ & $4dr$ \\
\midrule
\textbf{Low Rank Decomposition} & RoCoFT (Row) & $O(d \times r)$ & $O(d \times r)$ & $rd$ & 0 \\
\textbf{Low Rank Decomposition} & RoCoFT (Column) & $O(d \times r)$ & $O(d \times r)$ & $rd$ & 0 \\
\midrule
\textbf{Scaling Adaptation} & Propulsion & $O(d)$ & $O(d)$ & $d$ & $d$ \\
\bottomrule
\end{tabular}}
\caption{Comparison of PEFT methods and their computational complexity. Here, \textbf{TTPs} refers to the \textit{total trainable parameters}, and \textbf{APs} refers to the \textit{additional parameters} introduced by the fine-tuning method.}
\label{tab:complexity}
\end{table*}

\begin{table*}[htp!]
\centering
\scalebox{0.75}{
\begin{tabular}{l|l|p{10cm}}
\hline
\textbf{Method}        & \textbf{$\Delta W$ Reparameterization} & \textbf{Notes} \\ \hline
Intrinsic SAID         & $\Delta W  = F(W^r)$   & F : $\mathbb{R}^r \rightarrow \mathbb{R}^d$, $W^r \in \mathbb{R}^r$ are parameters to be optimized, and $r \ll d$. \\ \hline
LoRA                   & $\Delta W = W_{\text{down}} W_{\text{up}}$  & $W_{\text{down}} \in \mathbb{R}^{d \times r}$, $W_{\text{up}} \in \mathbb{R}^{r \times d}$, and $r \ll \{k,d\}$. \\ \hline
KronA                  & $\Delta W = W_{\text{down}} \otimes W_{\text{up}}$ & rank($W_{\text{down}} \otimes W_{\text{up}}$) = rank($W_{\text{down}}$) $\times$ rank($W_{\text{up}}$). \\ \hline
DyLoRA                 & $\Delta W = W_{\text{down}\downarrow b} W_{\text{up}\downarrow b}$ & $W_{\text{down}\downarrow b} = W_{\text{down}}[:b,:], \quad W_{\text{up}\downarrow b} = W_{\text{up}}[:, :b], \quad b \in \{r_{\text{min}}, \dots, r_{\text{max}}\}$. \\ \hline
AdaLoRA                & $\Delta W  = PAQ$ & $PP^\top = P^\top P \neq I = QQ^\top = Q^\top Q$, $\Lambda = \text{diag}(\sigma_1, \sigma_2, \dots, \sigma_r)$. \\ \hline
IncreLoRA              & $\Delta W = W_{\text{down}} \Lambda W_{\text{up}}$ & $\Lambda = [\lambda_1, \lambda_2, \dots, \lambda_r]$ with $\lambda_i$ being an arbitrary constant. \\ \hline
DeltaLoRA              & $\Delta W = W_{\text{down}} W_{\text{up}}$ & $W^{(t+1)} \gets W^{(t)} + \big(W_{\text{down}}^{(t+1)} W_{\text{up}}^{(t+1)} - W_{\text{down}}^{(t)} W_{\text{up}}^{(t)}\big)$. \\ \hline
LoRAPrune              & $\Delta W = W_{\text{down}} W_{\text{up}} \odot M$ & $\delta = (W + W_{\text{down}} W_{\text{up}}) \odot M, \quad M \in \{0, 1\}^{1 \times G}, \quad G \text{ is group number.}$ \\ \hline
QLoRA                  & $\Delta W = W_{\text{down}}^{BF16} W_{\text{up}}^{BF16}$ & $Y^{BF16} = X^{BF16} \cdot \text{doubleDequant}(c_1^{FP32}, c_2^{FP8}, W^{NF4}) 
+ X^{BF16} W_{\text{down}}^{BF16} W_{\text{up}}^{BF16}$. \\ \hline
QA-LoRA                & $\Delta W = W_{\text{down}} W_{\text{up}}$ & $W_{\text{down}} \in \mathbb{R}^{d \times r}$, $W_{\text{up}} \in \mathbb{R}^{r \times L}$, $L$ is the quantization group number of W. \\ \hline
LoFTQ                  & $\Delta W = SVD(W - Q_t$) & $Q_t = q_N \big(W - W_{\text{down}}^{t-1} W_{\text{up}}^{t-1} \big), \quad q_N \text{ is } N\text{-bit quantization function.}$ \\ \hline
Kernel-mix             & $\Delta W^h = \big(B_{\text{LoRA}}, B^h\big) 
\begin{pmatrix}
A_{\text{LoRA}}^h \\ 
A^h
\end{pmatrix}$ & $B_{\text{LoRA}} \text{ is shared across all heads, } B^h, A^h \text{ provide rank-}r  \text{ update  in each head.}$. \\ \hline
LoRA-FA                & $\Delta W = W_{\text{down}} W_{\text{up}} = Q R W_{\text{up}}$ & $W_{\text{down}}$ is frozen, and only $W_{\text{up}}$ is updated. \\ \hline
RoCoFT            & 
$\begin{array}{c} 
\Delta W = W_0 + R  \nonumber\\

\Delta W = W_0 + C
\end{array}$
& $R$ and $C$ are restricted weight matrices such that only at most $r$ of the rows or columns are non-zero.\\ \hline
Propulsion               & $\Delta W = (WX)\odot Z$ & $W$ is frozen, $X$ is input, and  $Z$ is propulsion parameters. \\ \hline
\end{tabular}
}
\caption{Comparison of delta weight reparameterization across various PEFT methods. Representations of the baseline methods are taken from  \cite{xu2023parameter}.} \label{tab:delta_cmp}
\end{table*}

\section{Complexity of PEFT Methods}
\label{sec:complexitypeft}

Table~\ref{tab:complexity} provides a detailed comparison of various PEFT methods based on their space and time complexity, as well as the total number of trainable parameters (TTPs) and additional parameters (APs) introduced during fine-tuning. Traditional full fine-tuning (FT) modifies all parameters of the model, resulting in a quadratic complexity of $O(d \times d)$ in both space and time, with a high memory footprint and zero additional modularity.

Adapter-based methods such as (IA)$^3$ reduce the fine-tuning burden by introducing small modules within the transformer layers, yielding linear complexity and maintaining trainable parameter counts at $3d$. Soft prompt-based methods, like Prompt and Prefix tuning, encode task-specific knowledge into learnable token embeddings, with complexity tied to the prompt length $l_p$ and number of layers $L$. These methods allow for highly modular adaptation while keeping training costs manageable.

Structured fine-tuning approaches like LoRA and its variant LoRA-FA factorize weight updates into low-rank matrices, reducing the number of trainable parameters to $O(dr)$ or less. Adaptive rank methods, such as AdaLoRA, dynamically adjust the rank during training, offering a flexible trade-off between performance and efficiency, though at a slightly higher parameter count due to the inclusion of $r^2$ terms.

Hybrid approaches like LoHA further extend the expressiveness of LoRA by introducing hierarchical adaptation, doubling the parameter footprint ($4dr$) in exchange for better task generalization. Similarly, RoCoFT applies a low-rank decomposition at either the row or column level of weight matrices, maintaining very low complexity ($O(d \times r)$) with no additional overhead beyond trainable parameters.

Finally, Propulsion represents an extremely lightweight and scalable fine-tuning mechanism, introducing only $O(d)$ space and time complexity, with both TTPs and APs capped at $d$. This makes it particularly attractive for edge and low-resource deployment.

In  addtion, Table \ref{tab:delta_cmp} provides a comprehensive comparison of various PEFT methods based on their reparameterization of the delta weight matrix \( \Delta W \). Each method uses different strategies for adjusting the weight updates during fine-tuning, optimizing parameter efficiency while maintaining performance. 

Overall, the table illustrates the diverse trade-offs between efficiency, modularity, and expressivity across PEFT techniques, offering a toolkit of strategies tailored to specific deployment constraints and task complexities.

\section{Strengths and Weaknesses of PEFT}
\label{sec:strengths}

PEFT has emerged as a transformative approach in adapting large pre-trained models to downstream tasks, offering a compelling balance between computational efficiency  \cite{kowsher2024rocoft, hu2021lora, zhang2023adalora, prottasha2024parameter, hayou2024lora+} and task-specific performance. One of its primary strengths lies in its ability to significantly reduce computational and memory costs by updating only a small subset of the model's parameters or introducing lightweight adapters, making it feasible to fine-tune large models on resource-constrained hardware  \cite{kowsher2024propulsion, hu2024adafish, zaken2021bitfit}. This efficiency extends to faster training times and lower energy consumption, which is particularly advantageous in environmentally conscious applications. Additionally, PEFT mitigates the risk of catastrophic forgetting by preserving the general knowledge encoded in the pre-trained model, while still enabling effective transfer learning, especially in low-data regimes  \cite{kowsher2024propulsion, zaken2021bitfit}.

However, PEFT is not without its limitations. It may underperform in tasks requiring significant adaptation of the pre-trained model, as the constraints imposed by limited parameter updates can restrict the model's ability to fully capture task-specific nuances  \cite{liu2024moe, lin2024peft, zhou2024autopeft}. Furthermore, some PEFT methods introduce architectural complexity, making implementation and debugging more challenging compared to standard fine-tuning. The approach can also be sensitive to hyperparameters, such as the size of adapter layers or the rank of low-rank approximations, necessitating extensive experimentation to achieve optimal performance  \cite{lu2023uniadapter, zheng2021adapting}. Additionally, PEFT may struggle with tasks that require a drastic shift from the pre-training Application, as it is most effective when the downstream task is closely related to the original training data  \cite{pfeiffer2020adapterhub}.

Despite these challenges, PEFT remains a powerful tool for scaling large models across diverse applications, and ongoing research aims to address its limitations, such as improving flexibility for diverse tasks and reducing hyperparameter sensitivity, to further enhance its utility in the field of machine learning.

\section{Discussion}
\label{sec:discussion}

Despite the remarkable progress of PEFT techniques in reducing computational and memory demands for adapting large language and vision models, several pressing challenges remain unresolved. Current methods often rely on heuristics rather than principled understanding, leading to inconsistent performance across tasks, architectures, and modalities. The lack of theoretical grounding regarding parameter sensitivity, the opaque nature of learned prompts and adapter modules, and the absence of unified benchmarks hinder reproducibility and generalization. Moreover, most PEFT approaches operate without consideration of task structure, domain knowledge, or semantic alignment—resulting in adaptations that, while efficient, are often suboptimal or cognitively naive. These limitations highlight the need for deeper analysis of model internals, architecture-aware design, and standardized evaluation to realize the full potential of PEFT in real-world, multimodal, and evolving scenarios.

\section{Future Research Directions}
\label{sec:futuredirections}

PEFT methods have emerged as essential tools for adapting large-scale foundation models under computational and storage constraints, the current trajectory of research reveals several key areas where further investigation is both necessary and promising. These directions are outlined below to guide the evolution of PEFT toward greater generalizability, robustness, and theoretical maturity.

\subsection{ Theoretical Understanding of Parameter Influence}

Most PEFT methods are grounded in empirical success rather than analytical rigor. Future research must prioritize the development of theoretical frameworks that explain how small subsets of trainable parameters influence overall model adaptation. Concepts from information theory, such as mutual information between adapted modules and output prediction, or from optimization theory, such as curvature of the loss landscape around modular updates, could be leveraged to quantify adaptation efficiency. A better theoretical grounding would not only enhance interpretability but also inform principled design choices across diverse architectures.

\subsection{ Layer-wise Sensitivity and Structural Adaptation}

In transformer-based architectures, not all layers contribute equally to downstream task performance. Existing PEFT approaches often insert adapter modules or low-rank projections uniformly across layers, which may be suboptimal. Future work should explore sensitivity-based placement strategies—using tools such as Jacobian analysis or Fisher Information Matrix estimates—to identify layers where fine-tuning yields the highest performance-to-parameter ratio. Additionally, research should focus on adaptive placement strategies, where modules are dynamically activated based on input complexity or layer activation statistics during training.

\subsection{ Task-Aware and Domain-Specific PEFT}

While current PEFT methods are generally task-agnostic, real-world applications often involve domain-specific constraints and task structures. For example, tasks in legal or medical NLP involve complex semantic dependencies, while vision tasks in robotics may require temporally aligned fine-tuning. Future PEFT frameworks should incorporate inductive biases tailored to task semantics, perhaps through structure-aware prompts, hierarchical adapters, or task-conditioned reparameterization schemes. Integrating symbolic reasoning elements, causal graphs, or domain ontologies may also enhance generalization in low-data or high-stakes scenarios.

\subsection{Generalization to Multimodal and Non-Transformer Architectures}

Most PEFT techniques have been developed and tested primarily on large transformer-based LLMs. However, an increasing number of vision models (e.g., CNN-Transformer hybrids) and multimodal architectures (e.g., CLIP, Flamingo, Gato) demand adaptation strategies that account for modality entanglement, asynchronous inputs, and stream-wise attention fusion. Future research should design PEFT modules that maintain cross-modal coherence, minimize information bottlenecks, and support modality-specific adaptation while preserving inter-modal alignment. Exploration of fine-tuning strategies for non-transformer backbones, such as graph neural networks or recurrent models, also remains largely uncharted.

\subsection{Continual and Lifelong Learning Integration}

PEFT methods are typically designed for static, single-task adaptation. However, real-world environments demand continual adaptation to evolving tasks and distributions. Incorporating lifelong learning principles—such as replay-based memory modules, regularization-based knowledge retention, or dynamically growing parameter banks—into PEFT frameworks would enable more resilient and context-aware models. Sparse adapter stacking, delta compression, and orthogonal subspace training are promising avenues for enabling memory-efficient continual PEFT without catastrophic forgetting.

\subsection{Interpretability and Explainability of PEFT Modules}

The modular nature of PEFT methods presents an opportunity for improved interpretability, yet this potential remains underexploited. Few studies have systematically investigated what adapter layers or learned prompts actually encode. Future work should develop attribution techniques and visualization tools to trace the flow of information through PEFT modules. Interpretable tuning may involve aligning adapter activations with human-understandable concepts, analyzing prompt token behavior across tasks, or quantifying attention shifts induced by fine-tuning. Such developments are particularly crucial in applications where explainability is legally or ethically mandated.

\subsection{Privacy-Preserving and Federated PEFT}

The intersection of PEFT with differential privacy and federated learning is a promising but underdeveloped area. Given the proliferation of LLM deployment in privacy-sensitive contexts—such as healthcare, finance, and education—future research must explore methods to fine-tune models without centralized data access. Approaches like differentially private LoRA, secure adapter aggregation, or decentralized prompt tuning may offer viable paths forward. These methods should aim to maintain fine-tuning efficiency while rigorously protecting user data and ensuring compliance with regulatory standards.

\subsection{Standardization of Benchmarks and Evaluation Protocols}

There is an urgent need for standardized, multimodal benchmark suites designed specifically for evaluating PEFT methods. These should span diverse task types (e.g., classification, generation, reasoning), data regimes (low-resource, zero-shot, few-shot), and domains (general-purpose, biomedical, legal). Additionally, evaluation protocols should include robustness tests under domain shift, noise injection, and adversarial perturbations. Establishing such benchmarks will enhance reproducibility, allow fair comparisons, and accelerate the iterative improvement of PEFT methodologies.

\subsection{Hardware-Aware and Sustainable PEFT}

As AI systems are increasingly deployed on edge devices and in environmentally sensitive settings, PEFT research must align with the goals of hardware-awareness and energy efficiency. Techniques should be optimized for specific accelerators (e.g., TPUs, NPUs, FPGAs), and evaluated not only on accuracy and parameter count but also on latency, power consumption, and carbon footprint. Green AI practices, including low-bit quantized PEFT modules or adaptive update schedules that terminate early on easy samples, may contribute to more sustainable large-scale model use.

\subsection{Meta-PEFT: Learning to Tune Efficiently}

A promising meta-direction involves designing systems that automatically learn how to fine-tune models efficiently. Meta-PEFT approaches may employ reinforcement learning, neural architecture search, or gradient-based meta-learning to discover optimal PEFT strategies across tasks and models. This could lead to generalizable policies for adapter placement, prompt design, or rank selection, significantly reducing manual trial-and-error and improving portability across diverse domains.

\section{Conclusion}
\label{sec:conclusion}

As the scale and ubiquity of large language, vision, and multimodal models continue to expand, the demand for computationally efficient and scalable fine-tuning strategies has become increasingly urgent. PEFT techniques have emerged as a pragmatic and powerful response to these demands, enabling adaptation of large-scale models to diverse downstream tasks with minimal additional resource overhead. This survey has provided a comprehensive synthesis of PEFT methodologies, categorizing them into additive, selective, reparameterized, hybrid, and unified frameworks. By analyzing their design principles, parameter behaviors, and architectural integration, we have highlighted the core mechanisms that underlie their effectiveness.

We have also demonstrated the broad applicability of PEFT methods across language processing, visual understanding, and generative modeling, emphasizing how these strategies bridge the gap between performance and efficiency. Moreover, we have identified critical challenges in areas such as interpretability, task generalization, continual learning, and theoretical grounding. Addressing these challenges will be essential for building adaptive, robust, and sustainable AI systems.

Looking forward, the role of PEFT is poised to become even more central in future AI development—particularly in settings where privacy, environmental constraints, or domain specificity limit the feasibility of traditional fine-tuning. By distilling the current landscape and charting key research directions, this work aims to serve as a foundational reference for researchers and practitioners committed to advancing efficient, equitable, and accessible model adaptation in the era of foundation models.

\bibliographystyle{plain}

\bibliography{main}


\appendix
\newpage

\begin{table}[htbp]
\centering

\scriptsize
\setlength{\tabcolsep}{2pt}
\begin{tabular}{@{}lclllrr@{}}
\toprule
\textbf{Dataset} & \textbf{Domain} & \textbf{Year} & \textbf{Task Type} & \textbf{I/O Format} & \textbf{Size} & \textbf{Train/Test} \\
\midrule

20 Newsgroups  \cite{twenty_newsgroups_113} & NLP & 1999 & Text Classification & Text $\rightarrow$ Category & 20K & 11K / 7.5K \\
CoNLL03  \cite{DBLP:journals/corr/cs-CL-0306050} & NLP & 2003 & NER & Text $\rightarrow$ Entity tags & 2,302 & 946 / 231 \\
CoNLL04  \cite{roth2004linear} & NLP & 2004 & Relation extraction & Text $\rightarrow$ BIO tags & 1,437 & 8,936 / 1,671 \\
Caltech101  \cite{fei2004learning} & CV & 2004 & Image Classification & Images $\rightarrow$ Class & 9,144 & - \\
SST-2  \cite{socher2013recursive} & NLP & 2005 & Sentiment Analysis & Text $\rightarrow$ Binary & 215K & 67K / 1.8K \\
MRPC  \cite{dolan-brockett-2005-automatically} & NLP & 2005 & Paraphrase Detection & Sentence pairs $\rightarrow$ Binary & 5,800 & 3,669 / 1,724 \\
ACE2005  \cite{walker2006ace} & NLP & 2005 & Information Extraction & Text $\rightarrow$ Entities & 599 & 529 / 40 \\
\bottomrule
\end{tabular}
\caption{Comprehensive overview of datasets used for PEFT methods across various domains including NLP, Computer Vision, Multimodal, and Language Models.}
\label{tab:dataset-overview}
\end{table}


\begin{table}[htbp]
\centering

\scriptsize
\setlength{\tabcolsep}{2pt}
\begin{tabular}{@{}lclllrr@{}}
\toprule
\textbf{Dataset} & \textbf{Domain} & \textbf{Year} & \textbf{Task Type} & \textbf{I/O Format} & \textbf{Size} & \textbf{Train/Test} \\
\midrule

ro-en & MT & 2016 & Machine Translation & Romanian $\rightarrow$ English & 614K & 610K / 2K / 2K \\
de-en & MT & 2014 & Machine Translation & German $\rightarrow$ English & 4.55M & 4.54M / 3K / 2.2K \\
Ha–En & MT & 2021 & Machine Translation & Hausa $\rightarrow$ English & 50K & 45K / 2.5K / 2.5K \\
(De-En) & MT & 2019 & Machine Translation & German $\rightarrow$ English & 3.9M & 3.9M / 3K / 3K \\
Fr–De & MT & 2020 & Machine Translation & French $\rightarrow$ German & 2.8M & 2.8M / 3K / 3K \\
Fr–Es & MT & 2018 & Machine Translation & French $\rightarrow$ Spanish & 2M & 2M / 3K / 3K \\
News Commentary \cite{tiedemann2012parallel} & MT & 2017 & Machine Translation & Multilingual $\rightarrow$ Translations & 370K & 90\% / 5\% / 5\% \\
\bottomrule
\end{tabular}
\caption{Comprehensive overview of datasets used for PEFT methods across various domains, including NLP, Computer Vision, Multimodal, and Language Models.}
\label{tab:dataset-overview1}
\end{table}

\begin{table}[htbp]
\centering

\scriptsize
\setlength{\tabcolsep}{2pt}
\begin{tabular}{@{}lclllrr@{}}
\toprule
\textbf{Dataset} & \textbf{Domain} & \textbf{Year} & \textbf{Task Type} & \textbf{I/O Format} & \textbf{Size} & \textbf{Train/Test} \\
\midrule

Kinetics-400 \cite{kay2017kinetics} & CV & 2017 & Action Class. & Video clips $\rightarrow$ 400 labels & 650K & - \\
Something-v2 \cite{goyal2017something} & CV & 2017 & Action Class. & Short video clips $\rightarrow$ 174 labels & 221K & 169K / 27K \\
VisDA-C \cite{peng2017visda} & CV & 2017 & Domain Adapt. & Images $\rightarrow$ Domain Class & 280K & 152K / 55K \\
EuroSAT \cite{helber2019eurosat} & CV & 2018 & Land Use Class. & Satellite images (64x64) $\rightarrow$ 10 labels & 27K & 24.3K / 2.7K \\
iNaturalist \cite{bateux2024improving} & CV & 2018 & Species Class. & Wildlife images $\rightarrow$ 8,142 species & ~612K & 437K / 149K \\
Taskonomy \cite{zamir2018taskonomy} & CV & 2018 & Multi-Task Vision & Indoor images $\rightarrow$ depth, seg., normals & ~4M & ~4M \\
ScanObjectNN \cite{uy2019scanobjectnn} & CV & 2019 & 3D Obj. Class. & Point clouds $\rightarrow$ 15 labels & ~15K & 11.4K \\
\bottomrule
\end{tabular}
\caption{Comprehensive overview of datasets used for PEFT methods across various domains including NLP, Computer Vision, Multimodal, and Language Models.}
\label{tab:dataset-overview2}
\end{table}

\begin{table}[t!]
\centering
\scriptsize

\setlength{\tabcolsep}{2pt}
\newcolumntype{n}[1]{>{\raggedright\arraybackslash}p{#1}}
\begin{tabular}{@{}l n{2.8cm} n{3.4cm} n{3.6cm} n{3.2cm}@{}}
\toprule
\textbf{Year} & \textbf{Model} & \textbf{Application} & \textbf{PEFT Method} & \textbf{Datasets} \\
\midrule

2025 \citep{thomas2025exploring} & Mistral 7B  \citep{jiang2023mistral} & 
Code conversion system & 
LoRA & 
APPS, Conala, CodeAlpacaPy \\

2025 \citep{ghasemi2025multilingual}  & BERT  \citep{devlin2019bert}, mT5  \citep{2020t5}, mGPT  \citep{achiam2023gpt} & 
Low-resource sentiment analysis, NER, QA & 
Full-Finetune, LoRA, AdaLoRA, DoRA & 
Persian NLP \\

2025 \citep{jia2025task}  & GLM4-9B  \citep{du2021glm} & 
Language understanding, sentiment analysis & 
MoE-LoRA, X-LoRA, TAAS-Net, IA\textsuperscript{3}, LoRA & 
GLUE, IMDB, Agnews \\

2025 \citep{gajulamandyam2025domain} & LLaMA  \citep{touvron2023llama}, Gemma  \citep{team2024gemma}, GPT-4o  \citep{achiam2023gpt} & 
Immigration law and insurance & 
LoRA, QLoRA, DoRA, Prompt Tuning & 
InsuranceQA, USCIS data  \\

2025 \citep{ali2025learning} & DistilBERT  \citep{sanh2019distilbert}, RoBERTa  \citep{liu2019roberta}, LLaMa-7B  \citep{touvron2023llama} & 
Common sense reasoning & 
LoRAShear, LLM-Pruner, LoRAPrune, LoRA variants & 
BoolQ, PIQA, HellaSwag  \\

2024 \citep{das2024natural}  & LLaMA  \citep{touvron2023llama}, RoBERTa  \citep{liu2019roberta}, TinyLlama  \citep{zhang2024tinyllama} & 
Token prediction, pre-training & 
Full-Rank, GaLore variants, LoRA, ReLoRA & 
C4, GLUE, TinyAgent \\

2024 \citep{singh2024three} & Llama-2  \citep{touvron2023llama}, GIZA++  \citep{och2001giza++}, xlm-roberta  \citep{conneau2019unsupervised} & 
Cross-lingual transfer & 
Handholding ICL/PEFT variants & 
Amazon Massive \\
\bottomrule
\end{tabular}
\caption{Parameter-Efficient Fine-Tuning (PEFT) Methods in NLP Models (2024-2025)}
\label{tab:peft_methods_nlp1}
\end{table}

\begin{table}[htbp]
\centering
\scriptsize
\setlength{\tabcolsep}{2pt}
\newcolumntype{q}[1]{>{\raggedright\arraybackslash}p{#1}}
\begin{tabular}{@{}l q{2.8cm} q{3.4cm} q{3.6cm} q{3.2cm}@{}}
\toprule
\textbf{Year} & \textbf{Model} & \textbf{Application} & \textbf{PEFT Method} & \textbf{Datasets} \\
\midrule

2024 \cite{xie2024pedro} & LlaMA-2 7B \cite{touvron2023llama}, LlaMA-2 13B  \cite{touvron2023llama}, Gemma 2B \cite{team2024gemma} & Question-answering, sentence level tasks, instruction tuning & PEDRO, BitFit, IA\textsuperscript{3}, SSP, AdaLoRA, LoRA, Learned-Adapter, Housbly-Adapter, LPT, P-tuning v2, Full-FT & SQuAD, SuperGLUE (BoolQ, COPA, and ReCoRD), GLUE (SST-2, RTE, QNLI), Alpaca dataset, MT-Bench, MMLU, BBH \\

2024 \cite{miao2024flexllm} & OPT-13B \cite{zhang2022opt}, LLaMA30B, LLaMA-2-70B  \cite{touvron2023llama} & Co-serving system for token generation & HuggingFace (HF) PEFT, S-LoRA, LoRA, IA\textsuperscript{3}, Adapter & Chatbot instruction prompts, ChatGPT Prompts, WebQA, Alpaca, PIQA \\

2024 \cite{runwal2024peft} & RoBERTa\textsubscript{Large} \cite{liu2019roberta}, BERT\textsubscript{BASE} \cite{devlin2019bert} \cite{koroteev2021bert}, T5\textsubscript{SMALL}, T5\textsubscript{BASE} \cite{2020t5}, Flan-T5-base, Flan-T5-xl \cite{chung2024scaling}, OPT \cite{zhang2022opt}, GPT2 \cite{liu2024gpt} \cite{zheng2021adapting} \cite{liu2024gpt}, ViT \cite{dosovitskiy2020image} & Sentiment classification, paraphrase detection, natural language inference, linguistic acceptability, semantic textual similarity, question answering & DEFT, ADA-DEFT, PEFT, ADA-PEFT, LoRA, Adapter, Prefix-Tuning, Prompt-Tuning & MNLI, QQP, QNLI, SST-2, STS-B, MRPC, RTE, SQuAD \\

2024 \cite{gao2024dlora} & OPT6.7B \cite{zhang2022opt}, BLOOM-7B \cite{luccioni2023estimating}, LLaMA-7B  \cite{touvron2023llama} & QA, multichoice science questions, problem compilation and concluding tasks & DLoRA, FT & OBQA, PIQA, SIQA, Winograde, BoolQ, HellaSwag, ARC-easy, ARC-challenge \\
\bottomrule
\end{tabular}
\caption{Parameter-Efficient Fine-Tuning (PEFT) Methods in NLP Models (2024-2025)}
\label{tab:peft_methods_nlp}
\end{table}

\begin{table}[t!]
\centering
\scriptsize

\setlength{\tabcolsep}{2pt}
\newcolumntype{z}[1]{>{\raggedright\arraybackslash}p{#1}}
\begin{tabular}{@{}l z{2.8cm} z{3.4cm} z{3.6cm} z{3.2cm}@{}}
\toprule
\textbf{Year} & \textbf{Model} & \textbf{Application} & \textbf{PEFT Method} & \textbf{Datasets} \\
\midrule
2023 \cite{wang2023aprompt} & T5-Base, T5-Large, T5-XL \cite{2020t5} & Natural language understanding tasks & FT, PT, Prompt Tuning, XPrompt, ResPrompt, APROMPT & SuperGLUE\\

2023 \cite{chen2023ptp} & BERT \cite{koroteev2021bert}, RoBERTa \cite{liu2019roberta}, ALBERTAxxlarge-v2 \cite{ryu2021re} & Natural language understanding tasks in few shot and fully supervised settings & P-Tuning, P-Tuning v2, PTP+RM, PTP+RG, PTP+A2T, PTP+PGD, PET Best & SuperGLUE, FewGLUE \\

2023 \cite{choi2023smop} & T5 \cite{2020t5} & Natural language understanding & Full FT, Prompt Tuning, P-Tuning, SMoP & SuperGLUE \\

2023 \cite{shi2023dept} & T5-Base \cite{2020t5}, CLIP-T5\citep{radford2021learning} & Natural language processing (NLP), visual question-answering task, image caption generation & LoRA, FT, Adapters, BitFit, PT, SPoT, ATTEMPT, MPT, DEPT & GLUE, SuperGLUE, WinoGrande, Yelp-2, SciTail, PAWS-Wiki, VQA, MSCOCO \\

2023 \cite{zadouri2023pushing} & T5 v1.1 + LM adaptation & Instruction tuning & T0, (IA)\textsuperscript{3}, LoRA, MoV, MoLoRA & ANLI, CB, RTE, WSC, WiC, COPA, WNG, HS \\

2023 \cite{lu2023inference} & GPT-2, GPT-3, GPT-3.5, GPT-4 \cite{radford2018improving}\cite{liu2024gpt}\cite{achiam2023gpt}, DialoGPT \cite{zhang2019dialogpt}, GODEL \cite{peng2022godel}, Blenderbot \cite{shuster2022blenderbot}, ChatGPT, DOHA \cite{prabhumoye2021focused}, T5 \cite{2020t5} & Toxicity reduction, lexically constrained generation, open-ended generation, dialogue safety control, and knowledge grounded dialogue & IPA, IPA\textsuperscript{*}, PPLM, GeDi, DEXPERTS, DAPT, PPO, QUARK & RealToxicityPrompts, XSum, CommonGen, ASAFETY, WoW \\

2023 \cite{lawton2023neural} & RoBERTa-Large \cite{liu2019roberta} & Natural language understanding & LoRA, FFT, BitFit, Adapters, MaM, S-MaM, U-MaM, WARP, S-BitFit, U-BitFit & GLUE \\
\bottomrule
\end{tabular}
\caption{Parameter-Efficient Fine-Tuning (PEFT) Methods in NLP Models (2023)}
\label{tab:peft_methods_nlp_combined2}
\end{table}

\begin{table*}[t!]
\centering
\scriptsize
\setlength{\tabcolsep}{2pt}
\newcolumntype{y}[1]{>{\raggedright\arraybackslash}p{#1}}
\begin{tabular}{@{}l y{2.8cm} y{3.4cm} y{3.6cm} y{3.2cm}@{}}
\toprule
\textbf{Year} & \textbf{Model} & \textbf{Application} & \textbf{PEFT Method} & \textbf{Datasets} \\
\midrule

2023 \cite{zhou2024autopeft} & BERTbase \cite{koroteev2021bert} & Text classification tasks & LoRA, FFT, Prefix, AdaMix, Serial, UniPELT, Parallel, MAM, AutoPEFT & GLUE, SuperGLUE \\

2023 \cite{hu2023llm} & LLaMA \cite{touvron2023llama}, BLOOMz \cite{nair2024investigating}, GPT-J \cite{chuang2024use} & Arithmetic reasoning and
commonsense reasoning & LoRA, Prefix, Series, Parallel & MultiArith, GSM8K, AddSub, AQuA, SingleEq, SVAMP, BoolQ, PIQA, SIQA, HellaSwag, WinoGrande, ARC-e, ARC-c, OBQA \\

2023 \cite{zhao2023prototype} & T5-Base, T5-Small, T5-Large \cite{2020t5} & Multi-task learning for natural language understanding & Adapter, FT, PT, SPoT, HF, ATP, HD, MPT, PHA & CoLA, SST-2, STS-B, MRPC, QQP, MNLI, QNLI, RTE, SciTail, BoolQ, WiC, CB, WSC \\

2023 \cite{xu2023fwdllm} & ALBERT \cite{lan2019albert}, DistilBERT-base \cite{sanh2019distilbert}, BERT-base \cite{devlin2019bert}, RoBERTa-large \cite{liu2019roberta}, LLaMA-7B (INT4) \cite{touvron2023llama} & News classification, topic classification, QA using federated fine-tuning of LLMs & LoRA, Adapter, BitFit, Full-FT & AGNEWS, YAHOO, YELP-Polarity, SQuAD-v1.1 \\

2022 \cite{ma2022xprompt} & T5-Large, T5-XL, and T5-XXL \cite{2020t5} & Natural language understanding & Fine-Tuning, P-Tuning, Prefix-Tuning, Prompt-Tuning, XPROMPT & SuperGLUE \\

2022 \cite{wu2022idpg} & RoBERTa \cite{liu2019roberta} and EFL \cite{wang2021entailment} & Natural language understanding & Transformer fine-tuning, Prompt tuning, Adapter, S-IDPG-PHM, S-IDPG-DNN, M-IDPG-PHM-GloV, M-IDPG-PHM, M-IDPG-DNN, Compacter, P-Tuningv2 & MPQA, Subj, CR, MR, SST-2, QNLI, RTE, MRPC, STS-B, QQP \\
\bottomrule
\end{tabular}
\caption{Parameter-Efficient Fine-Tuning (PEFT) Methods in NLP Models (2021-2023)}
\label{tab:peft_methods_nlp_combined3}
\end{table*}

\begin{table}[t!]
\centering
\scriptsize
\setlength{\tabcolsep}{2pt}
\newcolumntype{X}[1]{>{\raggedright\arraybackslash}p{#1}}
\begin{tabular}{@{}l X{2.8cm} X{3.4cm} X{3.6cm} X{3.2cm}@{}}
\toprule
\textbf{Year} & \textbf{Model} & \textbf{Application} & \textbf{PEFT Method} & \textbf{Datasets} \\
\midrule
2021 \cite{ansell2021composable} & mBERT \cite{xu2021bert} \cite{devlin2019bert}, XLM-R \cite{conneau2019unsupervised} & Zero-shot cross-lingual transfer & LT-SFT, RAND-SFT, MAD-X, BITFIT, LT-SFT TA-ONLY, MAD-X TA-ONLY & Universal Dependencies 2.7, MasakhaNER, CoNLL 2003, AmericasNLI, MultiNLI \\

2021 \cite{mahabadi2021parameter} & T5 \cite{2020t5} & Natural language understanding & HYPERFORMER, Adapters & GLUE \\
2021 \cite{xu2021raise} & BERTLARGE \cite{devlin2019bert}, XLNetLARGE, RoBERTaLARGE \cite{liu2019roberta}, ELECTRALARGE \cite{clark2020electra} & Single-sentence and a pair of sentence classification tasks & Vanilla Fine-tuning, CHILD-TUNING, Weight Decay, Top-K Tuning, Mixout, RecAdam, R3F & GLUE \\

2021 \cite{zaken2021bitfit} & BERT \cite{devlin2019bert}, RoBERTa \cite{liu2019roberta} & Sentence level and token level NLP tasks & Full-FT, BitFit & GLUE, PTB POS-tagging \\

2021 \cite{gheini2021cross} & Transformer \cite{vaswani2017attention} & Machine translation & scratch (100\%), {src,tgt} (8\%), {src,tgt}+body (75\%), {src,tgt}+xattn (17\%), {src,tgt}+randxattn (17\%) & Ro–En, Ja–En, De–En, Ha–En, Fr–Es, Fr–De \\

2021 \cite{hu2021lora} & GPT-2, GPT-3 \cite{liu2024gpt} & Natural language to SQL queries, natural language inference, conversation summarization, natural language generation tasks & LoRA, FT, BitFit, Adapter, Prefix-layer tuning, Prefix-embedding tuning & E2E, WikiSQL, MultiNLI, SAMSum \\

2021 \cite{henderson2021compacter} & T5 \cite{2020t5} & Natural language understanding tasks & ADAPTER, PFEIFFER-ADAPTER, ADAPTERDROP, ADAPTER-LOWRANK, PROMPT TUNING, INTRINSIC-SAID, BITFIT, PHM-ADAPTER, COMPACTER & GLUE, SuperGLUE \\
\bottomrule
\end{tabular}
\caption{Parameter-Efficient Fine-Tuning (PEFT) Methods in NLP Models (2019-2021)}
\label{tab:peft_methods_nlp4}
\end{table}

\begin{table*}[t!]
\centering

\caption{Parameter-Efficient Fine-Tuning (PEFT) Methods in Vision Models (2023-2024)}
\label{tab:peft_methods_vision5}
\end{table*}

\begin{table}[t!]
\centering
\scriptsize
\setlength{\tabcolsep}{2pt}
\newcolumntype{u}[1]{>{\raggedright\arraybackslash\hspace{0pt}}p{#1}}
\begin{tabular}{@{}l u{2.8cm} u{3.4cm} u{3.6cm} u{3.2cm}@{}}
\toprule
\textbf{Year} & \textbf{Model} & \textbf{Application} & \textbf{PEFT Method} & \textbf{Datasets} \\
\midrule
2023 \cite{he2023sensitivity} & ViT-B/16 \cite{dosovitskiy2020image} & Recognition tasks & MLP-3, PROMPT-SHALLOW, PROMPT-DEEP, ADAPTER, ADAPTFORMER, NOAH, SPT-ADAPTER, LINEAR, PARTIAL-1, BIAS, LORA, SPT-LORA, SPT-ADAPTER (FULL), VPT & CUB-200-2011, NABirds, Oxford Flowers, Stanford Cars, Stanford Dogs, VTAB-1k \\

2023 \cite{chen2023understanding} & ResNet-18, ResNet-50  \cite{he2016deep} & Transfer learning tasks & RLMVP, FLM-VP, ILM-VP & Flowers102, DTD, UCF101, Food101, GTSRB, SVHN, EuroSAT, OxfordPets, StanfordCars, SUN397, CIFAR10/100, ABIDE \\

    2023 \cite{gao2023unified} & ViT-B/16  \cite{dosovitskiy2020image} & Continual learning tasks & Sequential finetuning, LAE, Joint-FT, Prompt Tuning, Prefix Tuning, LoRA, Adapter & CIFAR100, ImageNet-R \\
    
    2023 \cite{jie2023fact} & ViT-B/16  \cite{dosovitskiy2020image} & Classification tasks & BitFit, VPT-Shallow, VPT-Deep, Adapter, AdapterFormer, LoRA, NOAH, Full FT, Linear probing, FacT-TT, FacT-TK & VTAB-1K \\

    2023 \cite{zha2023instance} & Point-BERT  \cite{devlin2019bert}, Point-MAE  \cite{bank2023autoencoders}, ACT  \cite{dong2022autoencoders} & Object classification, few-shot learning, part segmentation & Full FT, IDPT, VPT & ScanObjectNN, ModelNet40 \\

    2023 \cite{zhu2023visual} & RGB-based foundation tracking model & Multimodal tracking & Full FT, ViPT, Prompt-shaw, Prompt-deep, ViT-shaw & Depthtrack, VOT22RGBD, RGBT234, LasHeR, VisEvent \\

2022 \cite{chen2022vision} & ViT  \cite{dosovitskiy2020image} & Object detection, instance segmentation, and semantic segmentation & PVT-Tiny, PVTv2-B, ViT, ViTDet, ViT-Adapter & COCO, ADE20K \\
\bottomrule
\end{tabular}
\caption{Parameter-Efficient Fine-Tuning (PEFT) Methods in Vision Models (2022-2020)}
\label{tab:vision-peft6}
\end{table}

\end{document}